\RequirePackage{fix-cm}
\documentclass[10pt,twocolumn,letterpaper]{article}
\ifdefined\pdfcompresslevel\pdfcompresslevel=9\fi
\ifdefined\pdfobjcompresslevel\pdfobjcompresslevel=3\fi
\usepackage[pagenumbers]{cvpr}
\usepackage[T1]{fontenc}
\usepackage{microtype}
\usepackage{mathtools}
\usepackage{array,tabularx,multirow,makecell,colortbl}
\usepackage{adjustbox}
\usepackage{listings}
\usepackage{needspace,placeins}
\newcommand{\beginsupplement}{%
  \clearpage\onecolumn\appendix
  \setcounter{page}{1}
  \raggedbottom
  \renewcommand{\thepage}{S\arabic{page}}
  \ifdefined\nolinenumbers\nolinenumbers\fi
  \begin{center}
    {\Large\bfseries\thetitle\par}
    \vspace{0.5em}{\large Supplementary Material\par}
  \end{center}\vspace{0.8em}
}

\usepackage{algorithm,algpseudocode,tikz,titletoc}
\usepackage{amsmath,amsfonts,bm}

\def\eqref#1{equation~\ref{#1}}

\def\1{\bm{1}}

\DeclareMathAlphabet{\mathsfit}{\encodingdefault}{\sfdefault}{m}{sl}
\SetMathAlphabet{\mathsfit}{bold}{\encodingdefault}{\sfdefault}{bx}{n}

\renewcommand{\eqref}[1]{\textup{(\ref{#1})}}
\definecolor{vlmlistingbg}{HTML}{F7F8FA}
\definecolor{vlmlistingrule}{HTML}{CDD3DA}
\definecolor{vlmlistingnum}{HTML}{68717A}
\colorlet{meowDarkGray}{black!85}
\newcolumntype{N}{>{\centering\arraybackslash}X}
\definecolor{EFAccent}{HTML}{21658B}
\colorlet{EFTableHeader}{EFAccent!12}
\colorlet{EFTableGroup}{black!4}
\colorlet{EFTableOurs}{EFAccent!10}
\colorlet{EFTableVariant}{EFAccent!5}
\newcommand{\EFTableTypography}{%
  \small\setlength{\tabcolsep}{4pt}\renewcommand{\arraystretch}{1.08}}

\newcommand{\EFWrapTypography}{%
  \footnotesize\setlength{\tabcolsep}{2pt}\renewcommand{\arraystretch}{1.08}}
\newcommand{\EFMetricGroup}[1]{\textbf{#1}~$\uparrow$}

\titlecontents{appsection}[1.8em]{\addvspace{4pt}\small\bfseries}
  {\contentslabel{1.8em}}{}{\titlerule*[0.65em]{.}\contentspage}
\titlecontents{appsubsection}[4.4em]{\small}
  {\contentslabel{2.6em}}{}{\titlerule*[0.65em]{.}\contentspage}


\usepackage{fontawesome5}
\definecolor{cvprblue}{rgb}{0.21,0.49,0.74}
\usepackage[pagebackref,breaklinks,colorlinks,allcolors=cvprblue]{hyperref}
\hypersetup{pdftitle={From Scores to Samples: Elastic Forcing for Autoregressive Video Generation},
  pdfauthor={Chi Zhang, Yueyi Liu, Shi Haoyang, Ruichuan An, Haoyu Li, Yuhang Wu, Sen Cui, Miao Liu}}
\def\confName{CVPR}
\def\confYear{2027}
\newif\ifincludesupplement
\includesupplementtrue
\title{From Scores to Samples: Elastic Forcing for Autoregressive Video Generation}
\author{Chi Zhang\textsuperscript{1*$\dagger$}\quad Yueyi Liu\textsuperscript{1,2*}\quad
Shi Haoyang\textsuperscript{1,3*}\quad Ruichuan An\textsuperscript{4}\\
Haoyu Li\textsuperscript{2}\quad Yuhang Wu\textsuperscript{1}\quad
Sen Cui\textsuperscript{1}\quad Miao Liu\textsuperscript{1$\ddagger$}\\[3pt]
{\normalsize\textsuperscript{1}College of AI, Tsinghua University\quad
\textsuperscript{2}IAIR, Xi'an Jiaotong University}\\
{\normalsize\textsuperscript{3}Xianghui Academy, Fudan University\quad
\textsuperscript{4}Peking University}\\
{\tt\small imzc.2004@gmail.com\quad miaoliu@mail.tsinghua.edu.cn}\\
{\small\textsuperscript{*}Equal contribution.\quad\textsuperscript{$\dagger$}Project lead.\quad
\textsuperscript{$\ddagger$}Corresponding author.}}

\newcommand{\projectlinks}{%
  \href{https://zhangchi2004.github.io/ElasticForcing/}{\faGlobe\ Project Page}\qquad
  \href{https://github.com/zhangchi2004/ElasticForcing}{\faGithub\ Code}\qquad
  \href{https://huggingface.co/collections/liuyueyi-8/elastic-forcing-6aba0e62a4416b13ae7b13bf}{\raisebox{-0.15em}{\includegraphics[height=1.05em]{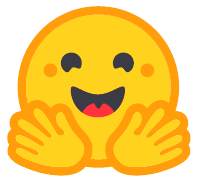}}\ Model}}
\newcommand{\firstpageteaser}{%
  \vspace{-1.5em}
  \begin{center}\small\projectlinks\end{center}
  \vspace{0.1em}
  \begin{minipage}{\textwidth}
  \centering
  \includegraphics[width=\linewidth]{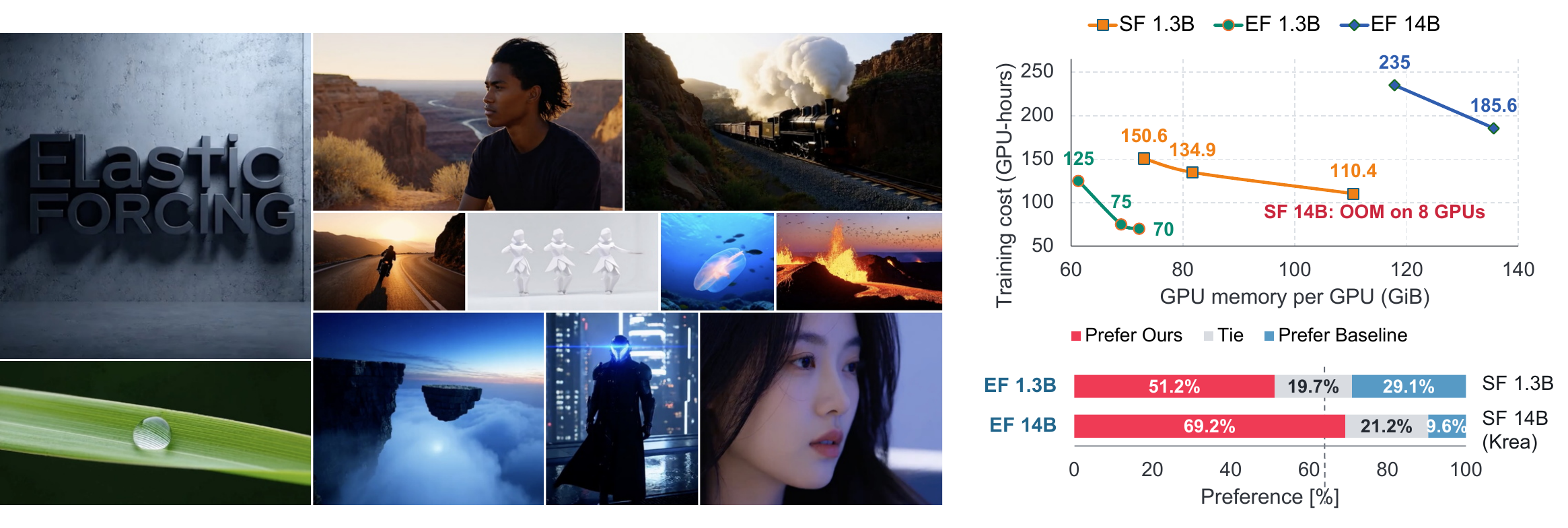}
  \captionof{figure}{\textbf{Elastic Forcing enables efficient few-step autoregressive video generation.}
Left: five-step samples from our Elastic Forcing 14B model.
Right: post-training cost from an initialization checkpoint versus per-GPU memory usage, and human preferences.
At 1.3B, EF wins 51.2\% of comparisons against Self-Forcing, loses 29.1\%, and ties 19.7\%.
At 14B, EF wins 69.2\% against Self-Forcing 14B (Krea Realtime), loses 9.6\%, and ties 21.2\%.}
  \label{fig:teaser}
  \label{fig:offloading}
  \end{minipage}
  \par\vspace{1em}}
\makeatletter
\apptocmd{\@maketitle}{\firstpageteaser}{}{\PackageError{ElasticForcing}{Cannot attach first-page teaser}{}}
\makeatother

\begin{document}
\maketitle
\begin{abstract}
Few-step autoregressive video generation commonly relies on Distribution Matching Distillation (DMD), which requires a bidirectional diffusion teacher and an auxiliary fake-score model. We introduce \textbf{Elastic Forcing}, which learns the rollout distribution directly from reference videos by minimizing maximum mean discrepancy (MMD) in frozen representation spaces. To capture both spatial appearance and temporal dynamics, we match distributions in complementary frozen image and self-supervised video representation spaces. To make this matching efficient, estimation--differentiation decoupling uses a larger rollout population to estimate MMD and a smaller subset to update the generator. With the same architecture and initialization as Self Forcing, our 1.3B model improves VBench Total from 83.95 to 84.25 while retaining 17 FPS inference. Removing auxiliary score models also enables 14B post-training on eight H200 GPUs. Human evaluation favors our models over Self-Forcing at 1.3B and Self-Forcing 14B (Krea Realtime). Combining teacher-forcing initialization with real-video references further enables autoregressive adaptation without teacher-generated trajectories or references.

\end{abstract}
\section{Introduction}
\label{sec:introduction}
Diffusion models have substantially advanced video synthesis~\citep{ho2022video,wan2025}. Streaming video generation~\citep{chen2024diffusion,guo2025resampling,lin2025aapt,yang2025longlive} is becoming increasingly important for applications that require interactive and continuous synthesis, including game simulation~\citep{ICLR2025_b71ecea2,he2025matrix}, virtual livestream interaction~\citep{zhang2026interacvid}, world modeling~\citep{bruce2024genie,NEURIPS2024_7dbb5bfa}, and embodied intelligence~\citep{wu2023unleashinglargescalevideogenerative,cheang2024gr2generativevideolanguageactionmodel}. Streaming requires both causal continuation and low-latency synthesis, motivating \emph{few-step autoregressive models} that generate each chunk through a short denoising trajectory~\citep{yin2025causvid,huang2025selfforcing}. To train these models under their own generated histories, Self Forcing~\citep{huang2025selfforcing} performs distributional post-training on the generator's autoregressive rollouts.

Recent rollout-based video generators~\citep{huang2025selfforcing,yang2025longlive,lu2025reward} commonly adopt Distribution Matching Distillation (DMD)~\citep{yin2024onestep,yin2024improved}, which obtains distributional gradients from a pretrained bidirectional diffusion teacher and an auxiliary model estimating the generated distribution's scores. Maintaining these score networks incurs substantial training overhead, while the teacher's distribution defines the student's supervision target. This raises a natural question: \emph{can we optimize autoregressive rollouts directly against reference videos, without relying on scoring models?}

Maximum Mean Discrepancy (MMD)~\citep{gretton2012kernel} offers a sample-based alternative by matching distributions through kernel comparisons, and has been explored for image generation~\citep{li2015generative,dziugaite2015training,li2017mmdgan,irdm}. Recent representation-based generation methods further motivate comparing samples in frozen pretrained feature spaces~\citep{drifting,irdm,meng2026virdm}. However, extending this approach to causal video introduces two challenges. First, image representations capture spatial appearance but provide weak constraints on temporal dynamics. Second, generating and backpropagating through video rollouts is expensive, making the large batches used in image distribution matching impractical, while small batches yield noisier MMD estimates~\citep{binkowski2018demystifying,irdm}. A practical sample-based approach to video generation must therefore \textit{capture both spatial and temporal structure while obtaining reliable distribution estimates under a limited backpropagation budget.}

We introduce \textbf{Elastic Forcing}, an efficient sample-based framework designed to address these challenges in post-training few-step autoregressive video generators. We retain Self Forcing's rollout procedure and train the generator with MMD over complementary spatial and temporal representations. To make this objective practical for video, we further decouple distribution estimation from generator differentiation. Reference videos directly define the training target, eliminating the need for either an online diffusion teacher or a fake-score model during distributional post-training.

The first design choice concerns what the distributional objective should measure. Our spatial and temporal perturbation diagnostics reveal complementary strengths in image and self-supervised video representations: image features are sensitive to spatial distortions, while video features detect changes in temporal order. Combining them therefore constrains both visual appearance and temporal evolution. The second design choice concerns how this discrepancy can be estimated efficiently. Reliable MMD estimates benefit from more generated samples, but differentiating through every video rollout is prohibitively expensive. Our key observation is that comparing more videos does not require backpropagating through all of them. \emph{Estimation--differentiation decoupling} uses a larger rollout population to estimate MMD and a smaller subset to update the generator, allowing more videos to guide each update under a limited backpropagation budget.

Experiments show that Elastic Forcing improves VBench~\citep{huang2023vbench} performance over Self Forcing with references sampled from the same teacher distribution, while retaining the same inference efficiency. It also scales to 14B post-training on a single node of eight H200 GPUs. Human evaluation favors EF over Self-Forcing by 51.2\% versus 29.1\% at 1.3B (19.7\% ties), and over Self-Forcing 14B (Krea Realtime~\citep{krea_realtime_14b}) by 69.2\% versus 9.6\% (21.2\% ties). Ablations support both the complementary representations and the decoupled training strategy. Beyond teacher-distribution matching, combining teacher-forcing initialization with real videos from UltraVideo~\citep{xue2025ultravideo} enables adaptation without teacher trajectories or teacher-generated references. Together, these results point to a more flexible paradigm for streaming video post-training, where the target distribution is defined by reference data rather than inherited from a fixed diffusion teacher.
\section{Related Work}
\label{sec:related-work}

\begin{figure*}[t]
    \centering
    \includegraphics[width=\linewidth]{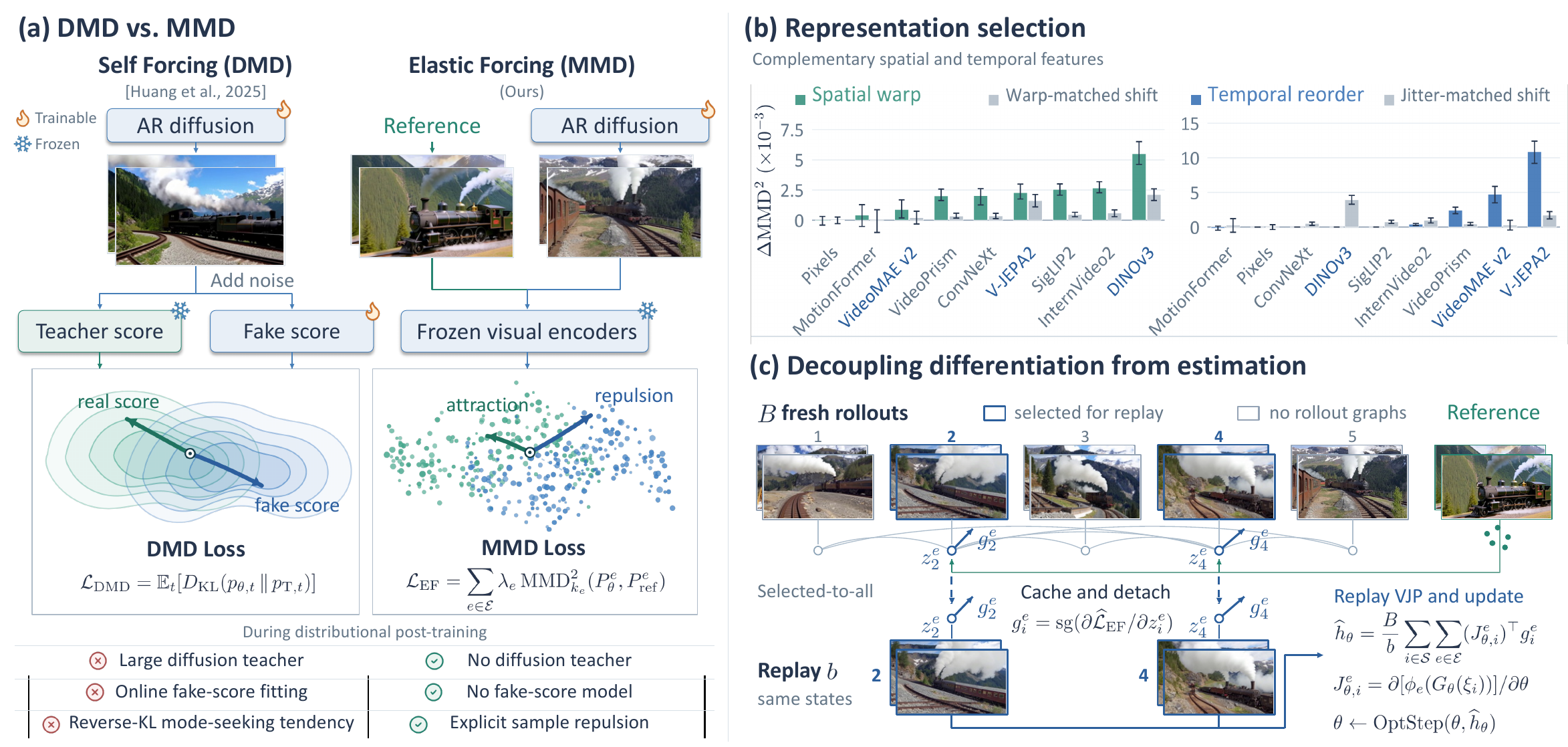}

   \caption{\textbf{Overview of Elastic Forcing.}
(a) Schematic comparison of DMD score supervision and MMD sample matching
in frozen video representations.
(b) Spatial and temporal diagnostics: changes measured
in unbiased MMD squared. Gray bars are rigid shifts matched to warp
RGB error (left) and jitter RGB error (right). Error bars are pointwise 95\% bootstrap intervals.
Suppl.~\ref{app:representation-diagnostic} gives the protocol.
(c) All $B$ fresh rollouts supply comparison features; only the selected
$b$ feature gradients are computed, cached, and differentiated.}
    
    \label{fig:elastic-forcing-overview}
    \label{fig:representation-diagnostic}
\end{figure*}

\subsection{Autoregressive Video Generation}
\label{sec:rw-autoregressive}

Autoregressive video models generate tokens or latent chunks sequentially~\citep{yan2021videogpt,hong2022cogvideo,kondratyuk2023videopoet,deng2024nova}.
Diffusion Forcing and Self-Resampling address the training--inference gap~\citep{chen2024diffusion,guo2025resampling,song2025history}; subsequent work improves long-form synthesis and interactive control~\citep{ai2025magi1autoregressivevideogeneration,chen2025skyreelsv2infinitelengthfilmgenerative,liu2025rolling,li2025stableinfinity,lin2025aapt,yang2025longlive,lu2025reward}.
CausVid distills a few-step causal generator, while Self Forcing trains on generated histories~\citep{yin2025causvid,huang2025selfforcing}.
EF retains this rollout training and replaces score supervision with reference matching.

\subsection{Distributional Distillation and Training}
\label{sec:rw-distillation}
\label{sec:rw-distributional-training}

DMD and SiD derive distributional gradients from teacher scores~\citep{yin2024onestep,yin2024improved,zhou2024sid}; SiDA and DFD retain teacher supervision while adding real-data signals~\citep{zhou2025sida,chen2026dataforcing}.
Direct distributional training instead matches samples through MMD or feature moments~\citep{gretton2012kernel,li2015generative,dziugaite2015training,irdm,yang2026frechet}.
MGFlow connects these objectives to feature-space gradient flows~\citep{zhang2026distributional,drifting}.
ViRDM~\citep{meng2026virdm} is concurrent work that extends representation matching to causal video. It was first released on September 24, after the September 15 cutoff that CVPR uses to define concurrent work. Our work was carried out independently of it, and we therefore do not make a comparison.
EF combines complementary features with gradient caching~\citep{gao2021gradcache,lezama2021runsort} to enlarge the estimation population under a fixed backward budget.

\subsection{Video Representation Learning}
\label{sec:rw-representations}

Self-supervised video encoders learn through masked reconstruction, contrastive learning, or latent prediction~\citep{tong2022videomae,wang2023videomaev2,qian2021cvrl,bardes2024vjepa,assran2025vjepa2}; video--text pretraining adds semantic alignment~\citep{wang2024internvideo2,zhao2024videoprism}.
Image encoders supply complementary spatial features~\citep{simeoni2025dinov3,radford2021clip,tschannen2025siglip2}.
Since video metrics can underweight temporal corruption~\citep{10655312}, EF tests both spatial and temporal sensitivity before selecting its matching spaces.

\section{Elastic Forcing}
\label{sec:method}

Elastic Forcing trains a few-step causal video generator directly from
reference videos. Its design addresses two requirements: representations
that expose spatial and temporal differences, and MMD updates informed by
a large rollout population under a limited backward budget.

\subsection{Preliminaries}
\label{sec:prelim}

\paragraph{Few-step autoregressive rollout learning.}
\label{sec:prelim_ar}
An autoregressive generator models a video
$x=(x_1,\ldots,x_T)$ as a sequence of causal chunks:
\begin{equation}
p_\theta(x)=\prod_{i=1}^{T}p_\theta(x_i\mid x_{<i}).
\label{eq:ar_factorization}
\end{equation}
Each chunk is synthesized through a short denoising
trajectory~\citep{yin2025causvid,huang2025selfforcing}.
Following Self Forcing~\citep{huang2025selfforcing}, we train on rollouts
conditioned on their own generated histories. We replace score-based
supervision with direct matching between the completed-rollout
distribution $p_\theta$ and the empirical distribution $p_{\rm ref}$ of a
fixed reference-video collection.

\paragraph{Maximum mean discrepancy.}
\label{sec:prelim_mmd}
MMD compares distributions through kernel mean
embeddings~\citep{gretton2012kernel,muandet2017kernel}.
For distributions $P,Q$ and a positive-definite kernel $k$,
\begin{equation}
\begin{aligned}
\operatorname{MMD}_{k}^{2}(P,Q)
={}&\mathbb{E}_{z,z'\sim P}[k(z,z')]
+\mathbb{E}_{u,u'\sim Q}[k(u,u')]\\
&-2\mathbb{E}_{z\sim P,\,u\sim Q}[k(z,u)],
\end{aligned}
\label{eq:mmd}
\end{equation}
where draws within each expectation are independent, and expectations are estimated from generated and reference samples.

\subsection{Representation Geometry for Video Matching}
\label{sec:representation_space}

\noindent\textbf{Spatial discrimination from image representations.}
Direct MMD matching in pixel space proved ineffective in our initial trials. This is consistent with the limited perceptual alignment of pixel-space distances~\citep{zhang2018unreasonable}. Following prior work on image generation~\citep{drifting,irdm}, we instead compare video distributions in frozen pretrained feature spaces. The geometry of the chosen representation determines which video differences are reflected in the MMD objective. \textit{However, image representations alone may be insufficient for video matching.} We first evaluate DINOv3~\citep{simeoni2025dinov3}, a strong image encoder, by extracting frame-wise features and stacking them along the temporal dimension. Although DINOv3-based matching produces videos with clear visual content, the generated videos exhibit limited motion, achieving a VBench Dynamic Degree of only 32.22.

To investigate this behavior, we conduct a controlled diagnostic by applying stochastic spatial warping and temporal reordering to a set of reference videos, and measuring the resulting MMD changes under different representation spaces (Figure~\ref{fig:representation-diagnostic}(b)). This experiment evaluates whether each representation can distinguish spatial and temporal changes at the distribution level. Raw pixel-space MMD exhibits near-zero changes under these perturbations, indicating limited sensitivity to the induced distortions. DINOv3 exhibits the strongest response to spatial warping, but its readout is nearly invariant to temporal reordering. Other image encoders, such as ConvNeXt~\citep{liu2022convnext} or SigLIP2~\citep{tschannen2025siglip2}, likewise exhibit near-zero MMD responses to temporal reordering.

\noindent\textbf{Self-supervised video representations supply temporal discrimination.}
We next investigate whether motion-oriented video encoders can address the temporal blindness of image representations. However, Motionformer~\citep{patrick2021motionformer} exhibits no clear MMD response to either spatial warping or temporal reordering in our diagnostic. This suggests that motion sensitivity alone is insufficient to provide a useful geometry for distribution matching. We hypothesize that \textit{effective matching spaces require semantically structured representations that capture the spatiotemporal properties of complete videos.}

We therefore consider self-supervised video representations, including V-JEPA~2~\citep{assran2025vjepa2}, VideoMAE~v2~\citep{wang2023videomaev2}, VideoPrism~\citep{zhao2024videoprism}, and InternVideo2~\citep{wang2024internvideo2}. These encoders exhibit clear MMD responses to both spatial warping and temporal reordering, indicating their ability to capture spatial and temporal distributional differences. However, their responses to spatial warping are weaker than DINOv3's, suggesting weaker spatial representations. This motivates combining video and image representations to provide complementary constraints.

\noindent\textbf{Combining spatial and temporal constraints.}
We combine DINOv3, V-JEPA~2, and VideoMAE~v2. Each frozen encoder $\phi_e$
maps a complete video to a vector of ordered patch features. We stack DINO representations of orderly selected frames along the temporal axis, and retain ordered token selections for the video encoders (details in Suppl.~\ref{app:training}).
Let $P_\theta^e$ and $P_{\rm ref}^e$ denote the feature distributions
obtained by applying $\phi_e$ to $p_\theta$ and $p_{\rm ref}$. We minimize
\begin{equation}
\mathcal L_{\rm EF}(\theta)
 =\sum_{e\in\mathcal E}\lambda_e
   \operatorname{MMD}_{k_e}^{2}(P_\theta^e,P_{\rm ref}^e),
\label{eq:ef_objective}
\end{equation}
with fixed nonnegative weights $\lambda_e$ and Gaussian kernels
$k_e(z,u)=\exp(-\lVert z-u\rVert_2^2/(2\sigma_e^2))$ of fixed bandwidth. Bandwidth is selected based on the numerical patterns of the distribution of each encoder.
Each complete video supplies one MMD sample.
Gaussian kernels are characteristic, so population MMD is zero only when
the compared feature distributions coincide~\citep{sriperumbudur2010hilbert,gretton2012kernel}.

\subsection{Decoupling Estimation from Differentiation}
\label{sec:decoupled_mmd}

\noindent\textbf{The batch-size challenge in video.}
The choice of representation determines which video differences MMD detects, while the reliability of distribution matching also depends on the number of generated samples. Recent image-generation methods use global batches of 8192~\citep{drifting} or 10240~\citep{irdm} images per update. Such batch sizes are impractical for video generation due to the substantially higher cost of video rollouts. Small sample sizes increase sampling uncertainty in MMD estimation~\citep{gretton2012kernel}, and we observe unstable gradients when reducing the video batch size. We address this challenge by decoupling the rollout population used for MMD estimation and for generator backpropagation.

\noindent\textbf{Estimating MMD over the full rollout batch.}
We first generate $B>1$ fresh rollouts and extract their features
$z_i^e=\phi_e(x_i)$ without retaining the generator computation graphs.
This allows all $B$ rollouts to participate in distribution matching
without storing the activations required for backpropagation.
On the reference side, we approximate the kernel mean
$m_{\rm ref}^e(z)=\mathbb E_{u\sim P_{\rm ref}^e}k_e(z,u)$
using our hybrid estimator $\widehat m_{\rm ref}^e$, which combines
Nystr\"om kernel-mean approximations~\citep{williams2000nystrom,chatalic2022nystrom}
and Monte Carlo estimates
(Suppl.~\ref{app:reference_estimation}). Dropping the constant
reference--reference term in Equation~\eqref{eq:mmd} gives
\begin{equation}
\widehat{\mathcal L}_{\rm EF}
=\frac{1}{B}\sum_{i,e}\lambda_e\left[
\frac{\sum_{j\ne i}k_e(z_i^e,z_j^e)}{B-1}
-2\widehat m_{\rm ref}^e(z_i^e)\right].
\label{eq:finite_batch_loss}
\end{equation}
The generated--generated term captures pairwise similarities among
current rollouts, while the cross term attracts them toward the
reference distribution~\citep{arbel2019mmdflow}.

\noindent\textbf{Selective backward through replay.}
To limit backward cost, we uniformly sample $b\le B$ distinct indices
$\mathcal S$, while retaining all $B$ rollouts in the loss. For each
$i\in\mathcal S$, we cache its feature gradient (cotangent)
$g_i^e=\nabla_{z_i^e}\widehat{\mathcal L}_{\rm EF}$:
\begin{equation}
g_i^e
=\frac{2\lambda_e}{B}\left[
\frac{\sum_{j\ne i}\nabla_1 k_e(z_i^e,z_j^e)}{B-1}
-\nabla\widehat m_{\rm ref}^e(z_i^e)\right],
\label{eq:representation_cotangent}
\end{equation}
where $\nabla_1$ denotes differentiation with respect to the first
kernel argument. Each cached gradient therefore incorporates all
$B-1$ other rollouts and the reference estimate, although only $b$
rollouts require cotangent computation.

We then replay the selected rollouts in microbatches using their saved
random choices and causal context~\citep{gao2021gradcache,lezama2021runsort,yang2026frechet}.
This recovers the same features $\widetilde z_i^e=z_i^e$ with the graphs
needed for backpropagation. With
$J_{\theta,i}^e=\partial\widetilde z_i^e/\partial\theta$ and the cached
$g_i^e$ held fixed, we compute
\begin{equation}
\begin{aligned}
v
=\frac{B}{b}\sum_{i\in\mathcal S,e}
(J_{\theta,i}^e)^\top g_i^e
=\nabla_\theta\left[
\frac{B}{b}\sum_{i\in\mathcal S,e}
\langle g_i^e,\widetilde z_i^e\rangle
\right].
\end{aligned}
\label{eq:selective_backward}
\end{equation}
This inner-product backward pass propagates through the frozen encoders
to the generator without forming the Jacobians explicitly. Conditional
on the rollouts and reference estimate, each index has inclusion
probability $b/B$. The factor $B/b$ thus makes $v$ unbiased for the
full-batch generator gradient under the same replay computation~\citep{horvitz1952sampling}
(Suppl.~\ref{app:selective_backward}). Thus, all $B$ rollouts inform
distribution matching, while generator backpropagation involves only
the selected $b$.

\begin{table*}[tp]
\centering
\EFTableTypography
\setlength{\tabcolsep}{3pt}
\renewcommand{\arraystretch}{1.12}
\captionof{table}{\textbf{Video generation on VBench and VLM evaluation}.
Models are grouped by generation scheme. Bold marks the best quality scores among autoregressive models.
Qual.: visual quality; Sem.: semantic consistency; Subj.: subject
consistency; Smoo.: motion smoothness. Wan and UV denote Wan-generated
and UltraVideo reference videos, respectively.}
\label{tab:short-video-comparison}
\begin{tabularx}{\linewidth}{lcc*{8}{N}}
\toprule
\rowcolor{EFTableHeader}
& & & \multicolumn{3}{c}{\EFMetricGroup{VBench}}
& \multicolumn{5}{c}{\EFMetricGroup{VLM scores}} \\
\rowcolor{EFTableHeader}
\multirow{-2}{*}{\textbf{Model}}
& \multirow{-2}{*}{\textbf{\# Params}}
& \multirow{-2}{*}{\makecell{\textbf{Throughput}\\\EFMetricGroup{(FPS)}}}
& \textbf{Total} & \textbf{Qual.} & \textbf{Sem.}
& \textbf{Total} & \textbf{Qual.} & \textbf{Sem.} & \textbf{Subj.} & \textbf{Smoo.} \\
\midrule
\rowcolor{EFTableGroup}
\multicolumn{11}{l}{\textit{Bidirectional Video Diffusion Models}} \\
LTX-Video~\citep{hacohen2024ltxvideo} & 1.9B & 8.98 & 80.00 & 82.30 & 70.79
& 4.01 & 3.87 & 3.94 & 4.16 & 4.07 \\
Wan 2.1~\citep{wan2025} & 1.3B & 0.78 & 84.26 & 85.30 & 80.09
& 4.44 & 4.35 & 4.48 & 4.60 & 4.34 \\
\midrule
\rowcolor{EFTableGroup}
\multicolumn{11}{l}{\textit{Autoregressive Video Diffusion Models}} \\
NOVA~\citep{deng2024nova} & 0.6B & 0.88 & 79.29 & 79.91 & 76.81
& 3.48 & 3.10 & 4.10 & 3.26 & 3.46 \\
Pyramid Flow~\citep{jin2024pyramidal} & 2.0B & 6.70 & 81.63 & 83.78 & 73.02
& 3.94 & 3.83 & 3.96 & 3.96 & 3.99 \\
SkyReels-V2~\citep{chen2025skyreelsv2infinitelengthfilmgenerative} & 1.3B & 0.49 & 82.26 & 83.61 & 76.85
& 4.48 & \textbf{4.42} & 4.26 & \textbf{4.73} & \textbf{4.52} \\
MAGI-1~\citep{ai2025magi1autoregressivevideogeneration} & 4.5B & 0.19 & 82.28 & 82.73 & 80.47
& 4.35 & 4.23 & 4.31 & 4.53 & 4.33 \\
\midrule
\rowcolor{EFTableGroup}
\multicolumn{11}{l}{\textit{Few-Step Autoregressive Video Models}} \\
CausVid~\citep{yin2025causvid} & 1.3B & 17.0 & 83.02 & 84.16 & 78.45
& 4.29 & 4.16 & 4.34 & 4.42 & 4.23 \\

Self-Forcing~\citep{huang2025selfforcing} & 1.3B & 17.0
& 83.95 & 84.73 & 80.83
& 4.39 & 4.29 & 4.50 & 4.45 & 4.33 \\
\rowcolor{EFTableVariant}
{\color{EFAccent}\bfseries Elastic Forcing (Wan)} & 1.3B & 17.0
& \textbf{84.25} & \textbf{85.06} & \textbf{80.99}
& 4.48 & 4.33 & 4.67 & 4.67 & 4.26 \\
\rowcolor{EFTableOurs}
{\color{EFAccent}\bfseries Elastic Forcing (UV)} & 1.3B & 17.0
& 83.65 & 84.42 & 80.57
& \textbf{4.49} & 4.29 & \textbf{4.70} & 4.69 & 4.27 \\
\bottomrule
\end{tabularx}

\end{table*}

\begin{algorithm}[t]
    \caption{Elastic Forcing training}
    \label{alg:elastic-forcing}
    \small
    \makeatletter
    \providecommand*{\theHALG@line}{}
    \renewcommand*{\theHALG@line}{\thealgorithm.\arabic{ALG@line}}
    \makeatother
    \begin{algorithmic}[1]
        \raggedright
        \Require Generator $G_\theta$ with parameters $\theta$;
        frozen encoders $\{\phi_e\}_{e\in\mathcal E}$;
        hybrid reference kernel-mean estimators
        $\{\widehat m_{\rm ref}^e\}_{e\in\mathcal E}$;
        distribution batch size $B$, differentiation batch size $b\le B$;
        number of updates $K$
        \For{$t=1,\ldots,K$}
            \State \textbf{Generate:} $\{(x_i,s_i)\}_{i=1}^{B}\gets
            \operatorname{Rollout}_\theta(B)$
            \Statex \hspace{\algorithmicindent}\textit{No retained graph;
            $s_i$ stores replay state.}
            \State \textbf{Encode:} $z_i^e\gets\phi_e(x_i)$ for all $i,e$
            \State Sample $b$ distinct indices $\mathcal S$ uniformly
            \State \textbf{Cache:}
            $g_i^e\gets\operatorname{Detach}\!\left(
            \nabla_{z_i^e}\widehat{\mathcal L}_{\rm EF}\right)$, $i\in\mathcal S$
            \Comment{Eq.~\eqref{eq:representation_cotangent}}
            \State \textbf{Replay:} $\widetilde z_i^e\gets
            \phi_e(\operatorname{Replay}_\theta(s_i))$, $i\in\mathcal S$
            \State \textbf{Accumulate:} $v\gets\frac{B}{b}\sum_{i\in\mathcal S,e}
            \nabla_\theta\langle g_i^e,\widetilde z_i^e\rangle$
            \Statex \hspace{\algorithmicindent}\textit{Replay and
            accumulation use microbatches.}
            \State \textbf{Update:}
            $\theta\gets\operatorname{OptimizerStep}(\theta,v)$
        \EndFor
    \end{algorithmic}
\end{algorithm}

\subsection{Overall Training}
\label{sec:overall_training}

Algorithm~\ref{alg:elastic-forcing} summarizes training.
We precompute reference features and
a persistent Nystr\"om summary, which is combined with sampled
reference features at each update to evaluate the hybrid
reference estimator (Suppl.~\ref{app:hybrid-definition}). Following Self Forcing's causal rollout
procedure~\citep{huang2025selfforcing}, we generate $B$ fresh
videos and compute their representations using frozen encoders.
All $B$ videos contribute to the MMD objective, while only
$b$ selected rollouts are replayed to propagate the cached
feature gradients to the generator. Replay preserves the
sampled denoising exits and stop-gradient boundaries of the
original rollout. Each microbatch's graph is released after backward,
while generator parameters remain fixed until all selected rollouts
have contributed. The optimizer then takes one step with the accumulated
gradients. See Suppl.~\ref{app:rollout-gradients}
and~\ref{app:selective_backward} for implementation details.

Elastic Forcing thus enables direct distribution matching of
complete causal rollouts without requiring an online diffusion
teacher or fake-score model. At inference, only the original
causal generator is used.

\begin{figure*}[tp]
    \centering
    \includegraphics[width=\linewidth,height=0.60\textheight,keepaspectratio]{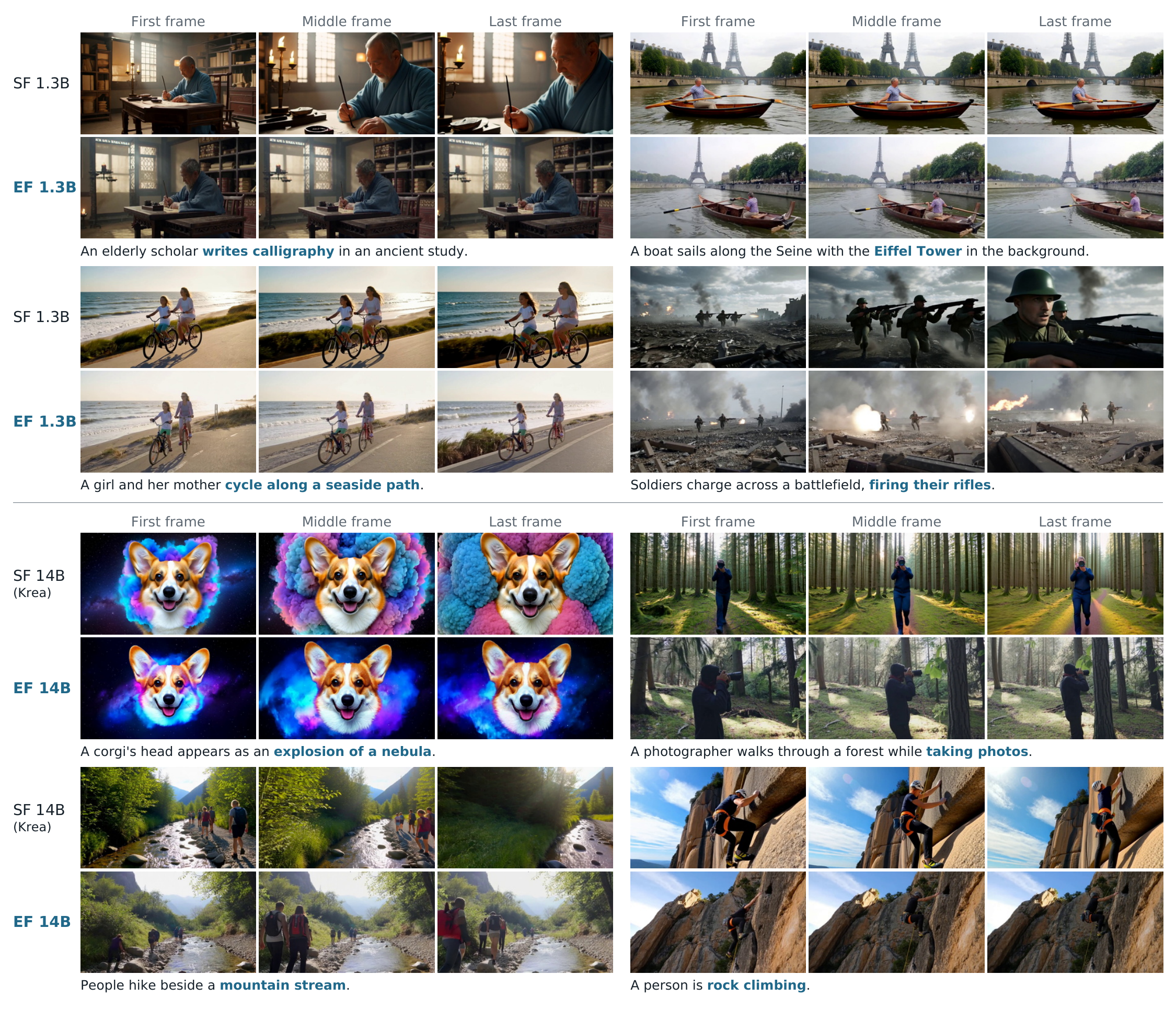}
    \caption{\textbf{Qualitative comparisons across model scales.}
    SF: Self-Forcing; EF: Elastic Forcing. Both EF models use UltraVideo
    references.
    The examples highlight surface texture and tonal variation relative
    to the corresponding baseline.}
    \label{fig:main-qualitative-comparison}
\end{figure*}

\section{Experiments}
\label{sec:experiments}

\subsection{Experimental Setup}
\label{sec:experimental-setup}

\noindent\textbf{Models and reference datasets.}
We evaluate Elastic Forcing (EF) on Wan2.1 backbones at 1.3B and
14B scales~\citep{wan2025}, using Self-Forcing's causal architecture
and rollout procedure~\citep{huang2025selfforcing}.
We train on two reference collections of equal size:
\emph{Wan} contains 8,000 videos generated by the Wan2.1 model at the
corresponding parameter scale, using VidProM
prompts~\citep{wang2024vidprom}; \emph{UV} contains 8,000 videos randomly
sampled from the UltraVideo~\citep{xue2025ultravideo} dataset.
We denote the resulting models by EF (Wan) and EF (UV).

\noindent\textbf{Training settings.}
The default matching spaces are DINOv3, V-JEPA~2, and VideoMAE~v2,
with loss weights $(10,1,1)/3$ and a 2:1 Nystr\"om--MC reference mixture.
At 1.3B, we use distribution and differentiation batches $B=256$ and $b=128$.
The standard three-encoder models train for 150 updates on four H200 GPUs
at 1.3B, and 80 updates on eight H200 GPUs at 14B. These training steps are chosen to balance computational cost and generation quality. The training dynamics of the remaining steps are provided in Supplementary Sec.~B.
Inference uses four denoising steps per chunk at 1.3B and five at 14B.
The standard runs start from Self-Forcing and Krea ODE-initialization checkpoints, respectively~\citep{huang2025selfforcing,krea_realtime_14b};
Section~\ref{sec:teacher-free} studies teacher-forcing initialization.
Detailed settings are in
Suppl.~\ref{app:implementation}.

\noindent\textbf{VBench.}
Following prior work, we report VBench Total, Quality, and Semantic
scores~\citep{huang2023vbench}, using five samples from fixed seeds for each of 946
VBench prompts. However, recent studies identify gaps in evaluating diversity and physical plausibility~\citep{zheng2025vbench2}, while VLM-based evaluators demonstrate better alignment with human preference~\citep{he2025videoscore2}.

\noindent\textbf{VLM based evaluation.}
We therefore introduce a model-blind GPT-6 evaluation alongside VBench.
Given the prompt and 16 uniformly sampled frames, the evaluator
scores visual quality, semantic consistency, subject consistency, and
motion smoothness. We divide the raw 0--100 scores
by 20 to report a 0--5 scale; VLM Total is their equal-weight mean.
For 14B comparisons and ablations, GPT-6 also ranks anonymous candidate
videos for each prompt, reading the entire group at one time. We report mean
rank points (higher is better), relative to the competing set and
comparable only within a table.
Suppl.~\ref{sec:vlm-prompt-design} gives the scoring and ranking prompts.


\subsection{Comparison with Baseline Methods}
\label{sec:small-model-comparison}
\label{sec:scaling}

Table~\ref{tab:short-video-comparison} compares Elastic Forcing with
bidirectional diffusion, autoregressive diffusion, and few-step
autoregressive models. Self-Forcing is our primary baseline, as it
shares the same 1.3B backbone, initialization, and causal rollout
procedure. Both EF variants retain its 17 FPS throughput.
We first train EF using Wan 2.1-generated videos as references,
matching the data distribution of the initialization and
Self-Forcing training. EF (Wan) improves VBench Total from 83.95
to 84.25 and VLM Total from 4.39 to 4.48, exhibiting better visual quality, semantic
consistency, and subject consistency.
Replacing generated references with real UltraVideo videos gives a VLM
Total of 4.49. Relative to Self-Forcing, EF (UV) matches visual quality
(4.29) and improves semantic and subject consistency, but lowers motion
smoothness.

We further conduct a human rating experiment with 20 raters. Each rater gives a preference between 2 generated videos from EF(UV) and SF.
EF (UV), which has a lower VBench total score, wins
51.2\% of comparisons versus 29.1\% for Self-Forcing, with 19.7\% ties
(Figure~\ref{fig:teaser}). This discrepancy
suggests that VBench does not fully capture the perceptual gains
brought by real-video references in this evaluation setting.
Figure~\ref{fig:main-qualitative-comparison} provides qualitative results, using UV references for EF. The 1.3B examples show more natural skin, clothing, and environmental textures.

\subsection{Efficient Scaling to Larger Backbones}

\begin{table}[t]
\centering
\EFWrapTypography
\captionsetup{hypcap=false}
\captionof{table}{\textbf{Scaling and reference choice at 14B.} 
UV: UltraVideo references. Wan: Wan-generated reference. Bold marks the best value in each column.}
\label{tab:vlm_comparison}
\begin{tabularx}{\linewidth}{l*{6}{N}}
\toprule
\rowcolor{EFTableHeader}
& \multicolumn{5}{c}{\EFMetricGroup{VLM scores}} & \EFMetricGroup{VLM} \\
\rowcolor{EFTableHeader}
\multirow{-2}{*}{\textbf{Model}} & \textbf{Total} & \textbf{Qual.} & \textbf{Sem.}
& \textbf{Subj.} & \textbf{Smoo.} & \textbf{Rank} \\
\midrule
Wan 2.1 14B & 4.55 & 4.42 & 4.62 & 4.79 & 4.37 & 2.11 \\
Krea &4.45& 4.29 & \textbf{4.79} & 4.50 & 4.21 & 2.49 \\
\rowcolor{EFTableVariant}
{\color{EFAccent}\bfseries EF (Wan)} & 4.59 & \textbf{4.47} & 4.77 & 4.77 & 4.36 & 2.67 \\
\rowcolor{EFTableOurs}
{\color{EFAccent}\bfseries EF (UV)} & \textbf{4.60} & 4.46 & 4.72 & \textbf{4.82} & \textbf{4.38} & \textbf{2.74} \\
\bottomrule
\end{tabularx}

\end{table}


\noindent\textbf{Scaling to 14B.}
Krea reports that naive scaling with ZeRO-3~\citep{rajbhandari2020zero}
encounters out-of-memory (OOM) errors
even on 64 H100 GPUs~\citep{millon2025krea}.
EF replaces these generator-scale score networks with frozen visual
encoders and post-trains the 14B backbone in
23.2 wall-clock hours on eight H200 GPUs (185.6 GPU-hours).
Figure~\ref{fig:offloading} shows the corresponding memory--cost trade-off: more aggressive offloading lowers peak GPU memory at the expense of additional computation. Costs include only post-training from the initialization checkpoint. At 1.3B, the plotted EF configurations require 70--125 GPU-hours, compared with 110.4--150.6 for Self-Forcing. At 14B, offloading reduces memory demand but increases EF's cost from 185.6 to 235 GPU-hours. Notably,
Self-Forcing at 14B still encounters OOM even under the most
aggressive offloading configuration.

\noindent\textbf{14B generation quality.}
Both EF variants improve VLM rank over Krea Realtime: 2.67 (Wan) and
2.74 (UV) versus 2.49 (Table~\ref{tab:vlm_comparison}). EF (UV) ranks
highest among the four models and also achieves the highest VLM Total
(4.60). Human evaluation supports this preference: EF (UV) wins 69.2\%
against Self-Forcing 14B (Krea Realtime), versus 9.6\% for the baseline
and 21.2\% ties (Figure~\ref{fig:teaser}).
Figure~\ref{fig:main-qualitative-comparison} shows finer surface detail
and less oversaturated colors in the EF examples; the baseline's
oversaturation is consistent with prior observations of
DMD~\citep{yin2024onestep,chen2026dataforcing}, which is rooted in DMD's mode seeking property of reverse KL divergence.

\subsection{Toward Teacher-Free Autoregressive Training}
\label{sec:teacher-free}
\label{sec:adaptation}

\begin{table}[t]
\centering
\EFWrapTypography
\captionof{table}{\textbf{Toward teacher-free autoregressive training.}
Three-encoder 1.3B models after 150 updates. TF: teacher forcing. Total, Qual., and Sem. are VBench scores. 
The highlighted TF + UV configuration removes ODE-trajectory distillation
and teacher-generated references. }
\label{tab:initialization_data}
\begin{tabularx}{\linewidth}{*{6}{N}}
\toprule
\rowcolor{EFTableHeader}
& & \multicolumn{3}{c}{\EFMetricGroup{VBench}} & \EFMetricGroup{VLM} \\
\rowcolor{EFTableHeader}
\multirow{-2}{*}{\textbf{Refs.}} & \multirow{-2}{*}{\textbf{Init.}}
& \textbf{Total} & \textbf{Qual.} & \textbf{Sem.} & \textbf{Rank} \\
\midrule
Wan & TF & 84.24 & \textbf{85.35} & 79.79 & 2.43 \\
Wan & ODE & \textbf{84.25} & 85.06 & \textbf{80.99} & 2.44 \\
\midrule
UV & ODE & 83.65 & 84.42 & 80.57 & \textbf{2.81} \\
\rowcolor{EFTableOurs}
\textbf{UV} & \textbf{TF} & 83.68 & 85.09 & 78.05 & 2.33 \\
\bottomrule
\end{tabularx}

\end{table}

\noindent\textbf{Removing diffusion-teacher dependence.}
Table~\ref{tab:initialization_data} compares TF and ODE initialization
with Wan and UV references on the same three-encoder 1.3B model after
150 updates. With Wan references, TF nearly matches ODE in VBench
Total (84.24 vs.\ 84.25). Combining TF with UV removes teacher
trajectories and teacher-generated references from post-training while
retaining the pretrained backbone. It achieves 83.68 VBench Total,
although its lower VLM rank than ODE + UV (2.33 vs.\ 2.81) indicates
a remaining quality trade-off.

\begin{figure}[t]
\centering
\includegraphics[width=\linewidth,height=0.48\textheight,keepaspectratio]{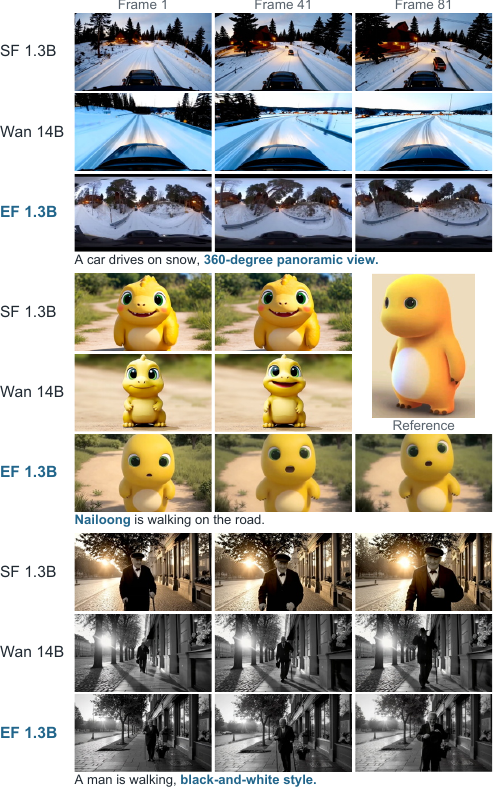}
\caption{\textbf{Changing the reference distribution.} Reference videos
supervise panoramic composition, character identity, and monochrome style.
The separate reference image identifies the target character.}
\label{fig:reference-adaptation}
\end{figure}

\noindent\textbf{Learning other targets from references.}
Figure~\ref{fig:reference-adaptation} illustrates EF 1.3B learning concepts
and styles that Wan 14B fails to generate in these examples. By changing
the reference videos, EF reproduces the Nailoong character and monochrome
appearance and learns panoramic composition. This demonstrates that
reference-defined supervision can teach new visual targets beyond those
expressed by the pretrained generator under the same prompts.
See Suppl.~\ref{app:adaptation} for quantitative results.

\subsection{Ablation Studies}
\label{sec:ablations}

\begin{table}[t]
\centering
\EFWrapTypography
\captionof{table}{\textbf{Encoder selection.} VBench scores and comparative VLM rank points for different matching spaces. VJ:
    V-JEPA~2; VM: VideoMAE; D: DINOv3. All EF variants use 150 updates.}
    \label{tab:video_encoder_ablation}
    \begin{tabularx}{\linewidth}{l*{4}{N}}
        \toprule
        \rowcolor{EFTableHeader}
        & \multicolumn{3}{c}{\EFMetricGroup{VBench}} & \EFMetricGroup{VLM} \\
        \rowcolor{EFTableHeader}
        \multirow{-2}{*}{\textbf{Encoder}}
        & \textbf{Total} & \textbf{Qual.} & \textbf{Sem.} & \textbf{Rank} \\
        \midrule
        \rowcolor{EFTableGroup}
        Self-Forcing & 83.95 & 84.73 & 80.83 & -- \\
        \midrule
        V-JEPA~2 & 83.91 & 84.62 & \underline{81.09}&2.40 \\
        VideoMAE & 83.97 & 85.02 & 79.79 &2.67\\
        DINOv3 & 81.96 & 82.32 & 80.54 &\underline{3.27}\\
        \midrule
        \rowcolor{EFTableOurs}
        \textbf{VJ + VM}
        & \textbf{84.64} & \textbf{85.43} & \textbf{81.48}&3.03 \\
        \rowcolor{EFTableVariant}
        VJ + VM + D
        & \underline{84.25} & \underline{85.06} & 80.99&\textbf{3.63} \\
        \bottomrule
    \end{tabularx}

\end{table}

\noindent\textbf{Representation geometry.}
Combining V-JEPA~2 and VideoMAE improves VBench Total over either encoder
alone (84.64 vs.\ 83.91 and 83.97), with gains in both Quality and Semantic
scores (Table~\ref{tab:video_encoder_ablation}). VLM rank also rises from
2.40 and 2.67 to 3.03, supporting complementary video representations.
DINOv3 alone yields lower VBench Total (81.96) but higher VLM rank (3.27),
showing that the evaluations favor different representations. The
three-encoder checkpoint further improves VLM rank to 3.63, trading a
modest VBench decrease from the pair (84.64 to 84.25) for stronger VLM
preference. Its VBench Total also remains above Self-Forcing (83.95),
supporting the three-encoder default when prioritizing VLM preference.

\begin{table}[t]
\centering
\EFWrapTypography
\captionof{table}{\textbf{Decoupling estimation and differentiation.} All variants use 150 updates. We report VBench Total, Quality (Qual.), Semantic (Sem.), and comparative VLM rank points.}
    \label{tab:sample_ablation}
    \begin{tabularx}{\linewidth}{*{6}{N}}
        \toprule
        \rowcolor{EFTableHeader}
        & & \multicolumn{3}{c}{\EFMetricGroup{VBench}} & \EFMetricGroup{VLM} \\
        \rowcolor{EFTableHeader}
        \multirow{-2}{*}{\textbf{Dist.} $B$} & \multirow{-2}{*}{\textbf{Diff.} $b$}
        & \textbf{Total} & \textbf{Qual.} & \textbf{Sem.} & \textbf{Rank} \\
        \midrule
        256 & 32  & 83.73 & 84.32 & \textbf{81.39} & 2.38 \\
        256 & 64  & 83.98 & 84.66 & 81.23 & 2.54 \\
        \rowcolor{EFTableOurs}
        256 & 128 & \textbf{84.25} & \textbf{85.06} & 80.99 & \textbf{3.04} \\
        \midrule
        128 & 128 & 84.09 & 84.77 & \textbf{81.38} & 2.04 \\
        \rowcolor{EFTableOurs}
        256 & 128 & \textbf{84.25} & \textbf{85.06} & 80.99 & \textbf{3.04} \\
        \bottomrule
    \end{tabularx}

\end{table}

\noindent\textbf{Estimation--differentiation decoupling.}
At a fixed backward budget of $b=128$, increasing the number of rollouts from $B=128$ to $B=256$ raises VBench Total from 84.09 to 84.25 and VLM rank from 2.04 to 3.04 (Table~\ref{tab:sample_ablation}). All variants use
150 updates. This gain comes cheaply: one update with $B=256$ takes 11\,min\,49\,s, only about 30\% longer than the 9\,min\,02\,s required by
$B=128$, while both metrics improve clearly. The benefit is even more evident when the backward budget is reduced rather than increased. At $B=256$ and $b=64$, an update costs about 8\,min\,40\,s, roughly 4.1\% less than $B=128,\,b=128$, yet VBench Total remains comparable (83.98 vs.\ 84.09) and VLM rank improves markedly (2.54 vs.\ 2.04). Lowering the backward budget further to $b=32$ reduces the cost to about 79.6\% of the
$B=128,\,b=128$ setting, and still outperforms it on VLM rank (2.38 vs.\ 2.04). Allocating more computation to forward estimation and less to backpropagation thus matches, and can even exceed, the quality obtained with a larger backward budget at a lower total cost.

\begin{table}[t]
\centering
\EFWrapTypography
\captionof{table}{\textbf{Marginal versus joint distribution matching.}
1.3B models with ODE initialization. }
\label{tab:marginal_joint}
\begin{tabularx}{\linewidth}{*{6}{N}}
\toprule
\rowcolor{EFTableHeader}
& & \multicolumn{3}{c}{\EFMetricGroup{VBench}} & \EFMetricGroup{VLM} \\
\rowcolor{EFTableHeader}
\multirow{-2}{*}{\textbf{Refs.}} & \multirow{-2}{*}{\textbf{Match}}
& \textbf{Total} & \textbf{Qual.} & \textbf{Sem.} & \textbf{Rank} \\
\midrule
\rowcolor{EFTableOurs}
UV & Marginal & 83.65 & 84.42 & \textbf{80.57} & \textbf{2.66} \\
UV & Joint & \textbf{83.74} & \textbf{84.53} & \textbf{80.57} & 2.48 \\
\midrule
\rowcolor{EFTableOurs}
Wan & Marginal & \textbf{84.25} & \textbf{85.06} & \textbf{80.99} & 2.31 \\
Wan & Joint & 83.90 & 84.72 & 80.62 & \textbf{2.56} \\
\bottomrule
\end{tabularx}

\end{table}

\noindent\textbf{Marginal versus joint matching.}
Native MMD matches only marginal distribution.
We compare marginal matching with joint matching over videos and text representations $(x,c)$.
For joint matching, we concatenate SigLIP text features with video features
along the channel dimension. The generator is text-conditioned in
both cases. Joint matching gives no consistent improvement across the
two reference datasets (Table~\ref{tab:marginal_joint}): VLM rank
falls from 2.66 to 2.48 with UV but rises from 2.31 to 2.56 with Wan,
while VBench Total rises from 83.65 to 83.74 with UV and falls from 84.25 to 83.90 with Wan. We therefore use
video-marginal matching by default. Full component scores are in
Suppl.~\ref{app:joint-matching}.

\section{Conclusion}

We introduced Elastic Forcing, a sample-based framework for post-training
few-step autoregressive video generators. MMD matches rollout and reference
distributions in frozen representation spaces, without an online diffusion
teacher or fake-score model.
Complementary video and image encoders provide spatial and temporal
discrimination, while decoupling estimation from differentiation makes
large rollout batches practical under memory constraints. With Self
Forcing's architecture and initialization, our 1.3B model improves generation
quality at unchanged inference speed, and 14B post-training fits on one
node of eight H200 GPUs. Reference-defined supervision enables learning
visual styles, semantic concepts, and spatial priors directly from videos.
These results demonstrate a practical sample-based route to efficient
streaming generation.

\iftoggle{cvprfinal}{

\subsubsection*{Acknowledgments}
We are deeply grateful to the Krea AI team for generously providing the ODE-initialized checkpoint for Wan2.1-T2V-14B. Their support allowed us to avoid reproducing an exceptionally computationally intensive initialization procedure at this scale, saving a substantial amount of GPU time and making our large-scale experiments possible.

We also sincerely thank all volunteers who participated in our human evaluations. Their time, careful judgment, and thoughtful feedback were invaluable to the empirical assessment of our work.

Finally, we gratefully acknowledge the creators and rights holders of the Nailoong character. Its distinctive visual identity provided a meaningful case study for evaluating the acquisition of novel semantic concepts from reference videos. Nailong is used in this work solely for non-commercial academic research and evaluation, and all associated intellectual-property rights remain with their respective owners.
}{}
{\small
\bibliographystyle{ieeenat_fullname}
\bibliography{main}}

\begin{thebibliography}{111}
\providecommand{\natexlab}[1]{#1}
\providecommand{\url}[1]{\texttt{#1}}
\expandafter\ifx\csname urlstyle\endcsname\relax
  \providecommand{\doi}[1]{doi: #1}\else
  \providecommand{\doi}{doi: \begingroup \urlstyle{rm}\Url}\fi

\bibitem[ai et~al.(2025)ai, Teng, Jia, Sun, Li, Li, Tang, Han, Zhang, Zhang, Luo, Kang, Sun, Cao, Huang, Lin, Fang, Tao, Zhang, Wang, Liu, Shi, Su, Sun, Pan, Wang, Sheng, Cui, Hu, Yan, Yin, Zhang, Liu, Yin, Yang, Song, Hu, Zhang, and Li]{ai2025magi1autoregressivevideogeneration}
Sand. ai, Hansi Teng, Hongyu Jia, Lei Sun, Lingzhi Li, Maolin Li, Mingqiu Tang, Shuai Han, Tianning Zhang, W.~Q. Zhang, Weifeng Luo, Xiaoyang Kang, Yuchen Sun, Yue Cao, Yunpeng Huang, Yutong Lin, Yuxin Fang, Zewei Tao, Zheng Zhang, Zhongshu Wang, Zixun Liu, Dai Shi, Guoli Su, Hanwen Sun, Hong Pan, Jie Wang, Jiexin Sheng, Min Cui, Min Hu, Ming Yan, Shucheng Yin, Siran Zhang, Tingting Liu, Xianping Yin, Xiaoyu Yang, Xin Song, Xuan Hu, Yankai Zhang, and Yuqiao Li.
\newblock {MAGI-1: Autoregressive Video Generation at Scale}.
\newblock \emph{arXiv preprint arXiv:2505.13211}, 2025.

\bibitem[Arbel et~al.(2019)Arbel, Korba, SALIM, and Gretton]{arbel2019mmdflow}
Michael Arbel, Anna Korba, Adil SALIM, and Arthur Gretton.
\newblock {Maximum Mean Discrepancy Gradient Flow}.
\newblock In \emph{Advances in Neural Information Processing Systems}. Curran Associates, Inc., 2019.

\bibitem[Assran et~al.(2025)Assran, Bardes, Fan, Garrido, Howes, Komeili, Muckley, Rizvi, Roberts, Sinha, Zholus, Arnaud, Gejji, Martin, Robert~Hogan, Dugas, Bojanowski, Khalidov, Labatut, Massa, Szafraniec, Krishnakumar, Li, Ma, Chandar, Meier, LeCun, Rabbat, and Ballas]{assran2025vjepa2}
Mahmoud Assran, Adrien Bardes, David Fan, Quentin Garrido, Russell Howes, Mojtaba Komeili, Matthew Muckley, Ammar Rizvi, Claire Roberts, Koustuv Sinha, Artem Zholus, Sergio Arnaud, Abha Gejji, Ada Martin, Francois Robert~Hogan, Daniel Dugas, Piotr Bojanowski, Vasil Khalidov, Patrick Labatut, Francisco Massa, Marc Szafraniec, Kapil Krishnakumar, Yong Li, Xiaodong Ma, Sarath Chandar, Franziska Meier, Yann LeCun, Michael Rabbat, and Nicolas Ballas.
\newblock {V-JEPA~2: Self-Supervised Video Models Enable Understanding, Prediction and Planning}.
\newblock \emph{arXiv preprint arXiv:2506.09985}, 2025.

\bibitem[Bardes et~al.(2024)Bardes, Garrido, Ponce, Chen, Rabbat, LeCun, Assran, and Ballas]{bardes2024vjepa}
Adrien Bardes, Quentin Garrido, Jean Ponce, Xinlei Chen, Michael Rabbat, Yann LeCun, Mahmoud Assran, and Nicolas Ballas.
\newblock {Revisiting Feature Prediction for Learning Visual Representations from Video}.
\newblock \emph{arXiv preprint arXiv:2404.08471}, 2024.

\bibitem[Bellemare et~al.(2017)Bellemare, Danihelka, Dabney, Mohamed, Lakshminarayanan, Hoyer, and Munos]{bellemare2017cramer}
Marc~G. Bellemare, Ivo Danihelka, Will Dabney, Shakir Mohamed, Balaji Lakshminarayanan, Stephan Hoyer, and R{\'e}mi Munos.
\newblock {The Cramer Distance as a Solution to Biased Wasserstein Gradients}.
\newblock \emph{arXiv preprint arXiv:1705.10743}, 2017.

\bibitem[Bertasius et~al.(2021)Bertasius, Wang, and Torresani]{bertasius2021timesformer}
Gedas Bertasius, Heng Wang, and Lorenzo Torresani.
\newblock {Is Space-Time Attention All You Need for Video Understanding?}
\newblock In \emph{Proceedings of the International Conference on Machine Learning}, pages 813--824, 2021.

\bibitem[Bi{\'n}kowski et~al.(2018)Bi{\'n}kowski, Sutherland, Arbel, and Gretton]{binkowski2018demystifying}
Miko{\l}aj Bi{\'n}kowski, Danica~J. Sutherland, Michael Arbel, and Arthur Gretton.
\newblock {Demystifying MMD GANs}.
\newblock In \emph{International Conference on Learning Representations}, 2018.

\bibitem[Bruce et~al.(2024)Bruce, Dennis, Edwards, Parker-Holder, Shi, Hughes, Lai, Mavalankar, Steigerwald, Apps, Aytar, Bechtle, Behbahani, Chan, Heess, Gonzalez, Osindero, Ozair, Reed, Zhang, Zolna, Clune, Freitas, Singh, and Rockt\"{a}schel]{bruce2024genie}
Jake Bruce, Michael~D Dennis, Ashley Edwards, Jack Parker-Holder, Yuge Shi, Edward Hughes, Matthew Lai, Aditi Mavalankar, Richie Steigerwald, Chris Apps, Yusuf Aytar, Sarah Maria~Elisabeth Bechtle, Feryal Behbahani, Stephanie~C.Y. Chan, Nicolas Heess, Lucy Gonzalez, Simon Osindero, Sherjil Ozair, Scott Reed, Jingwei Zhang, Konrad Zolna, Jeff Clune, Nando~De Freitas, Satinder Singh, and Tim Rockt\"{a}schel.
\newblock {Genie: Generative Interactive Environments}.
\newblock In \emph{Proceedings of the 41st International Conference on Machine Learning}, pages 4603--4623. PMLR, 2024.

\bibitem[Carreira and Zisserman(2017)]{carreira2017i3d}
Jo{\~a}o Carreira and Andrew Zisserman.
\newblock {Quo Vadis, Action Recognition? A New Model and the Kinetics Dataset}.
\newblock In \emph{Proceedings of the IEEE Conference on Computer Vision and Pattern Recognition}, pages 6299--6308, 2017.

\bibitem[Chatalic et~al.(2022)Chatalic, Schreuder, Rosasco, and Rudi]{chatalic2022nystrom}
Antoine Chatalic, Nicolas Schreuder, Lorenzo Rosasco, and Alessandro Rudi.
\newblock {{N}ystr{ö}m Kernel Mean Embeddings}.
\newblock In \emph{Proceedings of the 39th International Conference on Machine Learning}, pages 3006--3024. PMLR, 2022.

\bibitem[Cheang et~al.(2024)Cheang, Chen, Jing, Kong, Li, Li, Liu, Wu, Xu, Yang, Zhang, and Zhu]{cheang2024gr2generativevideolanguageactionmodel}
Chi-Lam Cheang, Guangzeng Chen, Ya Jing, Tao Kong, Hang Li, Yifeng Li, Yuxiao Liu, Hongtao Wu, Jiafeng Xu, Yichu Yang, Hanbo Zhang, and Minzhao Zhu.
\newblock Gr-2: A generative video-language-action model with web-scale knowledge for robot manipulation, 2024.

\bibitem[Chen et~al.(2024)Chen, Monso, Du, Simchowitz, Tedrake, and Sitzmann]{chen2024diffusion}
Boyuan Chen, Diego~Marti Monso, Yilun Du, Max Simchowitz, Russ Tedrake, and Vincent Sitzmann.
\newblock {Diffusion Forcing: Next-token Prediction Meets Full-Sequence Diffusion}.
\newblock \emph{arXiv preprint arXiv:2407.01392}, 2024.

\bibitem[Chen et~al.(2025)Chen, Lin, Yang, Lin, Zhu, Fan, Zhang, Chen, Chen, Ma, Xiong, Wang, Pang, Kang, Xu, Jin, Liang, Song, Zhao, Xu, Qiu, Li, Fei, Li, and Zhou]{chen2025skyreelsv2infinitelengthfilmgenerative}
Guibin Chen, Dixuan Lin, Jiangping Yang, Chunze Lin, Junchen Zhu, Mingyuan Fan, Hao Zhang, Sheng Chen, Zheng Chen, Chengcheng Ma, Weiming Xiong, Wei Wang, Nuo Pang, Kang Kang, Zhiheng Xu, Yuzhe Jin, Yupeng Liang, Yubing Song, Peng Zhao, Boyuan Xu, Di Qiu, Debang Li, Zhengcong Fei, Yang Li, and Yahui Zhou.
\newblock {SkyReels-V2: Infinite-length Film Generative Model}.
\newblock \emph{arXiv preprint arXiv:2504.13074}, 2025.

\bibitem[Chen et~al.(2026{\natexlab{a}})Chen, Liu, Jia, Wang, Ling, Qu, and Gao]{chen2026dataforcing}
Siyi Chen, Shaowei Liu, Yixuan Jia, Zian Wang, Huan Ling, Qing Qu, and Jun Gao.
\newblock {Data-Forcing Distillation: Restoring Diversity and Fidelity in Few-Step Video Generation}.
\newblock \emph{arXiv preprint arXiv:2606.18478}, 2026{\natexlab{a}}.

\bibitem[Chen et~al.(2026{\natexlab{b}})Chen, Wei, Sun, Nie, Zhou, Zhang, Yang, and Chen]{chen2026context}
Shuo Chen, Cong Wei, Sun Sun, Ping Nie, Kai Zhou, Ge Zhang, Ming-Hsuan Yang, and Wenhu Chen.
\newblock {Context Forcing: Consistent Autoregressive Video Generation with Long Context}.
\newblock \emph{arXiv preprint arXiv:2602.06028}, 2026{\natexlab{b}}.

\bibitem[Chen et~al.(2016)Chen, Xu, Zhang, and Guestrin]{chen2016sublinear}
Tianqi Chen, Bing Xu, Chiyuan Zhang, and Carlos Guestrin.
\newblock {Training Deep Nets with Sublinear Memory Cost}.
\newblock \emph{arXiv preprint arXiv:1604.06174}, 2016.

\bibitem[{DeepSeek-AI}(2026)]{deepseekai2026deepseekv4}
{DeepSeek-AI}.
\newblock Deepseek-v4: Towards highly efficient million-token context intelligence, 2026.

\bibitem[Deng et~al.(2024)Deng, Pan, Diao, Luo, Cui, Lu, Shan, Qi, and Wang]{deng2024nova}
Haoge Deng, Ting Pan, Haiwen Diao, Zhengxiong Luo, Yufeng Cui, Huchuan Lu, Shiguang Shan, Yonggang Qi, and Xinlong Wang.
\newblock Autoregressive video generation without vector quantization.
\newblock \emph{arXiv preprint arXiv:2412.14169}, 2024.

\bibitem[Deng et~al.(2026)Deng, Li, Li, Du, and He]{drifting}
Mingyang Deng, He Li, Tianhong Li, Yilun Du, and Kaiming He.
\newblock {Generative Modeling via Drifting}.
\newblock \emph{arXiv preprint arXiv:2602.04770}, 2026.

\bibitem[Deshpande et~al.(2018)Deshpande, Zhang, and Schwing]{deshpande2018sliced}
Ishan Deshpande, Ziyu Zhang, and Alexander~G. Schwing.
\newblock {Generative Modeling Using the Sliced Wasserstein Distance}.
\newblock In \emph{Proceedings of the IEEE Conference on Computer Vision and Pattern Recognition (CVPR)}, pages 3483--3491, 2018.

\bibitem[Dziugaite et~al.(2015)Dziugaite, Roy, and Ghahramani]{dziugaite2015training}
Gintare~Karolina Dziugaite, Daniel~M. Roy, and Zoubin Ghahramani.
\newblock {Training generative neural networks via Maximum Mean Discrepancy optimization}.
\newblock In \emph{Proceedings of the Conference on Uncertainty in Artificial Intelligence}, 2015.

\bibitem[Feichtenhofer et~al.(2019)Feichtenhofer, Fan, Malik, and He]{feichtenhofer2019slowfast}
Christoph Feichtenhofer, Haoqi Fan, Jitendra Malik, and Kaiming He.
\newblock {SlowFast Networks for Video Recognition}.
\newblock In \emph{Proceedings of the IEEE/CVF International Conference on Computer Vision}, pages 6202--6211, 2019.

\bibitem[Feichtenhofer et~al.(2022)Feichtenhofer, Fan, Li, and He]{feichtenhofer2022mae}
Christoph Feichtenhofer, Haoqi Fan, Yanghao Li, and Kaiming He.
\newblock {Masked Autoencoders As Spatiotemporal Learners}.
\newblock In \emph{Advances in Neural Information Processing Systems}, pages 35946--35958. Curran Associates, Inc., 2022.

\bibitem[Feng et~al.(2026)Feng, Li, Zablocki, Cord, and Alahi]{irdm}
Lan Feng, Wuyang Li, Eloi Zablocki, Matthieu Cord, and Alexandre Alahi.
\newblock {Representation Distribution Matching for One-Step Visual Generation}.
\newblock \emph{arXiv preprint arXiv:2607.02375}, 2026.

\bibitem[Feydy et~al.(2019)Feydy, S{\'e}journ{\'e}, Vialard, Amari, Trouv{\'e}, and Peyr{\'e}]{feydy2019sinkhorn}
Jean Feydy, Thibault S{\'e}journ{\'e}, Fran{\c{c}}ois-Xavier Vialard, Shun-ichi Amari, Alain Trouv{\'e}, and Gabriel Peyr{\'e}.
\newblock {Interpolating between Optimal Transport and MMD using Sinkhorn Divergences}.
\newblock In \emph{Proceedings of the Twenty-Second International Conference on Artificial Intelligence and Statistics}, pages 2681--2690. PMLR, 2019.

\bibitem[Gao et~al.(2024)Gao, Shi, Zhang, Wang, Xiao, and Chen]{gao2024ca2vdm}
Kaifeng Gao, Jiaxin Shi, Hanwang Zhang, Chunping Wang, Jun Xiao, and Long Chen.
\newblock {Ca2-VDM: Efficient Autoregressive Video Diffusion Model with Causal Generation and Cache Sharing}.
\newblock \emph{arXiv preprint arXiv:2411.16375}, 2024.

\bibitem[Gao et~al.(2021)Gao, Zhang, Han, and Callan]{gao2021gradcache}
Luyu Gao, Yunyi Zhang, Jiawei Han, and Jamie Callan.
\newblock {Scaling Deep Contrastive Learning Batch Size under Memory Limited Setup}.
\newblock In \emph{Proceedings of the 6th Workshop on Representation Learning for {NLP} ({R}ep{L}4{NLP}-2021)}, pages 316--321. Association for Computational Linguistics, 2021.

\bibitem[Ge et~al.(2024)Ge, Mahapatra, Parmar, Zhu, and Huang]{10655312}
Songwei Ge, Aniruddha Mahapatra, Gaurav Parmar, Jun-Yan Zhu, and Jia-Bin Huang.
\newblock On the content bias in fréchet video distance.
\newblock In \emph{2024 IEEE/CVF Conference on Computer Vision and Pattern Recognition (CVPR)}, pages 7277--7288, 2024.

\bibitem[Genevay et~al.(2018)Genevay, Peyr{\'e}, and Cuturi]{genevay2018sinkhorn}
Aude Genevay, Gabriel Peyr{\'e}, and Marco Cuturi.
\newblock {Learning Generative Models with Sinkhorn Divergences}.
\newblock In \emph{Proceedings of the Twenty-First International Conference on Artificial Intelligence and Statistics}, pages 1608--1617. PMLR, 2018.

\bibitem[Gretton et~al.(2012)Gretton, Borgwardt, Rasch, Sch{{\"o}}lkopf, and Smola]{gretton2012kernel}
Arthur Gretton, Karsten~M. Borgwardt, Malte~J. Rasch, Bernhard Sch{{\"o}}lkopf, and Alexander Smola.
\newblock {A Kernel Two-Sample Test}.
\newblock \emph{Journal of Machine Learning Research}, 13\penalty0 (25):\penalty0 723--773, 2012.

\bibitem[Guo et~al.(2025)Guo, Yang, He, Zhao, Wei, Yang, Huang, and Lin]{guo2025resampling}
Yuwei Guo, Ceyuan Yang, Hao He, Yang Zhao, Meng Wei, Zhenheng Yang, Weilin Huang, and Dahua Lin.
\newblock {End-to-End Training for Autoregressive Video Diffusion via Self-Resampling}.
\newblock \emph{arXiv preprint arXiv:2512.15702}, 2025.

\bibitem[HaCohen et~al.(2024)HaCohen, Chiprut, Brazowski, Shalem, Moshe, Richardson, Levin, Shiran, Zabari, Gordon, Panet, Weissbuch, Kulikov, Bitterman, Melumian, and Bibi]{hacohen2024ltxvideo}
Yoav HaCohen, Nisan Chiprut, Benny Brazowski, Daniel Shalem, Dudu Moshe, Eitan Richardson, Eran Levin, Guy Shiran, Nir Zabari, Ori Gordon, Poriya Panet, Sapir Weissbuch, Victor Kulikov, Yaki Bitterman, Zeev Melumian, and Ofir Bibi.
\newblock {LTX-Video}: Realtime video latent diffusion, 2024.

\bibitem[Han et~al.(2019)Han, Xie, and Zisserman]{han2019dpc}
Tengda Han, Weidi Xie, and Andrew Zisserman.
\newblock {Video Representation Learning by Dense Predictive Coding}.
\newblock In \emph{Proceedings of the IEEE/CVF International Conference on Computer Vision (ICCV) Workshops}, 2019.

\bibitem[He et~al.(2022)He, Chen, Xie, Li, Dollár, and Girshick]{he2022mae}
Kaiming He, Xinlei Chen, Saining Xie, Yanghao Li, Piotr Dollár, and Ross Girshick.
\newblock {Masked Autoencoders Are Scalable Vision Learners}.
\newblock In \emph{Proceedings of the IEEE/CVF Conference on Computer Vision and Pattern Recognition}, 2022.

\bibitem[He et~al.(2025{\natexlab{a}})He, Jiang, Nie, Liu, Jiang, Su, Ma, Lin, Ye, Lu, Wu, Schneider, Do, Li, Jia, Zhang, Cheng, Wang, Zhou, Lin, Zhang, Zhang, Huang, and Chen]{he2025videoscore2}
Xuan He, Dongfu Jiang, Ping Nie, Minghao Liu, Zhengxuan Jiang, Mingyi Su, Wentao Ma, Junru Lin, Chun Ye, Yi Lu, Keming Wu, Benjamin Schneider, Quy~Duc Do, Zhuofeng Li, Yiming Jia, Yuxuan Zhang, Guo Cheng, Haozhe Wang, Wangchunshu Zhou, Qunshu Lin, Yuanxing Zhang, Ge Zhang, Wenhao Huang, and Wenhu Chen.
\newblock {VideoScore2: Think before You Score in Generative Video Evaluation}.
\newblock \emph{arXiv preprint arXiv:2509.22799}, 2025{\natexlab{a}}.

\bibitem[He et~al.(2025{\natexlab{b}})He, Peng, Liu, Wang, Zhang, Cui, Kang, Jiang, An, Ren, et~al.]{he2025matrix}
Xianglong He, Chunli Peng, Zexiang Liu, Boyang Wang, Yifan Zhang, Qi Cui, Fei Kang, Biao Jiang, Mengyin An, Yangyang Ren, et~al.
\newblock Matrix-game 2.0: An open-source real-time and streaming interactive world model.
\newblock \emph{arXiv preprint arXiv:2508.13009}, 2025{\natexlab{b}}.

\bibitem[Heusel et~al.(2017)Heusel, Ramsauer, Unterthiner, Nessler, and Hochreiter]{heusel2017fid}
Martin Heusel, Hubert Ramsauer, Thomas Unterthiner, Bernhard Nessler, and Sepp Hochreiter.
\newblock {GANs Trained by a Two Time-Scale Update Rule Converge to a Local Nash Equilibrium}.
\newblock In \emph{Advances in Neural Information Processing Systems}, 2017.

\bibitem[Ho et~al.(2022)Ho, Salimans, Gritsenko, Chan, Norouzi, and Fleet]{ho2022video}
Jonathan Ho, Tim Salimans, Alexey Gritsenko, William Chan, Mohammad Norouzi, and David Fleet.
\newblock {Video Diffusion Models}.
\newblock In \emph{Advances in Neural Information Processing Systems}, pages 8633--8646. Curran Associates, Inc., 2022.

\bibitem[Hong et~al.(2022)Hong, Ding, Zheng, Liu, and Tang]{hong2022cogvideo}
Wenyi Hong, Ming Ding, Wendi Zheng, Xinghan Liu, and Jie Tang.
\newblock {CogVideo: Large-scale Pretraining for Text-to-Video Generation via Transformers}.
\newblock \emph{arXiv preprint arXiv:2205.15868}, 2022.

\bibitem[Horvitz and Thompson(1952)]{horvitz1952sampling}
D.~G. Horvitz and D.~J. Thompson.
\newblock A generalization of sampling without replacement from a finite universe.
\newblock \emph{Journal of the American Statistical Association}, 47\penalty0 (260):\penalty0 663--685, 1952.

\bibitem[Huang et~al.(2025)Huang, Li, He, Zhou, and Shechtman]{huang2025selfforcing}
Xun Huang, Zhengqi Li, Guande He, Mingyuan Zhou, and Eli Shechtman.
\newblock {Self Forcing: Bridging the Train-Test Gap in Autoregressive Video Diffusion}.
\newblock \emph{arXiv preprint arXiv:2506.08009}, 2025.

\bibitem[Huang et~al.(2024)Huang, He, Yu, Zhang, Si, Jiang, Zhang, Wu, Jin, Chanpaisit, Wang, Chen, Wang, Lin, Qiao, and Liu]{huang2023vbench}
Ziqi Huang, Yinan He, Jiashuo Yu, Fan Zhang, Chenyang Si, Yuming Jiang, Yuanhan Zhang, Tianxing Wu, Qingyang Jin, Nattapol Chanpaisit, Yaohui Wang, Xinyuan Chen, Limin Wang, Dahua Lin, Yu Qiao, and Ziwei Liu.
\newblock {VBench: Comprehensive Benchmark Suite for Video Generative Models}.
\newblock In \emph{Proceedings of the IEEE/CVF Conference on Computer Vision and Pattern Recognition (CVPR)}, pages 21807--21818, 2024.

\bibitem[Jiang et~al.(2025)Jiang, Han, Mao, Zhang, Pan, and Liu]{jiang2025vace}
Zeyinzi Jiang, Zhen Han, Chaojie Mao, Jingfeng Zhang, Yulin Pan, and Yu Liu.
\newblock {VACE: All-in-One Video Creation and Editing}.
\newblock \emph{arXiv preprint arXiv:2503.07598}, 2025.

\bibitem[Jin et~al.(2024)Jin, Sun, Li, Xu, Xu, Jiang, Zhuang, Huang, Song, Mu, and Lin]{jin2024pyramidal}
Yang Jin, Zhicheng Sun, Ningyuan Li, Kun Xu, Kun Xu, Hao Jiang, Nan Zhuang, Quzhe Huang, Yang Song, Yadong Mu, and Zhouchen Lin.
\newblock Pyramidal flow matching for efficient video generative modeling.
\newblock \emph{arXiv preprint arXiv:2410.05954}, 2024.

\bibitem[Kim et~al.(2024)Kim, Kang, Choi, and Han]{kim2024fifo}
Jihwan Kim, Junoh Kang, Jinyoung Choi, and Bohyung Han.
\newblock {FIFO-Diffusion: Generating Infinite Videos from Text without Training}.
\newblock \emph{arXiv preprint arXiv:2405.11473}, 2024.

\bibitem[Kondratyuk et~al.(2023)Kondratyuk, Yu, Gu, Lezama, Huang, Schindler, Hornung, Birodkar, Yan, Chiu, Somandepalli, Akbari, Alon, Cheng, Dillon, Gupta, Hahn, Hauth, Hendon, Martinez, Minnen, Sirotenko, Sohn, Yang, Adam, Yang, Essa, Wang, Ross, Seybold, and Jiang]{kondratyuk2023videopoet}
Dan Kondratyuk, Lijun Yu, Xiuye Gu, Jos{\'e} Lezama, Jonathan Huang, Grant Schindler, Rachel Hornung, Vighnesh Birodkar, Jimmy Yan, Ming-Chang Chiu, Krishna Somandepalli, Hassan Akbari, Yair Alon, Yong Cheng, Josh Dillon, Agrim Gupta, Meera Hahn, Anja Hauth, David Hendon, Alonso Martinez, David Minnen, Mikhail Sirotenko, Kihyuk Sohn, Xuan Yang, Hartwig Adam, Ming-Hsuan Yang, Irfan Essa, Huisheng Wang, David~A. Ross, Bryan Seybold, and Lu Jiang.
\newblock {VideoPoet: A Large Language Model for Zero-Shot Video Generation}.
\newblock \emph{arXiv preprint arXiv:2312.14125}, 2023.

\bibitem[Lezama et~al.(2021)Lezama, Chen, and Qiu]{lezama2021runsort}
Jose Lezama, Wei Chen, and Qiang Qiu.
\newblock {Run-Sort-ReRun: Escaping Batch Size Limitations in Sliced Wasserstein Generative Models}.
\newblock In \emph{Proceedings of the 38th International Conference on Machine Learning}, pages 6275--6285. PMLR, 2021.

\bibitem[Li et~al.(2017)Li, Chang, Cheng, Yang, and Póczos]{li2017mmdgan}
Chun-Liang Li, Wei-Cheng Chang, Yu Cheng, Yiming Yang, and Barnabás Póczos.
\newblock {MMD GAN: Towards Deeper Understanding of Moment Matching Network}.
\newblock In \emph{Advances in Neural Information Processing Systems}, 2017.

\bibitem[Li et~al.(2026)Li, He, Chen, Roy, Tsvetkov, Zettlemoyer, Ghazvininejad, and Han]{li2026isovrbenchvideorewardmodeling}
Shuyue~Stella Li, Jacqueline He, Jiahui Chen, Sreya~Dutta Roy, Yulia Tsvetkov, Luke Zettlemoyer, Marjan Ghazvininejad, and Xiaochuang Han.
\newblock {IsoVRBench: Video Reward Modeling Without Human Feedback}, 2026.

\bibitem[Li et~al.(2025)Li, Pan, Luan, Gao, and Alahi]{li2025stableinfinity}
Wuyang Li, Wentao Pan, Po-Chien Luan, Yang Gao, and Alexandre Alahi.
\newblock {Stable Video Infinity: Infinite-Length Video Generation with Error Recycling}.
\newblock \emph{arXiv preprint arXiv:2510.09212}, 2025.

\bibitem[Li et~al.(2015)Li, Swersky, and Zemel]{li2015generative}
Yujia Li, Kevin Swersky, and Rich Zemel.
\newblock {Generative Moment Matching Networks}.
\newblock In \emph{Proceedings of the 32nd International Conference on Machine Learning}, pages 1718--1727. PMLR, 2015.

\bibitem[Lin et~al.(2025{\natexlab{a}})Lin, Xia, Ren, Yang, Xiao, and Jiang]{lin2025apt}
Shanchuan Lin, Xin Xia, Yuxi Ren, Ceyuan Yang, Xuefeng Xiao, and Lu Jiang.
\newblock {Diffusion Adversarial Post-Training for One-Step Video Generation}.
\newblock \emph{arXiv preprint arXiv:2501.08316}, 2025{\natexlab{a}}.

\bibitem[Lin et~al.(2025{\natexlab{b}})Lin, Yang, He, Jiang, Ren, Xia, Zhao, Xiao, and Jiang]{lin2025aapt}
Shanchuan Lin, Ceyuan Yang, Hao He, Jianwen Jiang, Yuxi Ren, Xin Xia, Yang Zhao, Xuefeng Xiao, and Lu Jiang.
\newblock {Autoregressive Adversarial Post-Training for Real-Time Interactive Video Generation}.
\newblock In \emph{Advances in Neural Information Processing Systems}, pages 41061--41086. Curran Associates, Inc., 2025{\natexlab{b}}.

\bibitem[Liu et~al.(2026{\natexlab{a}})Liu, Wang, Gao, He, Ma, Wan, Zhang, Wei, and Chen]{liu2026opsdv}
Hongyu Liu, Chun Wang, Feng Gao, Xuanhua He, Yue Ma, Ziyu Wan, Yong Zhang, Xiaoming Wei, and Qifeng Chen.
\newblock {OPSD-V: On-Policy Self-Distillation for Post-Training Few-Step Autoregressive Video Generators}.
\newblock \emph{arXiv preprint arXiv:2607.08766}, 2026{\natexlab{a}}.

\bibitem[Liu et~al.(2026{\natexlab{b}})Liu, Liu, Mei, Wen, Yang, and Liu]{liu2026diagonal}
Jinxiu Liu, Xuanming Liu, Kangfu Mei, Yandong Wen, Ming-Hsuan Yang, and Weiyang Liu.
\newblock {Streaming Autoregressive Video Generation via Diagonal Distillation}.
\newblock \emph{arXiv preprint arXiv:2603.09488}, 2026{\natexlab{b}}.

\bibitem[Liu et~al.(2025)Liu, Hu, Xu, Shan, and Lu]{liu2025rolling}
Kunhao Liu, Wenbo Hu, Jiale Xu, Ying Shan, and Shijian Lu.
\newblock {Rolling Forcing: Autoregressive Long Video Diffusion in Real Time}.
\newblock \emph{arXiv preprint arXiv:2509.25161}, 2025.

\bibitem[Liu et~al.(2026{\natexlab{c}})Liu, Zhang, Cui, and Liu]{liu2026elasticttt}
Yueyi Liu, Chi Zhang, Sen Cui, and Miao Liu.
\newblock Elasticttt: Prior-preserving test-time tuning for video editing.
\newblock \emph{arXiv preprint arXiv:2607.21529}, 2026{\natexlab{c}}.

\bibitem[Liu et~al.(2022)Liu, Mao, Wu, Feichtenhofer, Darrell, and Xie]{liu2022convnext}
Zhuang Liu, Hanzi Mao, Chao-Yuan Wu, Christoph Feichtenhofer, Trevor Darrell, and Saining Xie.
\newblock {A ConvNet for the 2020s}.
\newblock In \emph{Proceedings of the IEEE/CVF Conference on Computer Vision and Pattern Recognition (CVPR)}, pages 11976--11986, 2022.

\bibitem[Liutkus et~al.(2019)Liutkus, {\c{S}}im{\c{s}}ekli, Majewski, Durmus, and St{\"o}ter]{liutkus2019slicedflows}
Antoine Liutkus, Umut {\c{S}}im{\c{s}}ekli, Szymon Majewski, Alain Durmus, and Fabian-Robert St{\"o}ter.
\newblock {Sliced-Wasserstein Flows: Nonparametric Generative Modeling via Optimal Transport and Diffusions}.
\newblock In \emph{Proceedings of the International Conference on Machine Learning}, pages 4104--4113, 2019.

\bibitem[Lu et~al.(2025)Lu, Zeng, Li, Ouyang, Wang, Cheng, Zhu, Cao, Zhang, Zhu, Shen, and Zhang]{lu2025reward}
Yunhong Lu, Yanhong Zeng, Haobo Li, Hao Ouyang, Qiuyu Wang, Ka~Leong Cheng, Jiapeng Zhu, Hengyuan Cao, Zhipeng Zhang, Xing Zhu, Yujun Shen, and Min Zhang.
\newblock {Reward Forcing: Efficient Streaming Video Generation with Rewarded Distribution Matching Distillation}.
\newblock \emph{arXiv preprint arXiv:2512.04678}, 2025.

\bibitem[Meng et~al.(2026)Meng, Ge, Huang, Zhou, and Jiang]{meng2026virdm}
Zichong Meng, Chongjian Ge, Chun-Hao~P. Huang, Yang Zhou, and Huaizu Jiang.
\newblock {ViRDM: Taming Representation Distribution Matching for Few-Step Causal Video Generation}.
\newblock \emph{arXiv preprint arXiv:2609.28923}, 2026.

\bibitem[Millon(2025{\natexlab{a}})]{krea_realtime_14b}
Erwann Millon.
\newblock Krea realtime 14b: Real-time video generation, 2025{\natexlab{a}}.

\bibitem[Millon(2025{\natexlab{b}})]{millon2025krea}
Erwann Millon.
\newblock {Krea Realtime 14B}: Real-time, long-form {AI} video generation.
\newblock Krea technical blog, 2025{\natexlab{b}}.

\bibitem[Muandet et~al.(2017)Muandet, Fukumizu, Sriperumbudur, and Schölkopf]{muandet2017kernel}
Krikamol Muandet, Kenji Fukumizu, Bharath Sriperumbudur, and Bernhard Schölkopf.
\newblock {Kernel Mean Embedding of Distributions: A Review and Beyond}.
\newblock \emph{Foundations and Trends in Machine Learning}, 10\penalty0 (1--2):\penalty0 1--141, 2017.

\bibitem[Nan et~al.(2025)Nan, Xie, Zhou, Fan, Yang, Chen, Li, Yang, and Tai]{nan2025openvid}
Kepan Nan, Rui Xie, Penghao Zhou, Tiehan Fan, Zhenheng Yang, Zhijie Chen, Xiang Li, Jian Yang, and Ying Tai.
\newblock {OpenVid-1M: A Large-Scale High-Quality Dataset for Text-to-video Generation}.
\newblock In \emph{International Conference on Learning Representations}, 2025.

\bibitem[Oquab et~al.(2024)Oquab, Darcet, Moutakanni, Vo, Szafraniec, Khalidov, Fernandez, Haziza, Massa, El-Nouby, Assran, Ballas, Galuba, Howes, Huang, Li, Misra, Rabbat, Sharma, Synnaeve, Xu, J{\'e}gou, Mairal, Labatut, Joulin, and Bojanowski]{oquab2023dinov2}
Maxime Oquab, Timoth{\'e}e Darcet, Th{\'e}o Moutakanni, Huy Vo, Marc Szafraniec, Vasil Khalidov, Pierre Fernandez, Daniel Haziza, Francisco Massa, Alaaeldin El-Nouby, Mahmoud Assran, Nicolas Ballas, Wojciech Galuba, Russell Howes, Po-Yao Huang, Shang-Wen Li, Ishan Misra, Michael Rabbat, Vasu Sharma, Gabriel Synnaeve, Hu Xu, Herv{\'e} J{\'e}gou, Julien Mairal, Patrick Labatut, Armand Joulin, and Piotr Bojanowski.
\newblock {DINOv2: Learning Robust Visual Features without Supervision}.
\newblock \emph{Transactions on Machine Learning Research}, 2024.

\bibitem[Patrick et~al.(2021)Patrick, Campbell, Asano, Misra, Metze, Feichtenhofer, Vedaldi, and Henriques]{patrick2021motionformer}
Mandela Patrick, Dylan Campbell, Yuki~M. Asano, Ishan Misra, Florian Metze, Christoph Feichtenhofer, Andrea Vedaldi, and João~F. Henriques.
\newblock {Keeping Your Eye on the Ball: Trajectory Attention in Video Transformers}.
\newblock In \emph{Advances in Neural Information Processing Systems}, 2021.

\bibitem[Qian et~al.(2021)Qian, Meng, Gong, Yang, Wang, Belongie, and Cui]{qian2021cvrl}
Rui Qian, Tianjian Meng, Boqing Gong, Ming-Hsuan Yang, Huisheng Wang, Serge Belongie, and Yin Cui.
\newblock {Spatiotemporal Contrastive Video Representation Learning}.
\newblock In \emph{Proceedings of the IEEE/CVF Conference on Computer Vision and Pattern Recognition (CVPR)}, pages 6964--6974, 2021.

\bibitem[Radford et~al.(2021)Radford, Kim, Hallacy, Ramesh, Goh, Agarwal, Sastry, Askell, Mishkin, Clark, Krueger, and Sutskever]{radford2021clip}
Alec Radford, Jong~Wook Kim, Chris Hallacy, Aditya Ramesh, Gabriel Goh, Sandhini Agarwal, Girish Sastry, Amanda Askell, Pamela Mishkin, Jack Clark, Gretchen Krueger, and Ilya Sutskever.
\newblock {Learning Transferable Visual Models From Natural Language Supervision}.
\newblock In \emph{Proceedings of the 38th International Conference on Machine Learning}, pages 8748--8763. PMLR, 2021.

\bibitem[Rajbhandari et~al.(2020)Rajbhandari, Rasley, Ruwase, and He]{rajbhandari2020zero}
Samyam Rajbhandari, Jeff Rasley, Olatunji Ruwase, and Yuxiong He.
\newblock {ZeRO: Memory optimizations Toward Training Trillion Parameter Models}.
\newblock In \emph{SC20: International Conference for High Performance Computing, Networking, Storage and Analysis}, pages 1--16. IEEE, 2020.

\bibitem[Siméoni et~al.(2025)Siméoni, Vo, Seitzer, Baldassarre, Oquab, Jose, Khalidov, Szafraniec, Yi, Ramamonjisoa, Massa, Haziza, Wehrstedt, Wang, Darcet, Moutakanni, Sentana, Roberts, Vedaldi, Tolan, Brandt, Couprie, Mairal, Jégou, Labatut, and Bojanowski]{simeoni2025dinov3}
Oriane Siméoni, Huy~V. Vo, Maximilian Seitzer, Federico Baldassarre, Maxime Oquab, Cijo Jose, Vasil Khalidov, Marc Szafraniec, Seungeun Yi, Michaël Ramamonjisoa, Francisco Massa, Daniel Haziza, Luca Wehrstedt, Jianyuan Wang, Timothée Darcet, Théo Moutakanni, Leonel Sentana, Claire Roberts, Andrea Vedaldi, Jamie Tolan, John Brandt, Camille Couprie, Julien Mairal, Hervé Jégou, Patrick Labatut, and Piotr Bojanowski.
\newblock {DINOv3}.
\newblock \emph{arXiv preprint arXiv:2508.10104}, 2025.

\bibitem[Song et~al.(2025)Song, Chen, Simchowitz, Du, Tedrake, and Sitzmann]{song2025history}
Kiwhan Song, Boyuan Chen, Max Simchowitz, Yilun Du, Russ Tedrake, and Vincent Sitzmann.
\newblock {History-Guided Video Diffusion}.
\newblock \emph{arXiv preprint arXiv:2502.06764}, 2025.

\bibitem[Sriperumbudur et~al.(2010)Sriperumbudur, Gretton, Fukumizu, Sch{{\"o}}lkopf, and Lanckriet]{sriperumbudur2010hilbert}
Bharath~K. Sriperumbudur, Arthur Gretton, Kenji Fukumizu, Bernhard Sch{{\"o}}lkopf, and Gert~R.G. Lanckriet.
\newblock {Hilbert Space Embeddings and Metrics on Probability Measures}.
\newblock \emph{Journal of Machine Learning Research}, 11\penalty0 (50):\penalty0 1517--1561, 2010.

\bibitem[Starodubcev et~al.(2026)Starodubcev, Drobyshevskiy, Kuznedelev, Babenko, and Baranchuk]{starodubcev2026scalewise}
Nikita Starodubcev, Ilya Drobyshevskiy, Denis Kuznedelev, Artem Babenko, and Dmitry Baranchuk.
\newblock {Scale-wise Distillation of Diffusion Models}.
\newblock In \emph{The Fourteenth International Conference on Learning Representations}, 2026.

\bibitem[Szegedy et~al.(2016)Szegedy, Vanhoucke, Ioffe, Shlens, and Wojna]{szegedy2016inception}
Christian Szegedy, Vincent Vanhoucke, Sergey Ioffe, Jonathon Shlens, and Zbigniew Wojna.
\newblock {Rethinking the Inception Architecture for Computer Vision}.
\newblock In \emph{Proceedings of the IEEE Conference on Computer Vision and Pattern Recognition}, 2016.

\bibitem[Tong et~al.(2022)Tong, Song, Wang, and Wang]{tong2022videomae}
Zhan Tong, Yibing Song, Jue Wang, and Limin Wang.
\newblock {VideoMAE: Masked Autoencoders are Data-Efficient Learners for Self-Supervised Video Pre-Training}.
\newblock In \emph{Advances in Neural Information Processing Systems}, 2022.

\bibitem[Tschannen et~al.(2025)Tschannen, Gritsenko, Wang, Naeem, Alabdulmohsin, Parthasarathy, Evans, Beyer, Xia, Mustafa, Hénaff, Harmsen, Steiner, and Zhai]{tschannen2025siglip2}
Michael Tschannen, Alexey Gritsenko, Xiao Wang, Muhammad~Ferjad Naeem, Ibrahim Alabdulmohsin, Nikhil Parthasarathy, Talfan Evans, Lucas Beyer, Ye Xia, Basil Mustafa, Olivier Hénaff, Jeremiah Harmsen, Andreas Steiner, and Xiaohua Zhai.
\newblock {SigLIP 2: Multilingual Vision-Language Encoders with Improved Semantic Understanding, Localization, and Dense Features}.
\newblock \emph{arXiv preprint arXiv:2502.14786}, 2025.

\bibitem[Unterthiner et~al.(2018)Unterthiner, van Steenkiste, Kurach, Marinier, Michalski, and Gelly]{unterthiner2018fvd}
Thomas Unterthiner, Sjoerd van Steenkiste, Karol Kurach, Raphael Marinier, Marcin Michalski, and Sylvain Gelly.
\newblock {Towards Accurate Generative Models of Video: A New Metric \& Challenges}.
\newblock \emph{arXiv preprint arXiv:1812.01717}, 2018.

\bibitem[Valevski et~al.(2025)Valevski, Leviathan, Arar, and Fruchter]{ICLR2025_b71ecea2}
Dani Valevski, Yaniv Leviathan, Moab Arar, and Shlomi Fruchter.
\newblock Diffusion models are real-time game engines.
\newblock In \emph{International Conference on Learning Representations}, pages 73754--73776, 2025.

\bibitem[Wan et~al.(2025)Wan, Wang, Ai, Wen, Mao, Xie, Chen, Yu, Zhao, Yang, Zeng, Wang, Zhang, Zhou, Wang, Chen, Zhu, Zhao, Yan, Huang, Feng, Zhang, Li, Wu, Chu, Feng, Zhang, Sun, Fang, Wang, Gui, Weng, Shen, Lin, Wang, Wang, Zhou, Wang, Shen, Yu, Shi, Huang, Xu, Kou, Lv, Li, Liu, Wang, Zhang, Huang, Li, Wu, Liu, Pan, Zheng, Hong, Shi, Feng, Jiang, Han, Wu, and Liu]{wan2025}
Team Wan, Ang Wang, Baole Ai, Bin Wen, Chaojie Mao, Chen-Wei Xie, Di Chen, Feiwu Yu, Haiming Zhao, Jianxiao Yang, Jianyuan Zeng, Jiayu Wang, Jingfeng Zhang, Jingren Zhou, Jinkai Wang, Jixuan Chen, Kai Zhu, Kang Zhao, Keyu Yan, Lianghua Huang, Mengyang Feng, Ningyi Zhang, Pandeng Li, Pingyu Wu, Ruihang Chu, Ruili Feng, Shiwei Zhang, Siyang Sun, Tao Fang, Tianxing Wang, Tianyi Gui, Tingyu Weng, Tong Shen, Wei Lin, Wei Wang, Wei Wang, Wenmeng Zhou, Wente Wang, Wenting Shen, Wenyuan Yu, Xianzhong Shi, Xiaoming Huang, Xin Xu, Yan Kou, Yangyu Lv, Yifei Li, Yijing Liu, Yiming Wang, Yingya Zhang, Yitong Huang, Yong Li, You Wu, Yu Liu, Yulin Pan, Yun Zheng, Yuntao Hong, Yupeng Shi, Yutong Feng, Zeyinzi Jiang, Zhen Han, Zhi-Fan Wu, and Ziyu Liu.
\newblock {Wan: Open and Advanced Large-Scale Video Generative Models}.
\newblock \emph{arXiv preprint arXiv:2503.20314}, 2025.

\bibitem[Wang et~al.(2023)Wang, Huang, Zhao, Tong, He, Wang, Wang, and Qiao]{wang2023videomaev2}
Limin Wang, Bingkun Huang, Zhiyu Zhao, Zhan Tong, Yinan He, Yi Wang, Yali Wang, and Yu Qiao.
\newblock {VideoMAE V2: Scaling Video Masked Autoencoders With Dual Masking}.
\newblock In \emph{Proceedings of the IEEE/CVF Conference on Computer Vision and Pattern Recognition (CVPR)}, pages 14549--14560, 2023.

\bibitem[Wang and Yang(2024)]{wang2024vidprom}
Wenhao Wang and Yi Yang.
\newblock Vidprom: A million-scale real prompt-gallery dataset for text-to-video diffusion models.
\newblock \emph{Advances in Neural Information Processing Systems}, 37:\penalty0 65618--65642, 2024.

\bibitem[Wang et~al.(2024)Wang, Li, Li, Yu, He, Wang, Chen, Pei, Yan, Zheng, Xu, Wang, Shi, Jiang, Li, Zhang, Huang, Qiao, Wang, and Wang]{wang2024internvideo2}
Yi Wang, Kunchang Li, Xinhao Li, Jiashuo Yu, Yinan He, Chenting Wang, Guo Chen, Baoqi Pei, Ziang Yan, Rongkun Zheng, Jilan Xu, Zun Wang, Yansong Shi, Tianxiang Jiang, Songze Li, Hongjie Zhang, Yifei Huang, Yu Qiao, Yali Wang, and Limin Wang.
\newblock {InternVideo2: Scaling Foundation Models for Multimodal Video Understanding}.
\newblock \emph{arXiv preprint arXiv:2403.15377}, 2024.

\bibitem[Wei et~al.(2022)Wei, Fan, Xie, Wu, Yuille, and Feichtenhofer]{wei2022maskfeat}
Chen Wei, Haoqi Fan, Saining Xie, Chao-Yuan Wu, Alan Yuille, and Christoph Feichtenhofer.
\newblock {Masked Feature Prediction for Self-Supervised Visual Pre-Training}.
\newblock In \emph{Proceedings of the IEEE/CVF Conference on Computer Vision and Pattern Recognition (CVPR)}, pages 14668--14678, 2022.

\bibitem[Williams and Seeger(2000)]{williams2000nystrom}
Christopher Williams and Matthias Seeger.
\newblock {Using the Nystr\"{o}m Method to Speed Up Kernel Machines}.
\newblock In \emph{Advances in Neural Information Processing Systems}. MIT Press, 2000.

\bibitem[Wu et~al.(2023)Wu, Jing, Cheang, Chen, Xu, Li, Liu, Li, and Kong]{wu2023unleashinglargescalevideogenerative}
Hongtao Wu, Ya Jing, Chilam Cheang, Guangzeng Chen, Jiafeng Xu, Xinghang Li, Minghuan Liu, Hang Li, and Tao Kong.
\newblock Unleashing large-scale video generative pre-training for visual robot manipulation, 2023.

\bibitem[Wu et~al.(2024)Wu, Yin, Feng, He, Li, Hao, and Long]{NEURIPS2024_7dbb5bfa}
Jialong Wu, Shaofeng Yin, Ningya Feng, Xu He, Dong Li, Jianye Hao, and Mingsheng Long.
\newblock {iVideoGPT: Interactive VideoGPTs are Scalable World Models}.
\newblock In \emph{Advances in Neural Information Processing Systems}, pages 68082--68119. Curran Associates, Inc., 2024.

\bibitem[Xu et~al.(2021)Xu, Ghosh, Huang, Okhonko, Aghajanyan, Metze, Zettlemoyer, and Feichtenhofer]{xu2021videoclip}
Hu Xu, Gargi Ghosh, Po-Yao Huang, Dmytro Okhonko, Armen Aghajanyan, Florian Metze, Luke Zettlemoyer, and Christoph Feichtenhofer.
\newblock {{V}ideo{CLIP}: Contrastive Pre-training for Zero-shot Video-Text Understanding}.
\newblock In \emph{Proceedings of the 2021 Conference on Empirical Methods in Natural Language Processing}, pages 6787--6800. Association for Computational Linguistics, 2021.

\bibitem[Xue et~al.(2026)Xue, Feng, Lin, Lin, Zeng, Zhang, Agrawala, Yan, and Pan]{xue2026ring}
Bowen Xue, Brandon~Y. Feng, Chenguo Lin, Yuchen Lin, Yujia Zeng, Lvmin Zhang, Maneesh Agrawala, Honglei Yan, and Panwang Pan.
\newblock {Ring Forcing: Towards Precise Long-Term Memory for Autoregressive Video Diffusion}.
\newblock \emph{arXiv preprint arXiv:2608.26794}, 2026.

\bibitem[Xue et~al.(2025)Xue, Zhang, Hu, He, Chen, Cai, Wang, Wang, Liu, Li, and Tao]{xue2025ultravideo}
Zhucun Xue, Jiangning Zhang, Teng Hu, Haoyang He, Yinan Chen, Yuxuan Cai, Yabiao Wang, Chengjie Wang, Yong Liu, Xiangtai Li, and Dacheng Tao.
\newblock {UltraVideo: High-Quality UHD Video Dataset with Comprehensive Captions}.
\newblock \emph{arXiv preprint arXiv:2506.13691}, 2025.

\bibitem[Yan et~al.(2021)Yan, Zhang, Abbeel, and Srinivas]{yan2021videogpt}
Wilson Yan, Yunzhi Zhang, Pieter Abbeel, and Aravind Srinivas.
\newblock Videogpt: Video generation using vq-vae and transformers.
\newblock \emph{arXiv preprint arXiv:2104.10157}, 2021.

\bibitem[Yang et~al.(2026)Yang, Geng, Ju, Tian, and Wang]{yang2026frechet}
Jiawei Yang, Zhengyang Geng, Xuan Ju, Yonglong Tian, and Yue Wang.
\newblock {Representation Fr\'echet Loss for Visual Generation}.
\newblock \emph{arXiv preprint arXiv:2604.28190}, 2026.

\bibitem[Yang et~al.(2025)Yang, Huang, Chu, Xiao, Zhao, Wang, Li, Xie, Chen, Lu, Han, and Chen]{yang2025longlive}
Shuai Yang, Wei Huang, Ruihang Chu, Yicheng Xiao, Yuyang Zhao, Xianbang Wang, Muyang Li, Enze Xie, Yingcong Chen, Yao Lu, Song Han, and Yukang Chen.
\newblock {LongLive: Real-time Interactive Long Video Generation}.
\newblock \emph{arXiv preprint arXiv:2509.22622}, 2025.

\bibitem[Yin et~al.(2024{\natexlab{a}})Yin, Gharbi, Park, Zhang, Shechtman, Durand, and Freeman]{yin2024improved}
Tianwei Yin, Micha\"{e}l Gharbi, Taesung Park, Richard Zhang, Eli Shechtman, Fr\'{e}do Durand, and William~T. Freeman.
\newblock {Improved Distribution Matching Distillation for Fast Image Synthesis}.
\newblock In \emph{Advances in Neural Information Processing Systems}, pages 47455--47487. Curran Associates, Inc., 2024{\natexlab{a}}.

\bibitem[Yin et~al.(2024{\natexlab{b}})Yin, Gharbi, Zhang, Shechtman, Durand, Freeman, and Park]{yin2024onestep}
Tianwei Yin, Micha\"el Gharbi, Richard Zhang, Eli Shechtman, Fr\'edo Durand, William~T. Freeman, and Taesung Park.
\newblock {One-step Diffusion with Distribution Matching Distillation}.
\newblock In \emph{Proceedings of the IEEE/CVF Conference on Computer Vision and Pattern Recognition (CVPR)}, pages 6613--6623, 2024{\natexlab{b}}.

\bibitem[Yin et~al.(2025)Yin, Zhang, Zhang, Freeman, Durand, Shechtman, and Huang]{yin2025causvid}
Tianwei Yin, Qiang Zhang, Richard Zhang, William~T. Freeman, Fredo Durand, Eli Shechtman, and Xun Huang.
\newblock {From Slow Bidirectional to Fast Autoregressive Video Diffusion Models}.
\newblock In \emph{Proceedings of the IEEE/CVF Conference on Computer Vision and Pattern Recognition (CVPR)}, pages 22963--22974, 2025.

\bibitem[Yu et~al.(2025)Yu, Kwak, Jang, Jeong, Huang, Shin, and Xie]{yu2024repa}
Sihyun Yu, Sangkyung Kwak, Huiwon Jang, Jongheon Jeong, Jonathan Huang, Jinwoo Shin, and Saining Xie.
\newblock {Representation Alignment for Generation: Training Diffusion Transformers Is Easier Than You Think}.
\newblock In \emph{International Conference on Learning Representations}, pages 87400--87442, 2025.

\bibitem[Yuan et~al.(2026)Yuan, Yin, Li, Huang, Yang, and Yuan]{yuan2026helios}
Shenghai Yuan, Yuanyang Yin, Zongjian Li, Xinwei Huang, Xiao Yang, and Li Yuan.
\newblock {Helios: Real Real-Time Long Video Generation Model}.
\newblock \emph{arXiv preprint arXiv:2603.04379}, 2026.

\bibitem[Zhai et~al.(2023)Zhai, Mustafa, Kolesnikov, and Beyer]{zhai2023siglip}
Xiaohua Zhai, Basil Mustafa, Alexander Kolesnikov, and Lucas Beyer.
\newblock {Sigmoid Loss for Language Image Pre-Training}.
\newblock In \emph{Proceedings of the IEEE/CVF International Conference on Computer Vision (ICCV)}, pages 11975--11986, 2023.

\bibitem[Zhang et~al.(2025{\natexlab{a}})Zhang, Chen, Zheng, and Zhu]{zhang2025voicebridge}
Chi Zhang, Zehua Chen, Kaiwen Zheng, and Jun Zhu.
\newblock Voicebridge: General speech restoration with one-step latent bridge models.
\newblock \emph{arXiv preprint arXiv:2509.25275}, 2025{\natexlab{a}}.

\bibitem[Zhang et~al.(2026{\natexlab{a}})Zhang, Haoyang, Liu, An, Zhou, Li, Lu, Zhang, Wang, Wu, Cui, and Liu]{zhang2026distributional}
Chi Zhang, Shi Haoyang, Yueyi Liu, Ruichuan An, Junkang Zhou, Chang Li, Xiuyuan Lu, Yichi Zhang, Bo Wang, Yuhang Wu, Sen Cui, and Miao Liu.
\newblock {Unifying Distributional Training for One-Step Visual Generation}.
\newblock \emph{arXiv preprint arXiv:2609.35763}, 2026{\natexlab{a}}.

\bibitem[Zhang et~al.(2026{\natexlab{b}})Zhang, Shi, Liu, Yan, Yin, Wu, and Liu]{zhang2026interacvid}
Chi Zhang, Haoyang Shi, Yueyi Liu, Zhaokun Yan, Yishu Yin, Yuhang Wu, and Miao Liu.
\newblock Interacvid: Building a real interactive audio-visual response dataset from live-chat videos.
\newblock \emph{arXiv preprint arXiv:2608.01157}, 2026{\natexlab{b}}.

\bibitem[Zhang et~al.(2008)Zhang, Tsang, and Kwok]{zhang2008nystrom}
Kai Zhang, Ivor~W. Tsang, and James~T. Kwok.
\newblock Improved {Nystr\"om} low-rank approximation and error analysis.
\newblock In \emph{Proceedings of the 25th International Conference on Machine Learning}, pages 1232--1239, 2008.

\bibitem[Zhang et~al.(2025{\natexlab{b}})Zhang, Cai, Li, Wetzstein, and Agrawala]{zhang2025framepack}
Lvmin Zhang, Shengqu Cai, Muyang Li, Gordon Wetzstein, and Maneesh Agrawala.
\newblock {Frame Context Packing and Drift Prevention in Next-Frame-Prediction Video Diffusion Models}.
\newblock \emph{arXiv preprint arXiv:2504.12626}, 2025{\natexlab{b}}.

\bibitem[Zhang et~al.(2018)Zhang, Isola, Efros, Shechtman, and Wang]{zhang2018unreasonable}
Richard Zhang, Phillip Isola, Alexei~A. Efros, Eli Shechtman, and Oliver Wang.
\newblock {The Unreasonable Effectiveness of Deep Features as a Perceptual Metric}.
\newblock In \emph{Proceedings of the IEEE Conference on Computer Vision and Pattern Recognition (CVPR)}, 2018.

\bibitem[Zhao et~al.(2024)Zhao, Gundavarapu, Yuan, Zhou, Yan, Sun, Friedman, Qian, Weyand, Zhao, Hornung, Schroff, Yang, Ross, Wang, Adam, Sirotenko, Liu, and Gong]{zhao2024videoprism}
Long Zhao, Nitesh~Bharadwaj Gundavarapu, Liangzhe Yuan, Hao Zhou, Shen Yan, Jennifer~J. Sun, Luke Friedman, Rui Qian, Tobias Weyand, Yue Zhao, Rachel Hornung, Florian Schroff, Ming-Hsuan Yang, David~A Ross, Huisheng Wang, Hartwig Adam, Mikhail Sirotenko, Ting Liu, and Boqing Gong.
\newblock {{V}ideo{P}rism: A Foundational Visual Encoder for Video Understanding}.
\newblock In \emph{Proceedings of the 41st International Conference on Machine Learning}, pages 60785--60811. PMLR, 2024.

\bibitem[Zheng et~al.(2025)Zheng, Huang, Liu, Zou, He, Zhang, Gu, Zhang, He, Zheng, Qiao, and Liu]{zheng2025vbench2}
Dian Zheng, Ziqi Huang, Hongbo Liu, Kai Zou, Yinan He, Fan Zhang, Lulu Gu, Yuanhan Zhang, Jingwen He, Wei-Shi Zheng, Yu Qiao, and Ziwei Liu.
\newblock {VBench-2.0: Advancing Video Generation Benchmark Suite for Intrinsic Faithfulness}.
\newblock \emph{arXiv preprint arXiv:2503.21755}, 2025.

\bibitem[Zhou et~al.(2024)Zhou, Zheng, Wang, Yin, and Huang]{zhou2024sid}
Mingyuan Zhou, Huangjie Zheng, Zhendong Wang, Mingzhang Yin, and Hai Huang.
\newblock {Score identity Distillation: Exponentially Fast Distillation of Pretrained Diffusion Models for One-Step Generation}.
\newblock In \emph{Proceedings of the 41st International Conference on Machine Learning}, pages 62307--62331. PMLR, 2024.

\bibitem[Zhou et~al.(2025)Zhou, Zheng, Gu, Wang, and Huang]{zhou2025sida}
Mingyuan Zhou, Huangjie Zheng, Yi Gu, Zhendong Wang, and Hai Huang.
\newblock {Adversarial Score identity Distillation: Rapidly Surpassing the Teacher in One Step}.
\newblock In \emph{International Conference on Learning Representations}, pages 83214--83247, 2025.

\bibitem[Zhu et~al.(2026)Zhu, Zhao, He, Su, Li, and Zhu]{zhu2026causal}
Hongzhou Zhu, Min Zhao, Guande He, Hang Su, Chongxuan Li, and Jun Zhu.
\newblock Causal forcing: Autoregressive diffusion distillation done right for high-quality real-time interactive video generation.
\newblock \emph{arXiv preprint arXiv:2602.02214}, 2026.

\bibitem[Zhuang et~al.(2026)Zhuang, Zhang, Bian, Li, Luo, Liu, Jin, Zhang, He, Zhang, et~al.]{zhuang2026self}
Junhao Zhuang, Shiyi Zhang, Yuxuan Bian, Yaowei Li, Yawen Luo, Yijun Liu, Weiyang Jin, Songchun Zhang, Xianglong He, Xuying Zhang, et~al.
\newblock Self gradient forcing: Native long video extrapolation.
\newblock \emph{arXiv preprint arXiv:2607.20368}, 2026.

\end{thebibliography}

\ifincludesupplement
\beginsupplement
\section*{LLM Usage Statement}

Generative AI tools were used for auxiliary tasks and automated video evaluation.
DeepSeek-V4-Flash~\citep{deepseekai2026deepseekv4} was used to expand prompts
drawn from VidProM for constructing the reference video dataset described
in Section~\ref{sec:experiments}. ChatGPT assisted with language editing and drafting portions of
the manuscript, while OpenAI Codex assisted in generating parts of the project
code. GPT-6 was also used to directly score generated videos and rank anonymous
groups of videos under the VLM evaluation protocols described in
Section~\ref{sec:experimental-setup} and Suppl.~\ref{sec:vlm-prompt-design}.
All AI-assisted content, including text and code, was carefully reviewed,
verified, and revised by the authors. These tools were not used to independently
formulate the research hypotheses, derive the proposed method, design the
experiments, or draw scientific conclusions. The authors take full
responsibility for the content and conclusions of this paper.
\section*{Supplementary Video}

\noindent Video examples and demonstrations are provided in the supplementary
video, with segments beginning at 00:00, 03:17, and 04:51.\par

\startcontents[appendix]
\section*{Suppl. Contents}
\begingroup
\hypersetup{linkcolor=meowDarkGray}
\printcontents[appendix]{app}{1}[2]{}
\endgroup
\par\medskip
\Needspace{8\baselineskip}

\section{Related Work}
\label{app:related-work}

\subsection{Autoregressive video generation}

Autoregressive generators predict ordered visual tokens or continuous
latent chunks~\citep{yan2021videogpt,kondratyuk2023videopoet,NEURIPS2024_7dbb5bfa,deng2024nova},
including action-conditioned world modeling~\citep{bruce2024genie}.
Independent temporal noise levels and history guidance connect sequence
prediction with diffusion~\citep{chen2024diffusion,song2025history}, enabling
long-video synthesis at
scale~\citep{chen2025skyreelsv2infinitelengthfilmgenerative,ai2025magi1autoregressivevideogeneration}.
Pyramidal computation, queued denoising, cache reuse, and context compression
improve streaming efficiency~\citep{jin2024pyramidal,kim2024fifo,gao2024ca2vdm,zhang2025framepack,yuan2026helios}.
To reduce exposure bias, self-resampled histories and recycled generation
errors expose denoising models to imperfect contexts during
training~\citep{guo2025resampling,li2025stableinfinity}.
Another line distills bidirectional models into causal
generators~\citep{yin2025causvid} and supervises self-generated rollouts
to reduce the train--test gap~\citep{huang2025selfforcing}.
Subsequent work improves causal initialization, rolling generation, reward
guidance, and context-gradient
reconstruction~\citep{zhu2026causal,liu2025rolling,lu2025reward,zhuang2026self}.
Long-context supervision and memory mechanisms address temporal
consistency~\citep{yang2025longlive,chen2026context,xue2026ring}, while
on-policy self-distillation and diagonal denoising schedules offer further
training and streaming strategies~\citep{liu2026opsdv,liu2026diagonal}.
Adversarial post-training instead learns through a discriminator, with
generated histories supporting real-time interactive
synthesis~\citep{lin2025apt,lin2025aapt,zhang2025voicebridge}.
We retain autoregressive rollout training but replace online score-based
supervision with a reference-sample objective.

\subsection{Distributional training}

Distributional objectives compare generated and target populations without
paired outputs. Maximum mean discrepancy measures differences between
kernel mean embeddings~\citep{sriperumbudur2010hilbert,gretton2012kernel,muandet2017kernel}
and directly trains generators from
samples~\citep{li2015generative,dziugaite2015training}; adversarially learned
representations strengthen these
objectives~\citep{li2017mmdgan,binkowski2018demystifying}.
Energy distances and sliced optimal transport provide other sample-based
discrepancies~\citep{bellemare2017cramer,deshpande2018sliced}, while
entropically regularized transport connects transport costs and kernel
matching~\citep{genevay2018sinkhorn,feydy2019sinkhorn}.
Distribution evolution can also be described by kernel or sliced-transport
gradient flows~\citep{arbel2019mmdflow,liutkus2019slicedflows}, or learned
through data-attraction and sample-repulsion fields~\citep{drifting}.
Memory-efficient feature-distribution training already separates large
statistical populations from smaller replay batches~\citep{lezama2021runsort}.
Recent approaches optimize Fr\'echet feature moments with decoupled
estimation and differentiation~\citep{yang2026frechet}, or use multiple
frozen encoders and persistent kernel summaries~\citep{irdm}.
ViRDM~\citep{meng2026virdm} studies representation distribution matching
for few-step causal video with stochastic clean exits, staged
vector--Jacobian products, and flow-based dynamics regularization.
Unlike score-based distribution distillation~\citep{yin2024onestep,yin2024improved},
these sample-based objectives can learn from reference data without an
online diffusion teacher. We apply this perspective to autoregressive
video rollouts, studying complementary spatial and temporal representations
and selective differentiation under video-training memory constraints.

\paragraph{Divergence geometry and mode coverage.}
DMD optimizes an approximate reverse KL between noise-perturbed student
and teacher distributions~\citep{yin2024onestep,yin2024improved}.
This objective can favor concentrated solutions under model or
optimization constraints; the original DMD includes regression
regularization to mitigate mode dropping. Data-Forcing Distillation
further studies diversity loss in video distillation and injects reference
data through teacher scores~\citep{chen2026dataforcing}.
Elastic Forcing instead changes both the target specification (reference
samples) and the discrepancy (MMD). These are distinct choices:
training on data alone does not guarantee coverage.

For a characteristic kernel, population MMD is zero only when the two
feature distributions coincide~\citep{gretton2012kernel,sriperumbudur2010hilbert}.
Its attraction term uses reference samples and its generated--generated
term penalizes concentration in kernel space. This motivates distributional
sample matching, but finite data, feature invariances, bandwidth, and
nonconvex optimization can still prevent recovery of missing modes;
even MMD gradient-flow convergence requires additional
conditions~\citep{arbel2019mmdflow}. Our VBench, VLM, FVD, and adaptation
results do not directly measure mode coverage. In particular, video-marginal
matching does not establish diversity within each text condition.
A controlled coverage comparison requires matched target data and
initialization, multiple seeds per prompt, and independent diversity and
precision--recall measures.

\subsection{Video representation learning}

Video encoders capture temporal structure through three-dimensional
convolutions, multiple temporal rates, or space--time
attention~\citep{carreira2017i3d,feichtenhofer2019slowfast,bertasius2021timesformer,patrick2021motionformer,liu2026elasticttt}.
Self-supervised objectives include dense predictive coding and temporal
contrastive learning~\citep{han2019dpc,qian2021cvrl}, masked pixel or
feature reconstruction~\citep{tong2022videomae,feichtenhofer2022mae,wei2022maskfeat,wang2023videomaev2},
and prediction in learned latent spaces~\citep{bardes2024vjepa,assran2025vjepa2}.
Video--text alignment and large-scale multimodal pretraining enrich semantic
features~\citep{xu2021videoclip,wang2024internvideo2};
image self-distillation and language--image alignment supply complementary
appearance cues~\citep{oquab2023dinov2,simeoni2025dinov3,radford2021clip,zhai2023siglip}.
Pretrained representations also improve generation by aligning denoiser
states with clean-image features~\citep{yu2024repa}, which differs from
matching distributions of completed outputs.
Feature selection remains important: video distances can underweight
temporal defects~\citep{unterthiner2018fvd,10655312}, and optimizing one
representation can conceal perceptual failures~\citep{yang2026frechet,irdm}.
This motivates combining video and image encoders and assessing them with
complementary automatic and human judgments.

\subsection{Relationship to Other Approaches}
\label{app:relationship-other-approaches}

\paragraph{Relationship to SwD.}
SwD~\citep{starodubcev2026scalewise} matches per-example spatial-token
statistics in a diffusion teacher's intermediate features, using
linear-kernel MMD in its final formulation. Its MMD-only results for
Wan2.1 establish prior use of standalone MMD for video distillation.
Elastic Forcing matches populations of complete
causal rollouts in external frozen representation spaces, rather than
spatial-token distributions within individual teacher feature maps.
It combines a persistent reference summary, Monte Carlo correction, and
selective differentiation. Our contribution is to make standalone,
reference-defined distribution learning practical for few-step
autoregressive video generation, including scaling to 14B models.

\paragraph{Relationship to RDM.}
Elastic Forcing builds on RDM's feature-space MMD, Nystr\"om reference
summaries, and use of gradient caching~\citep{irdm}. We address the
representation and computational constraints of causal video through
three complementary components. First, our encoder studies motivate
dense self-supervised features that preserve both spatial detail and
temporal evolution (Section~\ref{sec:representation_space} and
Suppl.~\ref{app:encoders}). The tested sparse-feature configuration did
not provide an effective video-training signal, and adding image-level
constraints alone did not resolve the observed temporal artifacts
(Figures~\ref{fig:app-fd} and~\ref{fig:adding-image-loss}). Selecting and
combining suitable representation spaces is therefore part of the video
training procedure.

Second, video rollouts are substantially more expensive than images,
making it useful to compare more videos than we differentiate.
RDM reports a FLUX post-training batch of 10,240 generated images and
uses gradient caching to accumulate the full-batch gradient.
Our estimation--differentiation decoupling instead compares 256 fresh
rollouts while backpropagating through only 128, with a gradient scale
of $256/128=2$ (Equation~\eqref{eq:selective_backward}). This reduces
backward computation as well as activation memory.
Table~\ref{tab:sample_ablation} isolates the benefit of enlarging the
estimation batch while keeping the backward batch fixed.

Third, hybrid reference estimation combines a Nystr\"om summary of the
full reference collection with a Monte Carlo component. The latter
reduces the summary's finite-rank approximation bias, while the
persistent summary reduces reference-minibatch variance
(Suppl.~\ref{app:hybrid-analysis}). Table~\ref{tab:reference_ablation} and
Figure~\ref{fig:nystrom-mc-full} evaluate this combination quantitatively
and qualitatively. The Nystr\"om-only ablation retains our video
representations and selective-differentiation pipeline; it isolates the
reference estimator and is not a reproduction of vanilla RDM.

\paragraph{Relationship to ViRDM.}
Concurrent ViRDM~\citep{meng2026virdm} emphasizes stochastic clean exits,
staged vector--Jacobian products, and flow-based dynamics regularization.
Elastic Forcing emphasizes complementary dense image and video representations for spatial and temporal distribution matching, together with estimation--differentiation decoupling that allows all fresh rollouts to inform the objective while backpropagating through only a selected subset.
Through this design, we provide the first demonstration of both 14B-scale
post-training and reference-driven adaptation within a standalone MMD-based
framework for few-step autoregressive video generation.

\section{Training and Evaluation Details}
\label{app:implementation}

This supplement first describes the experimental setup and comparison methods,
then provides the derivation of the finite-batch objective, reference-side
kernel-mean estimator, and selective-gradient update. We next report the
complete representation-space and optimization ablations, reference-data
adaptation, and additional 14B comparisons. Throughout, \emph{reference videos} specify
the distribution to be learned; a pretrained checkpoint supplies the
initialization but is not used as an online score oracle during Elastic
Forcing post-training.

\subsection{Training setup}
\label{app:training}

Our standard text-to-video experiments retain the autoregressive architecture
and rollout procedure of Self-Forcing~\citep{huang2025selfforcing}.
Each chunk conditions on previously generated content. The few-step
sampling schedule contains four denoising steps for the 1.3B model and
five for the 14B model. The 1.3B model starts from the same ODE-initialized
checkpoint as the Self-Forcing comparison. The 14B experiment uses the
ODE-initialized checkpoint supplied by Krea~\citep{krea_realtime_14b,millon2025krea}. These initialization stages
should be distinguished from the subsequent sample-based post-training.
The teacher-free study additionally evaluates a teacher-forcing
checkpoint and real UltraVideo references
(Section~\ref{sec:teacher-free}; Suppl.~\ref{app:initialization-data}).
The separate VACE experiment in Suppl.~\ref{app:vace} examines a different,
visually conditioned setting without a separate ODE initialization stage.

\begin{table}[tp]
    \centering
    \EFTableTypography
    \caption{\textbf{Reported settings for standard text-to-video post-training.}
    Both model scales use an ODE-initialized checkpoint.}
    \label{tab:app-training}
    \begin{tabularx}{\linewidth}{lNN}
        \toprule
        \rowcolor{EFTableHeader}
        \textbf{Setting} & \textbf{1.3B model} & \textbf{14B model} \\
        \midrule
        Backbone & Wan2.1-T2V-1.3B & Wan2.1-T2V-14B \\
        Initialization & Self-Forcing ODE checkpoint & Krea ODE checkpoint \\
        Denoising steps & 4 per chunk & 5 per chunk \\
        Hardware & 4 NVIDIA H200 GPUs & 8 NVIDIA H200 GPUs \\
        \bottomrule
    \end{tabularx}
\end{table}

For the 1.3B run, we use $B=256$ current rollouts to evaluate the
distributional objective and select $b=128$ of them for backpropagation. The three encoder variant in Table~\ref{tab:short-video-comparison}, which is the main variant and further used in other evaluations, is trained for 150 steps. Table~\ref{tab:app-encoders} summarizes the encoder settings used for post-training. DINOv3 and V-JEPA2 each provide 1024 tokens per video, while VideoMAE v2 provides 1280 tokens, with a feature dimension of 1024 per token across all three encoders. We use a 2:1 weighting to balance the Nystr\"om and MC estimators. The encoder-specific loss terms are combined as $(10N_{\mathrm{DINOv3}} + N_{\mathrm{V\text{-}JEPA2}} + N_{\mathrm{VideoMAE}})/3$, where the weights are roughly selected to balance the numerical values.

\paragraph{Hyperparameter selection.}
Implementation hyperparameters, such as the MMD bandwidths,
reference-estimator configuration, number of Nystr\"om landmarks, and
loss weights, were coarsely selected in early experiments on a small
dataset. In these preliminary experiments,
varying these hyperparameters within reasonable ranges had little
noticeable effect on the final evaluation results. We did not perform
fine-grained hyperparameter searches in the large-scale experiments or
tailor the settings to individual evaluation datasets, reducing the risk
of dataset-specific tuning. These preliminary observations suggest
robustness to moderate hyperparameter changes within the tested ranges.

\paragraph{Rollout gradients and replay.}
\label{app:rollout-gradients}
Training uses Self Forcing's stochastic denoising exit and truncated
gradient path. For each chunk, only the denoiser evaluation at the sampled
exit receives parameter gradients; its input from earlier denoising
steps and the cross-chunk history/KV cache are treated as constants.
The decoded video still receives gradients through the frozen feature
encoders. Replay reconstructs the same exit choices, pre-exit inputs,
and detached context, and reproduces this training graph.
Activation recomputation within the differentiable path preserves its
gradients; it does not restore gradients through the deliberately
detached denoising and history dependencies.
Accordingly, the selective-gradient identity in
Suppl.~\ref{app:selective_backward} is conditional on this truncated
graph.

\paragraph{Dense feature readouts.}
Each video is one sample for MMD. We retain an ordered feature matrix
$H_e(x)\in\mathbb R^{N_e\times d_e}$ and use
$\phi_e(x)=\operatorname{vec}H_e(x)$; Euclidean distance between these
vectors is the Frobenius distance between the complete feature matrices.
DINOv3 supplies 256 patch tokens for each of 20 sampled frames. We average
corresponding patches within four consecutive groups of five sampled
frames, then concatenate the four groups to obtain 1,024 tokens.
This local temporal pooling reduces the feature dimension by a factor of
five while preserving the ordering of the four groups.
V-JEPA~2 supplies 1,024 selected tokens from its last-64-frame window;
VideoMAE~v2 supplies 256 selected tokens from each of five consecutive
16-frame segments after skipping the leading frame.
The kernel uses these complete ordered features without an additional
global unit-norm projection. Each encoder's bandwidth is stored with
its reference summary and held fixed throughout post-training.

\begin{table}[tp]
    \centering
    \EFTableTypography
    \caption{\textbf{Encoder settings for post-training.}}
    \label{tab:app-encoders}
    \begin{tabularx}{\linewidth}{lNN}
        \toprule
        \rowcolor{EFTableHeader}
        \textbf{Encoder} & \textbf{Tokens per video} & \textbf{Dimension per token} \\
        \midrule
        DINOv3      & 1024 & 1024  \\
        V-JEPA2     & 1024 & 1024 \\
        VideoMAE v2 & 1280 & 1024 \\
        \midrule
        Loss & \multicolumn{2}{c}{%
            $\bigl(10N_{\mathrm{DINOv3}}
            + N_{\mathrm{V\text{-}JEPA2}}
            + N_{\mathrm{VideoMAE}}\bigr)/3$} \\
        Nystr\"om:MC hybrid & \multicolumn{2}{c}{$2:1$} \\
        \bottomrule
    \end{tabularx}
\end{table}

The main encoder summary in
Table~\ref{tab:video_encoder_ablation} includes the updated three-encoder
aggregate scores also reported in Table~\ref{tab:short-video-comparison}.
The $B/b$ and initialization ablations use 150 updates.
These 1.3B ablations are conducted on 4 H200 GPUs.
For the 14B model, we post-train it for 80 steps on 8 H200 GPUs, costing 23.2 hours. Details are listed in Table~\ref{tab:app-training}
We use two general-domain reference collections, each containing 8,000
videos. The Wan collection is generated by Wan2.1 at the corresponding
model scale~\citep{wan2025}, with prompts prepared from VidProM~\citep{wang2024vidprom}
using DeepSeek-V4-Flash~\citep{deepseekai2026deepseekv4}.
The UV collection consists of randomly sampled UltraVideo
clips~\citep{xue2025ultravideo}. The frozen encoders map
reference and generated videos into the same representation spaces. Only
the generator is updated: neither a real-score diffusion teacher nor an
online fake-score model is required by the post-training objective.
Reference-feature extraction and construction of the persistent
Nystr\"om summary can therefore be performed before the optimization loop.

We distinguish two representation configurations. The two-video-encoder
variant combines V-JEPA~2 and VideoMAE and gives the strongest aggregate
VBench result in the encoder ablation. The three-encoder variant also
includes DINOv3 and is used for the scaling and reference-adaptation experiments in
Sections~\ref{sec:scaling} and~\ref{sec:adaptation}.

\paragraph{Comparison with Krea Realtime at 14B.}
Both EF-14B and Krea Realtime use five denoising steps per chunk, and
EF-14B starts from the ODE-initialization checkpoint officially released
by Krea~\citep{krea_realtime_14b,millon2025krea}. For the Krea baseline,
we evaluate the officially released checkpoint. Despite substantial
effort to reproduce its training procedure, our attempts encountered
out-of-memory (OOM) errors on eight NVIDIA H200 GPUs, and we were unable
to successfully start training within this hardware budget. Krea
additionally incorporates optimizations to KV caching and related inference
components. These improvements are orthogonal to EF's
distribution-matching objective. To maintain consistency with our 1.3B
setup, EF-14B retains the inference formulation used in those experiments
and does not adopt Krea's additional optimizations. Even without these
optimizations, EF achieves higher VLM Total scores and mean rank points
than Krea Realtime (Table~\ref{tab:vlm_comparison}), demonstrating the
effectiveness of its distribution-matching post-training at 14B.

\paragraph{Comparison with Self-Forcing at 1.3B.}
EF-1.3B fully retains Self-Forcing's inference implementation and
settings~\citep{huang2025selfforcing}, including four denoising steps per
chunk and the same KV-cache configuration. We evaluate both methods
using the same prompt set and inference seeds, controlling these
sources of variation in the comparison.

\subsection{Evaluation protocols}
\label{app:evaluation}
\label{app:user_study}

\paragraph{VBench.}
Following the Self-Forcing evaluation setup, we use the
946 VBench prompts~\citep{huang2023vbench} and generate five videos per prompt with
independent random seeds, giving 4,730 videos per evaluated model.
We report Total, Quality, and Semantic scores on a 0--100 scale. In the
reported tables, Total is the weighted aggregate
$0.8\,\mathrm{Quality}+0.2\,\mathrm{Semantic}$, up to rounding.

\paragraph{Inference-seed variability.}
With respect to inference-seed noise, the estimated standard error of
the mean VBench Total score is 0.0935. Relative to Self-Forcing (83.95),
the three-encoder model (84.25) and two-encoder model (84.64) improve by
0.30 and 0.69, respectively (Table~\ref{tab:video_encoder_ablation}),
both larger than this standard error. We interpret the benchmark results
together with the qualitative and independent evaluations, rather than
treating the magnitude of the VBench Total difference alone as decisive.

\paragraph{VLM evaluation set.}
For independent VLM scoring, we use a fixed subset of 100 VBench prompts,
sampled uniformly without replacement from the 944 unique
prompts in the 946-entry official list using random seed 20260920.
For duplicate prompt strings, we retain their last official occurrence.
For each selected prompt, we evaluate the first generated sample
(sample index~0) from each method, yielding 100 videos per method.
The prompt subset and sample index are fixed across configurations and
reused without resampling or score-based selection. We will release this
prompt subset for reproducibility.

\paragraph{VLM ranking.}
For each matched prompt and inference seed, the evaluator jointly
compares anonymous videos generated by different methods and assigns
relative rank points. Matching the prompt and seed controls these
sources of variation and supports a consistent comparison across
methods. This protocol highlights differences among closely related
model variants and is therefore useful
for ablation studies. In the main paper, we use ranking for the encoder,
$B/b$, and marginal versus joint matching ablations
(Tables~\ref{tab:video_encoder_ablation}, \ref{tab:sample_ablation},
and~\ref{tab:marginal_joint}), as well as the 14B and initialization
comparisons (Tables~\ref{tab:vlm_comparison} and~\ref{tab:initialization_data}).
Rank points depend on the competing set and should be interpreted within
that set. The ranking prompt is provided below in
Listing~\ref{lst:vlm_joint_ranking}.

\paragraph{VLM scores.}
We use independent VLM scores to assess each video against a common,
predefined rubric. The evaluator scores visual quality, semantic
consistency, subject consistency, and motion smoothness; Total is their
equal-weight mean up to rounding. These scores are reported in the main
baseline and 14B comparisons (Tables~\ref{tab:short-video-comparison}
and~\ref{tab:vlm_comparison}), as well as the full initialization and
marginal versus joint matching results in the supplement
(Tables~\ref{tab:initialization_data_full} and~\ref{tab:marginal_joint_full}).
The main baseline table includes many methods, for which forced ranking
can assign substantially different rank points to videos of similar
perceptual quality. We therefore report only independent VLM scores in
Table~\ref{tab:short-video-comparison}. Rank points and independent scores
should not be compared numerically. The scoring prompt is provided below
in Listing~\ref{lst:single-video-vlm-prompt}.

\paragraph{VLM consistency and reliability.}
We schedule GPT-6 evaluations within as narrow a time window as
practicable to minimize potential effects of model-version changes.
Each request is stateless and contains no conversation history or
memory carried over from previous evaluations. Agreement between
pairwise VLM judgments and human preferences provides further
empirical support for the reliability of these judgments in the evaluated setting;
Suppl.~\ref{sec:vlm-human-correlation} reports this analysis, including
Figure~\ref{fig:vlm-human-correlation} and
Table~\ref{tab:vlm-human-correlation}.

\paragraph{Specialized adaptation.}
Table~\ref{tab:specialized_video_evaluation} reports ratings on a 0--10 scale
for monochrome appearance, Nailong character generation, and spatial
appearance. The task-specific score concerns adherence to the corresponding
target; the remaining columns describe general visual and temporal quality.
Task adherence and general quality are reported separately, since improving
one does not imply uniform improvement in the others. These task ratings, the large-model
VLM/human ratings, and VBench use different protocols and should be compared
within their respective tables.

\subsection{Distributional evaluation with FVD}
\label{app:fvd}
In addition, we adopt a distribution-based evaluation protocol to assess the generative capability of our method. Specifically, we synthesize the VBench evaluation set with Wan2.1-14B and conduct an FVD comparison, in which each student model's outputs are measured against the same reference distribution.

\begin{table}[tp]
    \centering
    \EFTableTypography
    \caption{\textbf{FVD evaluation using Wan-14B reference videos.}
    Lower FVD is better.}
    \label{tab:fvd_evaluation}
    \begin{tabularx}{\linewidth}{lN}
        \toprule
        \rowcolor{EFTableHeader}
        \textbf{FVD evaluation} & \textbf{Value} \\
        \midrule
        Reference videos         & $946 \times 5 = 4{,}730$ \\
        Feature extractor        & I3D \\
        Self-Forcing FVD $\downarrow$    & 266.86 \\
        \rowcolor{EFTableOurs}
        Elastic Forcing FVD $\downarrow$ & \textbf{217.12} \\
        Relative reduction       & \textbf{18.6\%} \\
        \bottomrule
    \end{tabularx}
\end{table}

\subsection{Comparison methods}
\label{app:baselines}

The main comparison focuses on Wan2.1, CausVid, and Self-Forcing.
Table~\ref{tab:extended-baselines} retains a broader set of full-sequence and
streaming generators for context. Their architectures, initialization,
and training procedures differ. Self-Forcing provides the closest
comparison to our post-training objective because the 1.3B architecture,
initialization, and rollout procedure are matched. Its VBench scores are
the updated values also used in Table~\ref{tab:short-video-comparison}.

\begin{table}[tp]

    \centering
    \EFTableTypography

    \captionof{table}{\textbf{Additional baseline context.} We report model size, inference throughput, and VBench total,
    quality (Qual.), and semantic (Sem.) scores. Shared models use the updated
    results in Table~\ref{tab:short-video-comparison}. Bold marks the best
    scores among autoregressive models.}
    \label{tab:extended-baselines}
    \begin{tabularx}{\linewidth}{l*{5}{N}}
        \toprule
        \rowcolor{EFTableHeader}
        & & & \multicolumn{3}{c}{\EFMetricGroup{VBench}} \\
        \rowcolor{EFTableHeader}
        \multirow{-2}{*}{\textbf{Model}} & \multirow{-2}{*}{\textbf{Size}}
        & \multirow{-2}{*}{\EFMetricGroup{FPS}}
        & \textbf{Total} & \textbf{Qual.} & \textbf{Sem.} \\
        \midrule
        \rowcolor{EFTableGroup}
        \multicolumn{6}{l}{\textit{Full-sequence diffusion}} \\
        LTX-Video & 1.9B & 8.98 & 80.00 & 82.30 & 70.79 \\
        Wan2.1 & 1.3B & 0.78 & 84.26 & 85.30 & 80.09 \\
        \midrule
        \rowcolor{EFTableGroup}
        \multicolumn{6}{l}{\textit{Autoregressive / streaming}} \\
        SkyReels-V2 & 1.3B & 0.49 & 82.26 & 83.61 & 76.85 \\
        MAGI-1 & 4.5B & 0.19 & 82.28 & 82.73 & 80.47 \\
        NOVA & 0.6B & 0.88 & 79.29 & 79.91 & 76.81 \\
        Pyramid Flow & 2B & 6.7 & 81.63 & 83.78 & 73.02 \\
        CausVid & 1.3B & 17.0 & 83.02 & 84.16 & 78.45 \\
        Self-Forcing & 1.3B & 17.0 & 83.95 & 84.73 & 80.83 \\
        LongLive & 1.3B & 20.7 & 83.22 & 83.68 & \textbf{81.37} \\
        Rolling Forcing & 1.3B & 17.5 & 81.22 & 84.08 & 69.78 \\
        Reward Forcing & 1.3B & 23.1 & 84.13 & 84.84 & 81.32 \\
        \midrule
        \rowcolor{EFTableVariant}
        EF (Wan) & 1.3B & 17.0
        & \textbf{84.25} & \textbf{85.06} & 80.99 \\
        \rowcolor{EFTableOurs}
        EF (UV) & 1.3B & 17.0
        & 83.65 & 84.42 & 80.57 \\
        \bottomrule
    \end{tabularx}

\end{table}

\paragraph{Full-sequence diffusion.}
\textbf{LTX-Video}~\citep{hacohen2024ltxvideo} uses a highly compressed video VAE and a diffusion
transformer for efficient full-sequence synthesis.
\textbf{Wan2.1}~\citep{wan2025} supplies the 1.3B and 14B foundation
backbones used in this work; its bidirectional models also provide
full-sequence quality references.

\paragraph{Other autoregressive generators.}
\textbf{SkyReels-V2}~\citep{chen2025skyreelsv2infinitelengthfilmgenerative}
uses Diffusion Forcing to extend videos autoregressively.
\textbf{MAGI-1}~\citep{ai2025magi1autoregressivevideogeneration}
performs chunk-wise autoregressive denoising with different noise levels
across chunks. \textbf{NOVA}~\citep{deng2024nova} uses non-quantized visual representations and
factorizes generation across temporal and spatial positions.
\textbf{Pyramid Flow}~\citep{jin2024pyramidal} combines spatial and temporal
pyramids to reduce the cost of video synthesis and history conditioning.

\paragraph{DMD-based streaming generators.}
\textbf{CausVid}~\citep{yin2025causvid} distills a bidirectional diffusion
model into a few-step causal generator.
\textbf{Self-Forcing}~\citep{huang2025selfforcing} applies distributional
supervision to rollouts conditioned on their own generated histories,
reducing the mismatch between training and inference.
\textbf{LongLive}~\citep{yang2025longlive} extends this setting with
long-horizon tuning, persistent frame-sink tokens, and prompt-dependent
KV recaching. \textbf{Rolling Forcing}~\citep{liu2025rolling}
uses joint denoising within rolling temporal windows and persistent
attention sinks. \textbf{Reward Forcing}~\citep{lu2025reward} combines
reward-guided distribution matching with an evolving memory of earlier
frames. The evaluated variants in this family retain score-based
supervision. Elastic Forcing instead changes how the rollout distribution
is supervised, and can be combined with improvements to temporal context
or memory handling.

\subsection{Training Dynamics}
The available three-encoder measurements show improvement from 100 to 150 updates. We selected the budgets by considering generation-quality convergence and computational cost. As shown in Figure~\ref{fig:training_dynamic}, with the training dataset and all other settings held fixed, our VBench scores remain stable throughout training, with no signs of overfitting or collapse. This indicates that our method is robust to the choice of the number of training steps.
\begin{figure}[tp]
    \centering
    \includegraphics[width=1\linewidth]{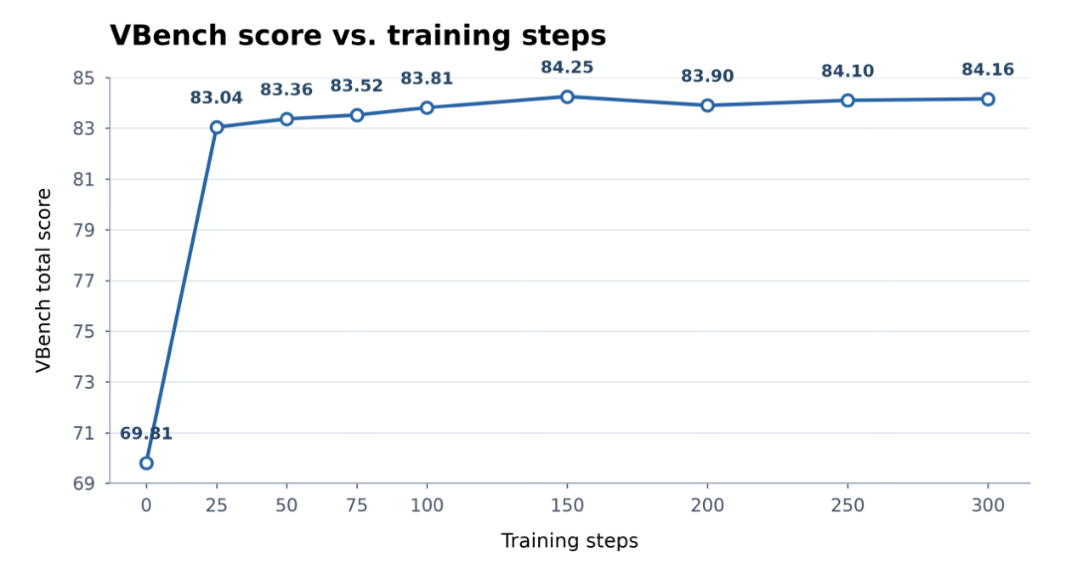}
    \caption{\textbf{VBench total score during post-training.}
The score increases sharply from 69.81 at initialization to 83.04 within 25 steps, after which it improves gradually and plateaus around 84.2, peaking at 84.25 at step 150. The score stays within 0.4 of this peak up to step 300, indicating that the gain is stable rather than a transient fluctuation.}
    \label{fig:training_dynamic}
\end{figure}

\subsection{Additional evaluation}
\label{app:additional-wan-evaluation}

The main paper reports human preferences for EF-14B (UV) against Krea
Realtime (Figure~\ref{fig:teaser}). Table~\ref{tab:additional-wan-evaluation}
additionally reports human ratings for EF-14B (Wan), together with a
separate VLM evaluation. The latter uses another subset of
100 VBench prompts, drawn using a different prompt-selection seed from the main evaluation. We will also release this additional prompt subset
for reproducibility. These auxiliary VLM scores should not be compared
numerically with the main because the prompt subset differ.

\begin{table}[tp]
    \centering
    \EFTableTypography
    \caption{\textbf{Additional VLM and human evaluation of EF-14B (Wan).}
    VLM scores are evaluated on a separately sampled 100-prompt subset.
    Total is the equal-weight mean of the four component scores.
    Qual.: visual quality; Sem.: semantic consistency; Subj.: subject
    consistency; Smoo.: motion smoothness.
    Bold marks the best value among autoregressive models in each column;
    -- denotes unreported human ratings.}
    \label{tab:additional-wan-evaluation}
    \begin{tabularx}{\linewidth}{l*{6}{N}}
        \toprule
        \rowcolor{EFTableHeader}
        & \multicolumn{5}{c}{\EFMetricGroup{VLM scores}} & \EFMetricGroup{Human} \\
        \rowcolor{EFTableHeader}
        \multirow{-2}{*}{\textbf{Model}}
        & \textbf{Total} & \textbf{Qual.} & \textbf{Sem.}
        & \textbf{Subj.} & \textbf{Smoo.} & \textbf{Score} \\
        \midrule
        \rowcolor{EFTableGroup}
        \multicolumn{7}{l}{\textit{Full-sequence diffusion}} \\
        Wan2.1-14B & 4.29 & 4.23 & 4.58 & 4.33 & 4.03 & -- \\
        \midrule
        \rowcolor{EFTableGroup}
        \multicolumn{7}{l}{\textit{Autoregressive / streaming}} \\
        Self-Forcing & 4.15 & 4.01 & 4.56 & 4.11 & 3.92 & -- \\
        Krea Realtime & 4.13 & 4.00 & 4.50 & 4.05 & 3.97 & 3.054 \\
        \rowcolor{EFTableOurs}
        EF-14B (Wan) & \textbf{4.30} & \textbf{4.21} & \textbf{4.61}
        & \textbf{4.35} & \textbf{4.01} & \textbf{3.354} \\
        \bottomrule
    \end{tabularx}
\end{table}

The additional human evaluation favors EF-14B (Wan) over Krea Realtime
(3.354 vs.\ 3.054). Despite the change in VLM version and prompt-selection
seed, the overall ordering of EF-14B (Wan) above Krea Realtime remains
consistent with the main evaluation in Table~\ref{tab:vlm_comparison} and
with the human ratings reported here. This consistency across evaluator
versions and separately sampled test subsets provides further evidence
for the reliability of the VLM-based pairwise comparison.

\section{Finite-Batch Objective and Efficient Gradients}
\label{app:optimization}

\subsection{Finite-batch training objective}
\label{app:empirical-objective}

For each encoder $e$, let $z_i^e=\phi_e(x_i)$ be the representation of the
$i$-th current rollout. With $B>1$ rollouts, the objective used to obtain
representation gradients is
\begin{equation}
    \widehat{\mathcal L}_{\mathrm{EF}}
    =\sum_{e\in\mathcal E}\lambda_e
    \left[
    \frac{1}{B(B-1)}\sum_{i\ne j}k_e(z_i^e,z_j^e)
    -\frac{2}{B}\sum_{i=1}^{B}\widehat m_{\alpha_e}^e(z_i^e)
    \right].
    \label{eq:ef_empirical}
\end{equation}
The generated--generated sum contains ordered pairs and excludes the
self-pairs $i=j$~\citep{gretton2012kernel}. The reference--reference term is omitted because the
reference distribution and encoders are fixed, so it contributes no
generator gradient. Thus, the displayed training loss need not be
nonnegative and should not be interpreted as an absolute MMD score.
The hybrid attraction term further approximates the original-kernel MMD
objective; it does not replace the generated--generated kernel by the
Nystr\"om kernel.

For the fixed batch, define each feature-gradient contribution as
$g_i^e=\partial\widehat{\mathcal L}_{\mathrm{EF}}/\partial z_i^e$.
Symmetry of the kernel gives
\begin{equation}
    g_i^e=\lambda_e\left[
    \frac{2}{B(B-1)}\sum_{j\ne i}\nabla_1 k_e(z_i^e,z_j^e)
    -\frac{2}{B}\nabla_z\widehat m_{\alpha_e}^e(z_i^e)
    \right],
    \label{eq:app-cotangent}
\end{equation}
where $\nabla_1$ differentiates the first kernel argument. The factor of
two in the repulsion term accounts for the appearance of each sample in
both positions of the ordered pair sum. Computing losses separately on
small microbatches would omit interactions between those microbatches.

\paragraph{Computing only selected feature gradients.}
We uniformly select $b$ indices $\mathcal S$ before feature
differentiation and evaluate Equation~\eqref{eq:app-cotangent} only for
$i\in\mathcal S$. All $B$ feature vectors remain available as comparison
values, while only the $b$ selected gradients per encoder are computed
and cached. Equivalently, these gradients can be obtained from the scalar
\begin{equation}
    \mathcal Q_{\mathcal S}
    =\frac{2}{B}\sum_{i\in\mathcal S,e}\lambda_e\left[
    \frac{\sum_{j\ne i}k_e(z_i^e,\operatorname{sg}(z_j^e))}{B-1}
    -\widehat m_{\alpha_e}^e(z_i^e)\right],
    \label{eq:app-selected-query-loss}
\end{equation}
where $\operatorname{sg}$ stops gradients through every comparison
argument, including those whose indices also lie in $\mathcal S$.
Differentiation with respect to the selected query features gives
$\nabla_{z_i^e}\mathcal Q_{\mathcal S}=g_i^e$ for $i\in\mathcal S$.
The factor of two retains both appearances of each query in the
full-batch pair sum. Thus only selected-to-all kernel derivatives and
selected reference queries are needed; interactions between two
unselected videos need not be evaluated.

\subsection{Selective differentiation and replay}
\label{app:selective_backward}

Condition on a fixed generated batch and the randomness used to construct
the reference estimate. For the analysis, define each rollout's parameter-gradient
contribution by
\begin{equation}
    h_i=\sum_{e\in\mathcal E}(J_{\theta,i}^{e})^\top g_i^e,
    \qquad
    J_{\theta,i}^{e}=\frac{\partial z_i^e}{\partial\theta}.
\end{equation}
The full-batch gradient is $\sum_i h_i$. For a uniformly sampled subset
$\mathcal S$ of size $b$ without replacement~\citep{horvitz1952sampling},
$\Pr(i\in\mathcal S)=b/B$, and therefore
\begin{equation}
    \mathbb E_{\mathcal S}\left[\frac{B}{b}
        \sum_{i\in\mathcal S}h_i\right]
    =\frac{B}{b}\sum_{i=1}^{B}\frac{b}{B}h_i
    =\nabla_\theta\widehat{\mathcal L}_{\mathrm{EF}}.
    \label{eq:app-unbiased-gradient}
\end{equation}
This is conditional unbiasedness for the \emph{fixed-batch hybrid objective},
not for an exact population MMD. It also assumes that the same rollout
computation and any prescribed gradient truncation are used during replay.
For $\bar h=B^{-1}\sum_i h_i$ and
$S_h=(B-1)^{-1}\sum_i(h_i-\bar h)(h_i-\bar h)^\top$, the conditional
covariance is
\begin{equation}
    \operatorname{Cov}_{\mathcal S}\left[\frac{B}{b}
        \sum_{i\in\mathcal S}h_i\right]
    =\frac{B^2}{b}\left(1-\frac{b}{B}\right)S_h.
    \label{eq:app-subsampling-variance}
\end{equation}
Thus reducing $b$ saves reverse computation but increases update variance;
it does not reduce the number of samples informing each cotangent.
At $b=B$, this source of variance vanishes.

\paragraph{Replay requirements.}
To reproduce the intended backward computation, the saved replay state
must permit reconstruction of the same random choices and segment inputs. Reverse execution restores
a boundary, reconstructs that segment, propagates its incoming cotangent,
and releases the segment's graph~\citep{chen2016sublinear,gao2021gradcache}. Frozen encoder weights still permit
derivatives with respect to their inputs. Reusing the same random choices,
preprocessing, and parameter values is essential: replaying a different
video would pair a cotangent with the wrong Jacobian. Boundary gradients
must be propagated wherever the original computation requires them;
detaching a boundary introduces truncation rather than an equivalent
memory-saving execution. The optimizer is updated only after all selected
contributions have been accumulated.

\begin{algorithm}[tp]
    \caption{Elastic Forcing with a persistent
    summary and selective replay}
    \label{alg:app-elastic-forcing}
    \small
    \begin{algorithmic}[1]
        \Require Generator $G_\theta$; reference collection $\mathcal D_{\rm ref}$;
        frozen $\{\phi_e,k_e,\lambda_e,\alpha_e\}_{e\in\mathcal E}$;
        $B>1$, $1\le b\le B$, $B_{\rm ref}\ge1$, $R$, $\varepsilon>0$
        \ForAll{$e\in\mathcal E$}
            \State Extract reference features and obtain $R$ landmarks by $k$-means
            \State Form $K_{UU}^e$, accumulate $\mathbf c_e$, and solve
            Equation~\eqref{eq:app-coefficient-system} for $\boldsymbol\beta_e$
        \EndFor
        \Repeat
            \State Generate $B$ current autoregressive rollouts without retained graphs;
            save replay states
            \State Extract and cache all comparison features $\{z_i^e\}_{i,e}$
            \State Sample a reference minibatch; combine its Monte Carlo mean with
            $\mathbf k_e(U_e,z)^\top\boldsymbol\beta_e$ using Equation~\eqref{eq:hybrid_reference}
            \State Uniformly sample $b$ distinct indices $\mathcal S$
            \State Compute, detach, and cache only $\{g_i^e\}_{i\in\mathcal S,e}$
            using all $B$ comparison features in Equation~\eqref{eq:app-cotangent}
            \State Clear the generator's accumulated gradient
            \ForAll{$i\in\mathcal S$}
                \State Replay rollout $i$ with its saved states and propagate
                $(B/b)g_i^e$ through each encoder and the generator
                \State Accumulate the parameter gradient and release the replayed graph
            \EndFor
            \State Apply one optimizer update to $\theta$
        \Until{the post-training budget is exhausted}
    \end{algorithmic}
\end{algorithm}

\section{Hybrid Reference Estimation}
\label{app:reference_estimation}

The reference collection is fixed during post-training, whereas generated
features must follow the current generator. This section gives the complete
reference-side implementation used in Equation~\eqref{eq:finite_batch_loss}:
Monte Carlo kernel means, a persistent Nystr\"om summary, their mixture,
and the corresponding analysis and ablations. The estimator complements
the representation and differentiation choices in the main paper.

\subsection{Nystr\"om approximation of the reference kernel mean}
\label{app:nystrom}

Fix an encoder, kernel, and reference collection. Write
$v_n^e=\phi_e(y_n)$ and use the empirical reference mean
\begin{equation}
    m_{\mathrm{ref}}^e(z)
    =\frac{1}{N_{\mathrm{ref}}}\sum_{n=1}^{N_{\mathrm{ref}}}k_e(z,v_n^e).
    \label{eq:app-exact-reference}
\end{equation}
This equality defines the finite-data target used in the main text; it
does not assume that the reference collection exactly represents an
underlying population distribution.
A minibatch of $B_{\rm ref}$ independent reference draws gives
\begin{equation}
    \widehat m_{\rm MC}^e(z)=\frac{1}{B_{\rm ref}}
    \sum_{j=1}^{B_{\rm ref}}k_e(z,\phi_e(y_j)),
    \qquad y_j\overset{\mathrm{i.i.d.}}{\sim}p_{\rm ref}.
    \label{eq:mc_reference}
\end{equation}
This estimate is unbiased for the empirical target, but its variance can
be large when the reference minibatch is small. We complement it with a
persistent summary of the full fixed reference collection.

\paragraph{Landmarks and coefficients.}
Let $U_e=\{u_r^e\}_{r=1}^{R}$ be the landmarks obtained by $k$-means in the
reference representation space~\citep{zhang2008nystrom}. The centroids need not coincide with
individual reference videos. Define
\begin{equation}
    [K_{UU}^e]_{rs}=k_e(u_r^e,u_s^e),\qquad
    \mathbf k_e(U_e,z)=
    [k_e(u_1^e,z),\ldots,k_e(u_R^e,z)]^\top,
    \label{eq:app-landmark-statistics}
\end{equation}
and
\begin{equation}
    \mathbf c_e=\frac{1}{N_{\mathrm{ref}}}
    \sum_{n=1}^{N_{\mathrm{ref}}}\mathbf k_e(U_e,v_n^e).
\end{equation}
We approximate the reference mean in the span of landmark kernel
functions~\citep{williams2000nystrom,chatalic2022nystrom,irdm}:
\begin{equation}
    m_{\mathrm{Nys}}^e(z)=\mathbf k_e(U_e,z)^\top\boldsymbol\beta_e.
    \label{eq:app-landmark-expansion}
\end{equation}
If $K_{UU}^e$ is invertible, matching the empirical mean at every landmark
amounts to solving $K_{UU}^e\boldsymbol\beta_e=\mathbf c_e$.
For numerical stability, use
\begin{equation}
    (K_{UU}^e+\varepsilon I)\boldsymbol\beta_e=\mathbf c_e,
    \qquad\varepsilon>0.
    \label{eq:app-coefficient-system}
\end{equation}
Regularization relaxes exact interpolation: the residual at the landmarks
is $\mathbf c_e-K_{UU}^e\boldsymbol\beta_e
=\varepsilon\boldsymbol\beta_e$. The coefficients $\boldsymbol\beta_e$
are distinct from the scalar mixture weight $\alpha_e$ below.

\paragraph{Equivalent feature map.}
The regularized Gram matrix is symmetric positive definite. Therefore,
\begin{equation}
    \psi_e(z)=(K_{UU}^e+\varepsilon I)^{-1/2}\mathbf k_e(U_e,z),
    \qquad
    \bar\mu_e=\frac{1}{N_{\mathrm{ref}}}\sum_n\psi_e(v_n^e)
    \label{eq:nystrom_features}
\end{equation}
yield
\begin{align}
    \psi_e(z)^\top\bar\mu_e
    &=\mathbf k_e(U_e,z)^\top
      (K_{UU}^e+\varepsilon I)^{-1}\mathbf c_e \nonumber\\
    &=\mathbf k_e(U_e,z)^\top\boldsymbol\beta_e
      =m_{\mathrm{Nys}}^e(z).
    \label{eq:nystrom_reference}
\end{align}
The stored mean incorporates all reference videos, while a query uses only
landmark kernel evaluations.
The associated kernel $\widetilde k_e(z,u)=\psi_e(z)^\top\psi_e(u)$ has
rank at most $R$.

\paragraph{Precomputation and cost.}
Once the representation space and reference set are fixed, we can solve
for $\boldsymbol\beta_e$ once and evaluate each subsequent query using
$R$ kernel values and a dot product. Forming the dense landmark Gram matrix
uses $O(R^2)$ kernel evaluations, and accumulating $\mathbf c_e$ uses
$O(N_{\mathrm{ref}}R)$; the latter can be streamed without storing the full
reference--landmark matrix. A dense factorization costs $O(R^3)$ arithmetic.
These costs exclude feature extraction and $k$-means. Using the coefficient
form avoids a dense inverse-square-root multiplication for every query.
If the encoder, kernel, or reference collection changes, the persistent
statistics must be recomputed.

\subsection{Hybrid estimation and its bias--variance trade-off}
\label{app:hybrid-definition}
\label{app:hybrid-analysis}

We combine the persistent approximation and the sampled estimate:
\begin{equation}
    \widehat m_{\alpha_e}^e(z)
    =(1-\alpha_e)m_{\rm Nys}^e(z)
      +\alpha_e\widehat m_{\rm MC}^e(z),
    \qquad \alpha_e\in[0,1].
    \label{eq:hybrid_reference}
\end{equation}
This is the evaluator denoted $\widehat m_{\rm ref}^e$ in the main text.
Our implementation uses a fixed $2:1$ Nystr\"om-to-Monte-Carlo weighting,
so $\alpha_e=1/3$. The generated--generated term retains its original
kernel. Figure~\ref{fig:hybrid_toy} illustrates the mixture on a toy
problem; the following analysis specifies what its bias--variance benefit
means.

\begin{figure}[tp]
    \centering
    \includegraphics[width=0.4\linewidth]{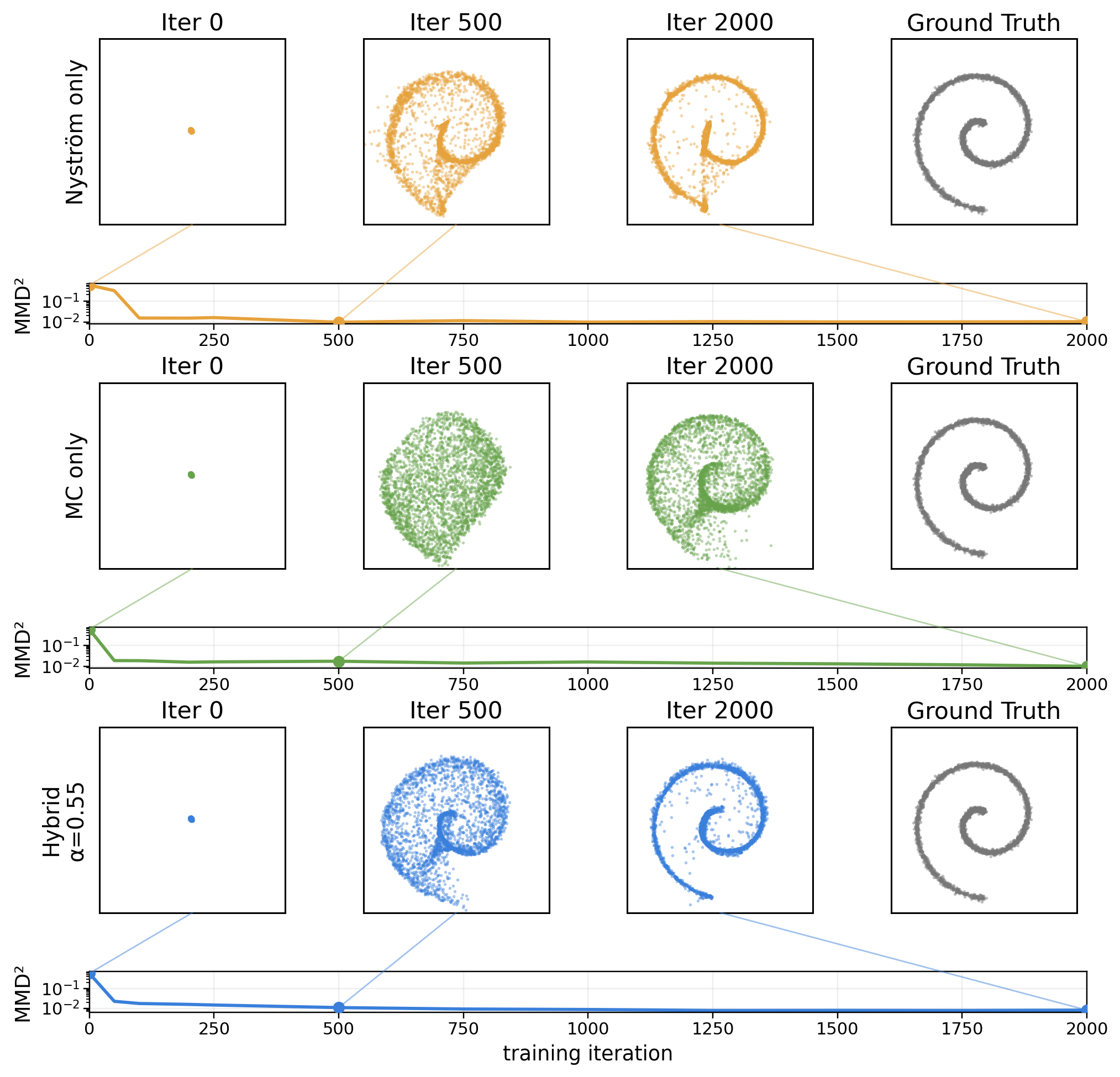}
    \caption{\textbf{Hybrid reference estimation in a 2D toy experiment.}
    The combined estimator complements a persistent reference summary with
    sampled kernel comparisons. This illustration is separate from the
    video representation and selective-differentiation experiments.}
    \label{fig:hybrid_toy}
\end{figure}

At a fixed query $z$, condition on the reference collection and landmarks.
Let
\begin{equation}
    \delta_e(z)=m_{\mathrm{Nys}}^e(z)-m_{\mathrm{ref}}^e(z),\qquad
    \eta_e(z)=\widehat m_{\mathrm{MC}}^e(z)-m_{\mathrm{ref}}^e(z).
\end{equation}
For independent reference draws with replacement,
$\mathbb E[\eta_e(z)]=0$ and
\begin{equation}
    v_e(z):=\operatorname{Var}[\eta_e(z)]
    =\frac{\operatorname{Var}_{Y\sim p_{\mathrm{ref}}}
        [k_e(z,\phi_e(Y))]}{B_{\mathrm{ref}}}.
\end{equation}
For a fixed $\alpha_e$, the hybrid error is
$(1-\alpha_e)\delta_e(z)+\alpha_e\eta_e(z)$. Hence
\begin{align}
    \operatorname{Bias}[\widehat m_{\alpha_e}^e(z)]
      &=(1-\alpha_e)\delta_e(z),\nonumber\\
    \operatorname{Var}[\widehat m_{\alpha_e}^e(z)]
      &=\alpha_e^2v_e(z),\nonumber\\
    \operatorname{MSE}[\widehat m_{\alpha_e}^e(z)]
      &=(1-\alpha_e)^2\delta_e(z)^2+\alpha_e^2v_e(z).
    \label{eq:app-hybrid-mse}
\end{align}
The cross term vanishes because the persistent approximation is fixed and
the Monte Carlo error has zero mean. This makes the trade-off precise:
pure Nystr\"om estimation has no reference-minibatch variance but can be
biased, whereas pure Monte Carlo estimation is unbiased for the empirical
reference mean. A hybrid generally retains some approximation bias.

For a fixed query distribution $Q$, let
$\mathcal B_e^2=\mathbb E_{z\sim Q}[\delta_e(z)^2]$ and
$\mathcal V_e=\mathbb E_{z\sim Q}[v_e(z)]$. Minimizing the expected squared
error gives the oracle coefficient
\begin{equation}
    \alpha_e^\star=\frac{\mathcal B_e^2}{\mathcal B_e^2+\mathcal V_e},
    \label{eq:app-oracle-alpha}
\end{equation}
when the denominator is nonzero. If both terms are positive, the oracle
mixture has lower mean-squared error than either endpoint. If both terms
vanish, every coefficient
has zero error. This expression explains the estimator's behavior; it is
not a claim that the unknown errors are measured or that this oracle
coefficient is used during training. It also concerns kernel-mean values,
not directly the error of their derivatives. Finally, unbiasedness for a
finite reference collection does not eliminate finite-data error relative
to the desired real-world distribution.

\subsection{Reference-estimator ablation}
\label{app:reference-ablation}

\begin{table}[tp]
    \centering
    \EFTableTypography
    \centering
\EFTableTypography
\captionof{table}{\textbf{Estimator ablation.} We report VBench scores and VLM judge results for different MMD reference kernel mean estimators.}
    \label{tab:reference_ablation}
    \begin{tabularx}{\linewidth}{l*{4}{N}}
        \toprule
        \rowcolor{EFTableHeader}
        & \multicolumn{3}{c}{\EFMetricGroup{VBench}} & \EFMetricGroup{VLM} \\
        \rowcolor{EFTableHeader}
        \multirow{-2}{*}{\textbf{Method}}
        & \textbf{Total} & \textbf{Qual.} & \textbf{Sem.} & \textbf{Rank} \\
        \midrule
        Nystr\"om & 84.59 & \textbf{85.47} & 81.08 &1.97\\
        MC & 84.06 & 84.84 & 80.93 &1.97\\
        \rowcolor{EFTableOurs}
        Full & \textbf{84.64} & 85.43 & \textbf{81.48}&\textbf{2.07} \\
        \bottomrule
    \end{tabularx}

\end{table}

Table~\ref{tab:reference_ablation} collects the aggregate scores reported
for the reference-side ablation. Relative to pure Monte
Carlo estimation, the hybrid improves VBench Total by 0.58 points; relative
to pure Nystr\"om estimation, it improves Total by 0.05 points. These
measurements support the combined estimator in the evaluated setting.
They do not imply that any mixture weight will outperform both endpoints,
as the error decomposition in Equation~\eqref{eq:app-hybrid-mse} makes clear.

As shown in Figure~\ref{fig:nystrom-mc-full}, the quantitative results are
also reflected in generation quality. Pure Nystr\"om estimation produces
duplicated or inconsistent liquid streams, while pure Monte Carlo estimation
leads to unstable bottle geometry and discontinuous pouring dynamics, as
highlighted by the red circles. Combining the two estimators better preserves
object structure and produces a coherent stream and smoothly expanding puddle
over time.

\begin{figure}[tp]
    \centering
    \includegraphics[width=\linewidth]{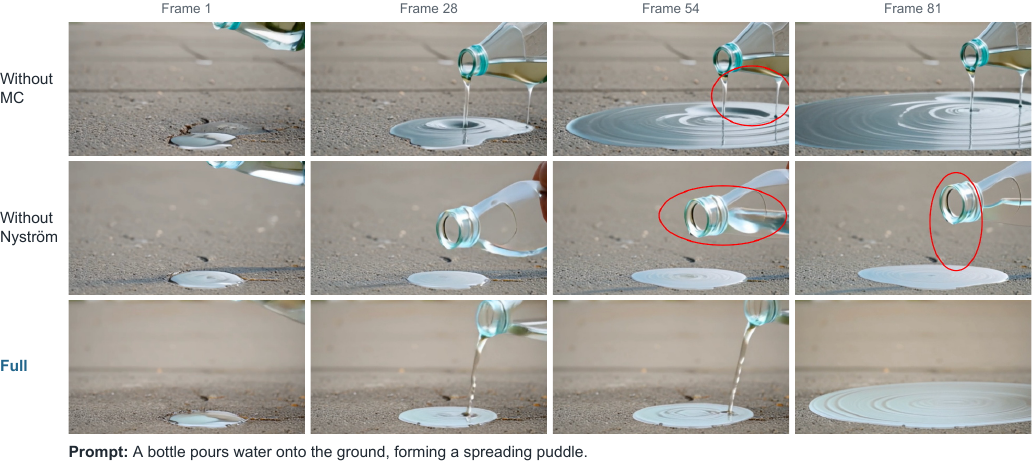}
    \caption{\textbf{Qualitative ablation of the hybrid reference estimator.}
Removing either the Monte Carlo or Nystr\"om component introduces visible
temporal and structural artifacts (red circles), while the full estimator
produces coherent pouring dynamics and stable object geometry.}
    \label{fig:nystrom-mc-full}
\end{figure}

\section{Additional Ablations}
\label{app:ablations}

\subsection{Initialization and reference data}
\label{app:initialization-data}

Table~\ref{tab:initialization_data_full} gives the available component
scores for the 1.3B configurations in Table~\ref{tab:initialization_data}.
The additional TF + UltraVideo configuration has aggregate VBench and
VLM scores of 83.68 and 4.35.
With Wan references, TF and ODE
initialization attain nearly identical VBench Totals after 150 updates,
but differ across the quality and semantic components. The VLM evaluation
favors ODE initialization in all four component scores.
Changing from Wan-generated to UltraVideo references under ODE
initialization raises VLM Total from 4.48 to 4.49, with improvements in
semantic consistency, subject consistency, and motion smoothness, but
lower visual quality (4.29 vs. 4.33) and VBench Total (83.65 vs. 84.25). Thus, comparable aggregate scores can conceal different
behavior, and reference selection should be assessed with multiple
criteria.

\begin{table}[tp]
\centering
\EFTableTypography
\captionof{table}{\textbf{Full initialization and reference-data results.}
1.3B models after 150 updates. VLM columns use independent scoring,
with Total the mean of the four component scores up to rounding.
Only completed configurations are reported.}
\label{tab:initialization_data_full}
\begin{tabularx}{\linewidth}{ll*{8}{N}}
\toprule
\rowcolor{EFTableHeader}
& & \multicolumn{3}{c}{\EFMetricGroup{VBench}}
& \multicolumn{5}{c}{\EFMetricGroup{VLM scores}} \\
\rowcolor{EFTableHeader}
\multirow{-2}{*}{\textbf{References}} & \multirow{-2}{*}{\textbf{Init.}}
& \textbf{Total} & \textbf{Qual.} & \textbf{Sem.}
& \textbf{Total} & \textbf{Qual.} & \textbf{Sem.} & \textbf{Subj.} & \textbf{Smoo.} \\
\midrule
Wan-generated & TF & 84.24 & 85.35 & 79.79 & 4.33 & 4.15 & 4.63 & 4.45 & 4.10 \\
Wan-generated & ODE & 84.25 & 85.06 & 80.99 & 4.48 & 4.33 & 4.67 & 4.67 & 4.26 \\
UltraVideo & ODE & 83.65 & 84.42 & 80.57 & 4.49 & 4.29 & 4.70 & 4.69 & 4.27 \\
\rowcolor{EFTableOurs}
UltraVideo & TF & 83.68 & 85.09 & 78.05 & 4.35 & 4.24 & 4.36 & 4.56 & 4.25 \\
\bottomrule
\end{tabularx}

\end{table}

The default UltraVideo runs match the video marginal. Their 14B VLM scores are
reported in Table~\ref{tab:vlm_comparison}; no 14B UltraVideo VBench
result is available. Table~\ref{tab:marginal_joint} compares 1.3B marginal
and joint matching with both reference sources.

\subsection{Representation choice and complementary encoders}
\label{app:encoders}
\label{app:single-encoder-vbench}

\paragraph{Qualitative behavior across representation spaces.}
Section~\ref{sec:representation_space} discusses MMD supervision in
image, video, and reconstruction spaces. Figure~\ref{fig:dino-comparison}
illustrates the effect of adding DINOv3 spatial constraints.
\begin{figure}[tp]
    \centering
    \includegraphics[width=\linewidth]{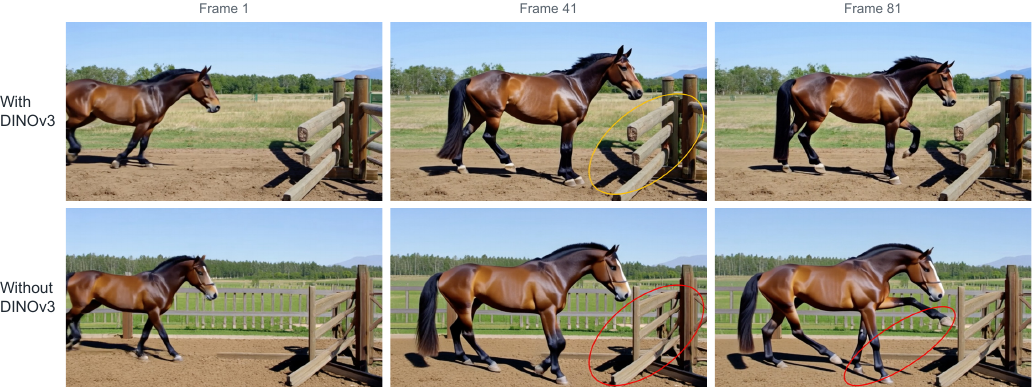}
    \caption{\textbf{Effect of incorporating DINOv3 features.}
Adding DINOv3 supervision improves perceptual quality by preserving sharper
details and more consistent object geometry. Without DINOv3, the generated
content becomes overly smooth and may exhibit structural tearing around the
horse and fence.}
    \label{fig:dino-comparison}
\end{figure}

\paragraph{Single-encoder results and demonstrations.}

Table~\ref{tab:single-encoder-vbench-full} reports the full VBench component
scores for seven single-encoder matching configurations, complementing
the aggregate comparison in Table~\ref{tab:video_encoder_ablation}.
V-JEPA~2 gives the highest aesthetic and imaging quality (66.59 and
70.21), but no encoder leads on every dimension.
Figure~\ref{fig:single-encoder-demonstrations} complements these scores with
visual examples from four individual matching encoders.

\begin{table}[tp]
\centering
\EFTableTypography
\caption{\textbf{Detailed VBench results for single-encoder matching.}
Scores are on a 0--100 scale; higher is better. The table reports all
seven quality dimensions and nine semantic dimensions for each encoder.
Bold marks the best value in each row, including ties.}
\label{tab:single-encoder-vbench-full}
\begin{tabularx}{\linewidth}{l*{7}{N}}
\toprule
\rowcolor{EFTableHeader}
\textbf{Metric}
& \makecell{\textbf{VideoMAE}\\\textbf{v2}}
& \textbf{V-JEPA~2}
& \textbf{SigLIP2}
& \makecell{\textbf{ConvNeXt-}\\\textbf{B}}
& \textbf{DINOv3}
& \makecell{\textbf{InternVideo}\\\textbf{2}}
& \makecell{\textbf{Motion}\\\textbf{former}} \\
\midrule
\rowcolor{EFTableGroup}
\multicolumn{8}{l}{\textit{Quality dimensions}} \\
Subject consistency & 95.89 & 96.28 & \textbf{97.50} & 96.05 & 97.27 & 94.86 & 97.32 \\
Background consistency & \textbf{96.75} & 95.73 & 96.44 & 94.80 & 96.21 & 93.48 & 96.67 \\
Temporal flickering & \textbf{99.57} & 99.21 & 99.06 & 99.42 & 98.64 & 98.60 & 99.44 \\
Motion smoothness & 98.45 & 98.72 & 98.67 & 97.99 & 98.50 & 95.91 & \textbf{99.16} \\
Dynamic degree & 66.94 & 58.06 & 43.61 & 49.17 & 32.22 & \textbf{80.83} & 15.00 \\
Aesthetic quality & 64.53 & \textbf{66.59} & 65.79 & 64.59 & 65.62 & 59.09 & 63.47 \\
Imaging quality & 69.45 & \textbf{70.21} & 68.49 & 67.28 & 69.63 & 63.34 & 64.81 \\
\midrule
\rowcolor{EFTableGroup}
\multicolumn{8}{l}{\textit{Semantic dimensions}} \\
Object class & 94.37 & 94.64 & 95.54 & 94.13 & 94.30 & 92.69 & \textbf{95.62} \\
Multiple objects & 83.52 & 87.33 & \textbf{87.70} & 87.01 & 86.88 & 79.25 & 79.22 \\
Human action & \textbf{96.80} & 96.00 & 96.60 & \textbf{96.80} & 96.40 & \textbf{96.80} & 96.20 \\
Color & 86.56 & \textbf{88.96} & 87.61 & 85.92 & 85.03 & 81.17 & 86.37 \\
Spatial relationship & 76.02 & 78.98 & \textbf{81.96} & 79.18 & 79.22 & 70.74 & 73.27 \\
Scene & 55.78 & 56.56 & 56.93 & 56.76 & \textbf{58.59} & 53.44 & 48.30 \\
Appearance style & 20.95 & 21.52 & 20.19 & 20.52 & 20.36 & 21.33 & \textbf{21.89} \\
Temporal style & 24.35 & 24.41 & \textbf{24.89} & 24.73 & 24.54 & 23.62 & 23.61 \\
Overall consistency & 26.50 & 26.49 & 26.57 & \textbf{26.98} & 26.62 & 26.53 & 26.15 \\
\bottomrule
\end{tabularx}

\end{table}

\begin{figure}[tp]
    \centering
    \includegraphics[width=\linewidth]{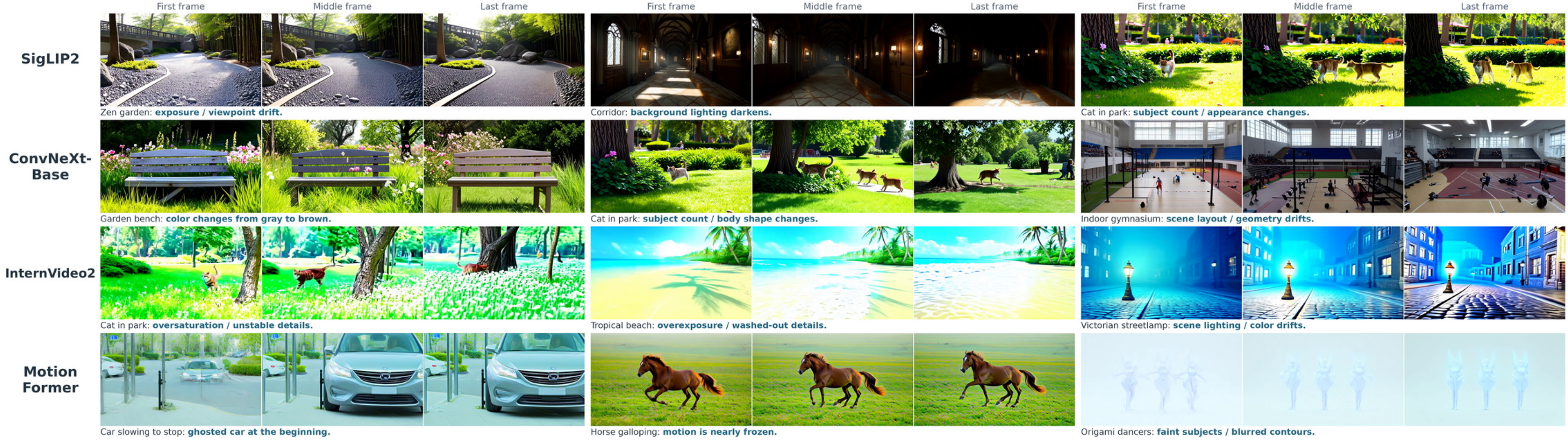}
    \caption{\textbf{Demonstrations with individual matching encoders.}
    Rows show SigLIP2, ConvNeXt-B, InternVideo2, and Motionformer,
    respectively. Each triplet shows frames labeled first, middle, and
    last from a generated video. The examples illustrate appearance and
    scene changes, color and exposure artifacts, and unnatural or limited
    motion under the tested configurations.}
    \label{fig:single-encoder-demonstrations}
\end{figure}

\paragraph{ConvNeXt-B.}
The generated videos exhibit abrupt, cut-like changes in appearance and
scene structure over time. This behavior is consistent with the weak
MMD response to temporal reordering in
Figure~\ref{fig:representation-diagnostic}(b), suggesting insufficient
temporal constraints in this matching representation.

\paragraph{InternVideo2.}
The examples show pronounced color and exposure problems, including
oversaturation, washed-out details, and drifting illumination.
Although InternVideo2 achieves the highest dynamic degree (80.83), its
color score (81.17) and imaging quality (63.34) are the lowest among
the reported encoders. Substantial motion therefore does not by itself
ensure faithful appearance.

\paragraph{Motionformer.}
The illustrated sequences show unnatural temporal evolution:
the car changes unnaturally and exhibits ghosting, while the horse and
origami sequences are almost static, with mainly local jitter.
These observations suggest inadequate temporal constraints in the
tested matching space. Dynamic degree is only 15.00, despite a motion
smoothness score of 99.16; limited motion can make outputs appear smooth
without producing natural actions.

\paragraph{SigLIP2.}
In the inspected videos, global color-tone drift is accompanied by rapid
image displacements across generated chunks, creating a jittery visual
impression. These discontinuities indicate that the tested image-feature
matching configuration does not sufficiently stabilize appearance and
motion across chunks.

\paragraph{Additional image-feature losses.}
Frame-wise image features can constrain appearance, but do not explicitly
encode temporal order or cross-frame dynamics. Figure~\ref{fig:adding-image-loss}
shows the configuration that adds SigLIP~\citep{zhai2023siglip}, Inception~\citep{szegedy2016inception}, and MAE~\citep{he2022mae} losses. The
highlighted regions exhibit changes in the street background, coastline, and model
ships across frames. One possible explanation is competition between
frame-level alignment and video-level constraints. The examples motivate
careful encoder and loss-weight selection; they do not isolate the effect
of each added image encoder.

\begin{figure}[tp]
    \centering
    \includegraphics[width=\linewidth,height=0.46\textheight,keepaspectratio]{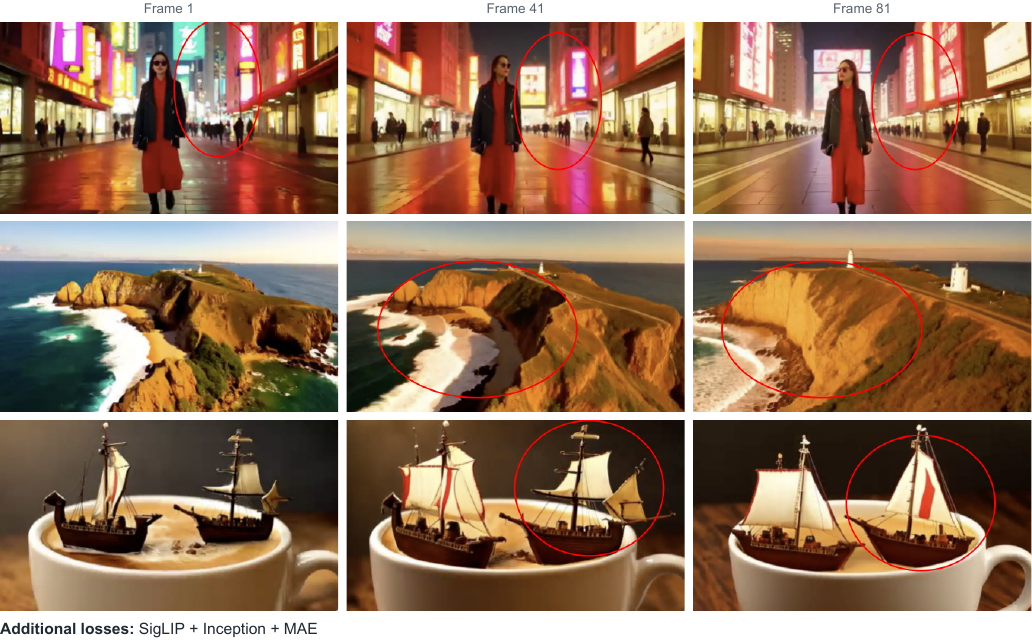}
    \caption{\textbf{Adding frame-wise representation losses.}
    Results with the combined SigLIP, Inception, and MAE losses.
    Columns show sampled frames from each example; red circles mark
    changes in subject appearance or scene geometry. These examples
    illustrate temporal inconsistencies that remain despite additional
    image-level supervision.}
    \label{fig:adding-image-loss}
\end{figure}

\paragraph{Final encoder selection.}
Together with the spatial--temporal diagnostic in
Figure~\ref{fig:representation-diagnostic}(b) and the combination ablation
in Table~\ref{tab:video_encoder_ablation}, these qualitative results
motivate our final choice of DINOv3, V-JEPA~2, and VideoMAE~v2.
DINOv3 supplies strong spatial discrimination, while V-JEPA~2 and
VideoMAE~v2 provide complementary temporal constraints.

\subsection{Controlled spatial and temporal distribution diagnostics}
\label{app:representation_diagnostic}
\label{app:representation-diagnostic}

\paragraph{Videos and perturbations.}
Figure~\ref{fig:representation-diagnostic}(b) uses three disjoint subsets of
the same 8,000-clip natural-video pool: 128 query clips, 256 reference
clips, and 32 bandwidth-calibration clips. Selection follows a fixed
hash-based ordering established before feature extraction. Each clip
contains 81 RGB frames, downsampled from $832\times480$ to $416\times240$
before each encoder's native preprocessing. The spatial perturbation is
a smooth nonrigid warp, obtained by bicubic upsampling of a $5\times7$
Gaussian displacement field with 6-pixel RMS per coordinate. The same
field is applied to every frame. Its rigid-translation control is
calibrated per clip to match RGB mean squared error without using encoder
features.

The displayed temporal permutation changes 62 of 81 frame positions.
It preserves the set of frames sampled by every encoder and every
native segment, and additionally preserves four groups of five sampled
image-encoder frames. This permutation was originally constructed as a
structural control for a temporally pooled DINO readout, hence its
protocol identifier \texttt{poolnull}. Figure~\ref{fig:representation-diagnostic}(b)
uses this training readout: average corresponding DINO patches within
each five-frame group, then concatenate the four groups in temporal
order. The displayed permutation preserves each group's membership, so
the DINO features are invariant up to numerical precision.
SigLIP2 and an additional unpooled DINO control retain all ordered
frame-patch features; their features do change under reordering.
The displayed pooled-DINO result consequently diagnoses a specific
readout ambiguity, rather than a general inability of DINO to distinguish
all temporal permutations. A second,
less constrained permutation preserves the native sampled frame sets
without preserving DINO's pooling groups and moves 64 of 81 frames.
The encoders sample different temporal windows, so they need not see
identical fractions of moved frames.
The right-hand gray bars in Figure~\ref{fig:representation-diagnostic}(b)
come from a separate rigid translation matched to per-frame
$\pm6$-pixel jitter. They are \emph{not} matched to the temporal
permutation and do not establish temporal selectivity at equal RGB error.

\paragraph{Feature readouts and MMD.}
Table~\ref{tab:diagnostic-readouts} specifies the diagnostic readouts.
DINO retains all spatial patch locations after local temporal averaging,
SigLIP2 concatenates all selected frame-patch features without averaging,
and the video encoders retain their original ordered token selections.
All features are converted to FP32 for distance computation. Backbone
inference uses FP32; TF32 is disabled. For each encoder $e$, we set $\sigma_e^2$ to
the median pairwise squared distance on the 32 clean calibration clips
and keep this bandwidth fixed across all perturbations. Thus the
calibration rule is shared, while the numerical bandwidths differ.
For query set $X$, reference set $R$, and perturbation $T$, we report
\begin{equation}
\Delta_e(T)=
\widehat{\operatorname{MMD}}_{U,k_e}^{\,2}
  \bigl(\phi_e(TX),\phi_e(R)\bigr)
-\widehat{\operatorname{MMD}}_{U,k_e}^{\,2}
  \bigl(\phi_e(X),\phi_e(R)\bigr).
\label{eq:diagnostic_mmd_change}
\end{equation}
The estimator omits within-set diagonal terms and includes all
query--reference pairs. A positive change indicates that the perturbed
set is farther from the natural reference set under this kernel.
Signed estimates are retained. This measures distributional separation,
which differs from the paired displacement
$\|\phi_e(Tx)-\phi_e(x)\|_2^2/\sigma_e^2$ of an individual clip.

\begin{table}[tp]
    \centering
    \EFTableTypography
    \caption{\textbf{Frozen readouts in the representation diagnostic.}
    Frame indices are zero-based. DINO pooled is labeled DINOv3 in
    Figure~\ref{fig:representation-diagnostic}(b); the unpooled variant and
    Wan VAE are additional checks. Layers follow the diagnostic code's
    indexing.}
    \label{tab:diagnostic-readouts}
    \begin{tabularx}{\linewidth}{lX}
        \toprule
        \rowcolor{EFTableHeader}
        \textbf{Representation} & \textbf{Readout} \\
        \midrule
        Pixels & All 81 RGB frames; no learned projection. \\
        Wan VAE & Wan2.1 VAE latents from all 81 frames. \\
        DINOv3 ordered & ViT-L final normalized patch features from 20 sampled frames;
            256 tokens per frame, concatenated in frame order into 5,120 tokens;
            an additional control. \\
        SigLIP2 & Large patch16-256 final post-layernorm features from the same
            20 frames; all 256 patch tokens per frame concatenated in order;
            pooling head unused~\citep{tschannen2025siglip2}. \\
        DINOv3 pooled & The same DINO frame features averaged within four
            consecutive groups of five sampled frames and concatenated into
            1,024 tokens; the training readout used in the main figure. \\
        Motionformer & Something--Something V2 Base 16$\times$4 checkpoint;
            16 sampled frames, block 11 before final layer normalization,
            all 1,568 patch tokens. \\
        V-JEPA~2 & ViT-L block 15; frames 17--80; 1,024 uniformly selected tokens. \\
        VideoMAE~v2 & Large block 15; five 16-frame segments covering frames 1--80;
            256 tokens per segment. \\
        \bottomrule
    \end{tabularx}
\end{table}

\paragraph{Uncertainty and interpretation.}
We use 4,000 paired source-clip bootstrap resamples of the query videos
and independent resamples of the references, sharing resampling draws
across encoders and conditions. Error bars are basic 95\% intervals
conditional on the calibration bandwidth, with no multiple-comparison
correction. Source-film identifiers are unavailable, so the resampling
unit is a clip. The intervals are descriptive uncertainty estimates;
near-zero U-MMD is degenerate, and these intervals are not an exact
test of the null. The experiment also evaluates $0.5\sigma_e$ and
$2\sigma_e$ without tuning bandwidth to individual corruptions.

The pooled DINO training readout has a spatial-warp MMD change of
$0.00612$. Under the displayed temporal permutation, its MMD change is
$-2.95\times10^{-10}$ and its mean normalized paired feature displacement
is $1.90\times10^{-15}$, consistent with the pooling invariance.
For the same permutation, the unpooled DINO control and SigLIP2 have
MMD changes of $2.97\times10^{-7}$ and $2.85\times10^{-6}$, respectively,
with intervals including zero. Their mean normalized paired feature
displacements are $0.0501$ and $0.1526$, confirming that the representations
are not invariant. Both respond positively to the spatial warp, with
MMD changes of $0.00549$ and $0.00255$. These measurements separate
spatial distribution discrimination from merely retaining ordered frames.
The unpooled control shows that a weak distribution-level response can
persist even when frame order is explicitly retained. Neither diagnostic
alone isolates the cause of the generation ablation's training behavior.

The less constrained temporal permutation gives MMD changes of
$0.01034$ and $0.00445$ for V-JEPA~2 and VideoMAE~v2, respectively;
their reported intervals remain above
zero at all three bandwidths. The pooled DINO readout has a small positive
response ($4.91\times10^{-5}$) to this permutation, unlike its numerical invariance in the
pooling-preserving control. Motionformer likewise has a mean normalized paired
displacement of $0.175$ under temporal reordering, yet its MMD change
is $-0.000178$ with interval $[-0.000459,0.000133]$. Thus changing features
does not ensure that the chosen finite-sample kernel discrepancy detects
a distribution shift.

Wan VAE also has a positive temporal MMD response ($0.00978$ for the
less constrained permutation), despite its unsuccessful generation
trials. This shows why diagnostic
sensitivity alone cannot predict optimization success. The diagnostic
fixes specific layers and readouts.
Our final representation choice consequently uses both the perturbation
study and the generation ablations, without treating the bar heights as
a universal ranking of perceptual quality or training performance.

\subsection{Joint text--video matching}
\label{app:joint-matching}

Let $c$ denote the text condition. Video-only matching uses $p_\theta(x)$
and $p_{\rm ref}(x)$, whereas joint matching compares representations
of $(x,c)$ from $p_\theta(x,c)=p(c)p_\theta(x\mid c)$ and the reference
joint distribution.

\paragraph{Full 1.3B comparison.}
Table~\ref{tab:marginal_joint_full} expands the main-paper comparison in
Table~\ref{tab:marginal_joint} with all VLM component scores.
With UltraVideo references, joint matching lowers all four VLM
components despite a small increase in VBench Total. With Wan-generated
references, joint matching lowers all four VLM component scores and all
three VBench aggregates. Independent VLM scoring therefore favors marginal
matching for both reference sources. The groupwise ranks in
Table~\ref{tab:marginal_joint} instead favor joint matching for Wan,
illustrating that the two evaluation protocols need not agree.

\begin{table}[tp]
\centering
\EFTableTypography
\captionof{table}{\textbf{Full marginal versus joint distribution results.}
1.3B models with ODE initialization. VBench and score-based VLM results
are reported separately for UltraVideo and Wan-generated references.
Qual.: visual quality; Sem.: semantic consistency; Subj.: subject
consistency; Smoo.: motion smoothness. Higher is better.
Table~\ref{tab:marginal_joint} reports the compact main-paper comparison.}
\label{tab:marginal_joint_full}
\begin{tabularx}{\linewidth}{l*{8}{N}}
\toprule
\rowcolor{EFTableHeader}
& \multicolumn{3}{c}{\EFMetricGroup{VBench}}
& \multicolumn{5}{c}{\EFMetricGroup{VLM scores}} \\
\rowcolor{EFTableHeader}
\multirow{-2}{*}{\textbf{Distribution}}
& \textbf{Total} & \textbf{Qual.} & \textbf{Sem.}
& \textbf{Total} & \textbf{Qual.} & \textbf{Sem.}
& \textbf{Subj.} & \textbf{Smoo.} \\
\midrule
\rowcolor{EFTableGroup}
\multicolumn{9}{l}{\textit{UltraVideo references}} \\
\rowcolor{EFTableOurs}
Marginal & 83.65 & 84.42 & 80.57 & 4.49 & 4.29 & 4.70 & 4.69 & 4.27 \\
Joint & 83.74 & 84.53 & 80.57 & 4.38 & 4.20 & 4.51 & 4.59 & 4.22 \\
\midrule
\rowcolor{EFTableGroup}
\multicolumn{9}{l}{\textit{Wan-generated references}} \\
\rowcolor{EFTableOurs}
Marginal & 84.25 & 85.06 & 80.99 & 4.48 & 4.33 & 4.67 & 4.67 & 4.26 \\
Joint & 83.90 & 84.72 & 80.62 & 4.43 & 4.29 & 4.63 & 4.57 & 4.23 \\
\bottomrule
\end{tabularx}

\end{table}

\paragraph{Full 14B comparison.}
Table~\ref{tab:marginal_joint_14b} extends the comparison to 14B models
trained with UltraVideo references for 80 updates. Both configurations
use the same 100-prompt evaluation set, GPT-6 evaluator, and independent
scoring prompt. The marginal configuration corresponds to EF (UV) in
Table~\ref{tab:vlm_comparison}. It achieves a higher VLM Total than joint
text--video matching (4.60 versus 4.51), with higher scores in all four
components. This result favors video-marginal matching in the tested
14B UltraVideo setting. Both generators remain text-conditioned;
the comparison concerns the distribution-matching loss.

\begin{table}[tp]
\centering
\EFTableTypography
\caption{\textbf{Full 14B marginal versus joint matching comparison.}
Models use UltraVideo references and 80 training updates. Independent VLM
scores are on a 0--5 scale; higher is better. Total is the equal-weight
mean of the four component scores, up to rounding.
Bold marks the best value in each column.}
\label{tab:marginal_joint_14b}
\begin{tabularx}{\linewidth}{l*{5}{N}}
\toprule
\rowcolor{EFTableHeader}
& \multicolumn{5}{c}{\EFMetricGroup{VLM scores}} \\
\rowcolor{EFTableHeader}
\multirow{-2}{*}{\textbf{Distribution}}
& \textbf{Total} & \textbf{Qual.} & \textbf{Sem.}
& \textbf{Subj.} & \textbf{Smoo.} \\
\midrule
\rowcolor{EFTableOurs}
Marginal (video-only)
& \textbf{4.60} & \textbf{4.46} & \textbf{4.72}
& \textbf{4.82} & \textbf{4.38} \\
Joint (text--video)
& 4.51 & 4.36 & 4.66 & 4.68 & 4.35 \\
\bottomrule
\end{tabularx}

\end{table}

\subsection{Why use MMD for dense video representations?}
\label{app:why-mmd}

A Fr\'echet-distance (FD) objective summarizes a $d$-dimensional
representation with a mean and a $d\times d$ covariance matrix~\citep{heusel2017fid,unterthiner2018fvd,yang2026frechet}. Dense
covariance storage costs $O(d^2)$ and matrix factorizations can cost
$O(d^3)$, which is restrictive for representations retaining many video
tokens. Pooling reduces this cost but also removes information the loss
could constrain.

MMD uses pairwise kernels without constructing the covariance matrix.
The Nystr\"om summary reduces reference comparisons, and selective replay
controls generator backpropagation. Direct generated--generated evaluation
still costs $O(B^2)$ kernel evaluations, and representation and kernel
choices remain consequential. Our exploratory sparse-feature FD experiment
produced blurred samples (Figure~\ref{fig:app-fd}), motivating dense
feature matching. This observation concerns the tested configuration,
rather than all FD formulations.

\begin{figure}[tp]
    \centering
    \includegraphics[width=\linewidth]{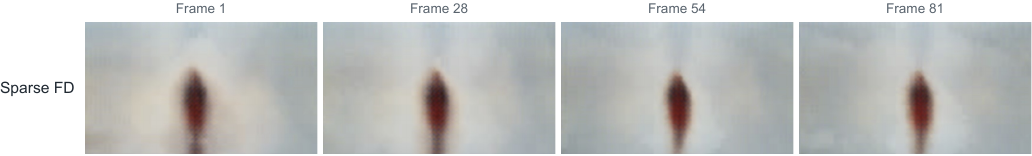}
    \caption{\textbf{Exploratory sparse-feature FD objective.} Sampled
    frames from the tested configuration show a blurred human silhouette
    with little background detail. This example motivates preserving dense
    video features but is not a comparison of all FD-based objectives.}
    \label{fig:app-fd}
\end{figure}

\subsection{Fresh rollouts versus a FIFO feature queue}
\label{app:fifo}

\begin{figure}[tp]
    \centering

    \includegraphics[width=0.8\linewidth,height=0.48\textheight,keepaspectratio]{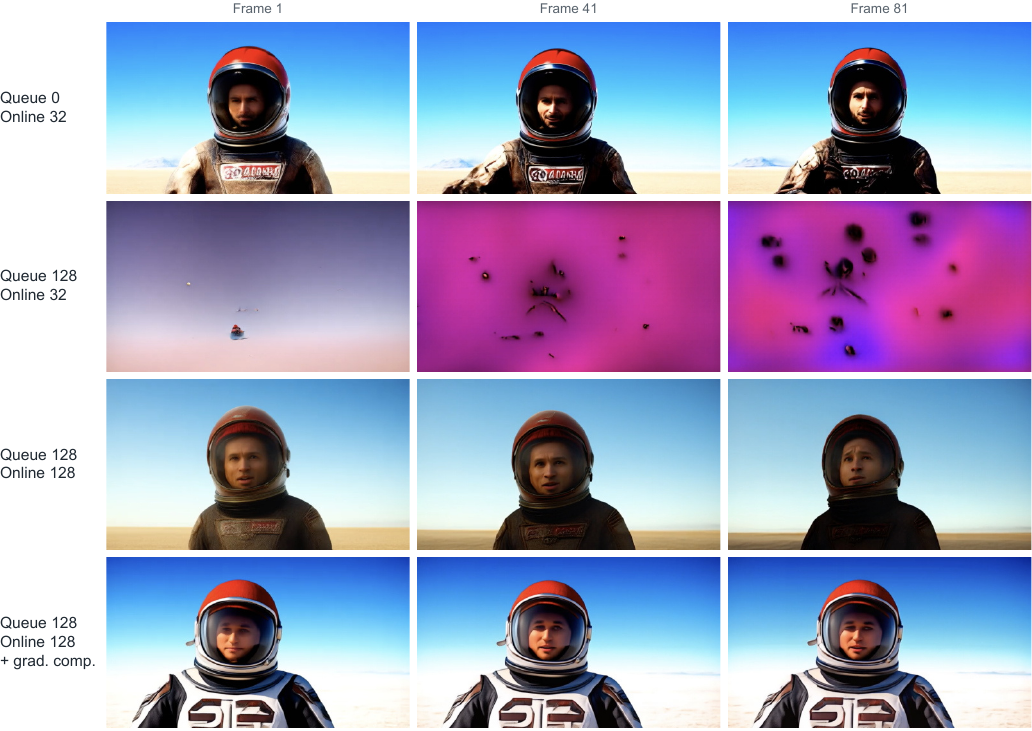}
    \caption{\textbf{FIFO feature reuse.} More online samples and gradient compensation improve queued variants, but retain historical features. The top row uses 32 fresh samples without a queue; ``grad. comp.'' denotes gradient compensation.}
    \label{fig:fifo_ablation}
\end{figure}

Selective differentiation uses current rollouts even when only a subset
is backpropagated. Reusing historical representations in a FIFO queue~\citep{yang2026frechet}
could save forward computation, but those features come from earlier
generator parameters. This feature queue is distinct from the KV cache
used for autoregressive generation.

Figure~\ref{fig:fifo_ablation} compares four settings: no queue with 32
online samples; a queue of 128 with 32 online samples; a queue of 128
with 128 online samples; and the last setting with gradient compensation.
With 32 online samples, the queue severely degrades the example.
Increasing the online population restores the subject and scene but
leaves softer details; compensation improves boundaries and appearance.

Queued features repel current samples, but their own reference-attraction
terms have no parameter-gradient path. Compensation can address part of
this asymmetry, but cannot make stale features on-policy or establish
Equation~\eqref{eq:app-unbiased-gradient}. The example does not isolate
staleness from gradient-scaling effects. We therefore use fresh rollouts
for the standard configuration.

\Needspace{0.42\textheight}
\subsection{Comparison under a matched training budget}
\label{app:matched-training-budget}

Table~\ref{tab:matched-training-budget} complements the equal-update
ablation in Table~\ref{tab:sample_ablation} with an approximately matched
training-compute comparison. We compare full differentiation
($B=b=256$) after 90 updates with our reported EF (Wan) configuration
($B=256$, $b=128$) after 150 updates. Selective differentiation achieves
higher VBench Total (84.25 versus 83.89) and VLM Total (4.48 versus 4.35),
as well as higher scores in all four VLM components. Full differentiation
has a slightly higher VBench Semantic score (81.05 versus 80.99).
These results favor the selective-differentiation configuration within
this training budget.

\begin{table}[H]
\centering
\EFTableTypography
\caption{\textbf{Full and selective differentiation under a matched training budget.}
$B$ denotes the generated rollouts used for MMD estimation and $b$ the
rollouts differentiated per update. Update counts approximately match
training compute. The $b=128$ row reproduces the reported EF (Wan) scores
in Table~\ref{tab:short-video-comparison}. VLM columns use independent
scoring, with Total the equal-weight mean of the four component scores
up to rounding. Bold marks the better score in each column.}
\label{tab:matched-training-budget}
\begin{tabularx}{\linewidth}{*{11}{N}}
\toprule
\rowcolor{EFTableHeader}
& & & \multicolumn{3}{c}{\EFMetricGroup{VBench}}
& \multicolumn{5}{c}{\EFMetricGroup{VLM scores}} \\
\rowcolor{EFTableHeader}
\multirow{-2}{*}{$\boldsymbol{B}$}
& \multirow{-2}{*}{$\boldsymbol{b}$}
& \multirow{-2}{*}{\textbf{Updates}}
& \textbf{Total} & \textbf{Qual.} & \textbf{Sem.}
& \textbf{Total} & \textbf{Qual.} & \textbf{Sem.}
& \textbf{Subj.} & \textbf{Smoo.} \\
\midrule
256 & 256 & 90 & 83.89 & 84.60 & \textbf{81.05}
& 4.35 & 4.22 & 4.50 & 4.48 & 4.21 \\
\rowcolor{EFTableOurs}
256 & 128 & 150 & \textbf{84.25} & \textbf{85.06} & 80.99
& \textbf{4.48} & \textbf{4.33} & \textbf{4.67}
& \textbf{4.67} & \textbf{4.26} \\
\bottomrule
\end{tabularx}

\end{table}

\section{Reference-Data Adaptation and Initialization}
\label{app:adaptation}

\subsection{Appearance, character, and spatial adaptation}
\label{app:specialized}

\paragraph{Adaptation to specialized reference data.}
Table~\ref{tab:specialized_video_evaluation} expands the three adaptation
settings in Section~\ref{sec:adaptation}. Elastic Forcing uses a
1.3B generator in these experiments; the reference videos change the
optimization target without requiring a target-specific diffusion teacher.
For character acquisition, videos containing Nailong are mixed with the
general-domain reference collection. The spatial experiment tests adaptation to a specified visual projection;
its score measures adherence to that appearance rather than verified
three-dimensional geometry.

\begin{table}[tp]
    \centering
    \EFTableTypography
    \caption{\textbf{Specialized reference-data adaptation.} Ratings are on
    a 0--10 scale; higher is better. TS: task-specific score; VQ: visual
    quality; TC: temporal consistency; SI: subject integrity.
    Bold and underlining denote the best and second-best score within
    each task, respectively.}
    \label{tab:specialized_video_evaluation}
    \label{tab:beyond_distillation}
    \begin{tabularx}{\linewidth}{l*{5}{N}}
        \toprule
        \rowcolor{EFTableHeader}
        \textbf{Method} & \textbf{TS} & \textbf{VQ} & \textbf{TC}
        & \textbf{SI} & \textbf{Overall} \\
        \midrule
        \rowcolor{EFTableGroup}
        \multicolumn{6}{l}{\textit{Monochrome appearance}} \\
        Self-Forcing & 8.822 & 8.340 & 8.140 & 8.106 & 8.308 \\
        Wan2.1-1.3B & 8.674 & 8.390 & 8.410 & 8.372 & 8.382 \\
        Wan2.1-14B & \underline{9.840} & \textbf{8.638} & \textbf{8.574} & \textbf{8.494} & \textbf{8.992} \\
        \rowcolor{EFTableOurs}
        Elastic Forcing & \textbf{9.910} & \underline{8.544} & \underline{8.526} & \underline{8.426} & \underline{8.942} \\
        \midrule
        \rowcolor{EFTableGroup}
        \multicolumn{6}{l}{\textit{Nailong character}} \\
        Self-Forcing & 7.532 & \underline{8.886} & 8.512 & \underline{8.220} & 8.122 \\
        Wan2.1-1.3B & 7.418 & 8.588 & 8.056 & 7.922 & 7.878 \\
        Wan2.1-14B & \underline{7.656} & \textbf{8.906} & \underline{8.580} & 8.084 & \underline{8.174} \\
        \rowcolor{EFTableOurs}
        Elastic Forcing & \textbf{8.072} & 8.834 & \textbf{8.584} & \textbf{8.272} & \textbf{8.380} \\
        \midrule
        \rowcolor{EFTableGroup}
        \multicolumn{6}{l}{\textit{Spatial appearance}} \\
        Self-Forcing & 5.000 & \underline{7.910} & 7.934 & 7.832 & 6.272 \\
        Wan2.1-1.3B & 3.990 & 7.816 & 8.192 & 7.966 & 5.674 \\
        Wan2.1-14B & \underline{6.334} & \textbf{8.072} & \underline{8.332} & \underline{8.110} & \underline{7.130} \\
        \rowcolor{EFTableOurs}
        Elastic Forcing & \textbf{8.544} & 7.896 & \textbf{8.664} & \textbf{8.396} & \textbf{8.292} \\
        \bottomrule
    \end{tabularx}
\end{table}

For monochrome appearance, Elastic Forcing has the highest task-specific
score (9.910), but Wan2.1-14B has the highest overall score (8.992 versus
8.942). This distinguishes target-style compliance from general quality.
For Nailong, Elastic Forcing leads in character fidelity and overall score,
while Wan2.1-14B retains the highest visual-quality rating. For spatial
appearance, Elastic Forcing improves the task-specific score from 5.000
for Self-Forcing to 8.544 and the overall score from 6.272 to 8.292.
Together, these results show adaptation to different reference-defined
targets, while retaining meaningful variation across quality dimensions.

\paragraph{General-domain retention}
Fine-tuning large pretrained generators on small, specialized datasets
can compromise general-domain performance. To assess this trade-off for
MMD-based adaptation, we evaluate the Nailong-adapted model on VBench
using the same protocol as Self-Forcing
(Table~\ref{tab:nailong-retention}).

\begin{table}[tp]
    \centering
    \EFTableTypography
    \caption{\textbf{General-domain retention after Nailong adaptation.}
    Matched VBench re-evaluation of 1.3B models. Scores are on a 0--100
    scale; higher is better.}
    \label{tab:nailong-retention}
    \begin{tabularx}{\linewidth}{lNNN}
        \toprule
        \rowcolor{EFTableHeader}
        & \multicolumn{3}{c}{\EFMetricGroup{VBench}} \\
        \rowcolor{EFTableHeader}
        \multirow{-2}{*}{\textbf{Model}}
        & \textbf{Total} & \textbf{Qual.} & \textbf{Sem.} \\
        \midrule
        Self-Forcing & 83.95 & 84.73 & 80.83 \\
        \rowcolor{EFTableOurs}
        Elastic Forcing after Nailong adaptation & 83.68 & 84.74 & 79.43 \\
        \bottomrule
    \end{tabularx}
\end{table}

The MMD-adapted model retains overall VBench performance comparable to
Self-Forcing while acquiring the target character, with Total scores of
83.68 and 83.95, respectively. Its Quality score is nearly identical to
that of Self-Forcing (84.74 vs.\ 84.73), while its Semantic score is lower
(79.43 vs.\ 80.83).
Figure~\ref{fig:reference-adaptation} also uses prompts unseen during
adaptation, providing evidence beyond reproducing the training clips.

The key capability demonstrated here is learning directly from
target-domain references. Standard DMD with a frozen diffusion teacher
remains tied to the teacher's distribution, whereas Elastic Forcing can
learn a new target distribution from reference videos without adapting
the diffusion teacher. These experiments demonstrate reference-driven
adaptation; they do not establish an advantage over methods given equal
access to target-domain data.

\begin{figure}[H]
    \centering
    \includegraphics[width=0.8\linewidth]{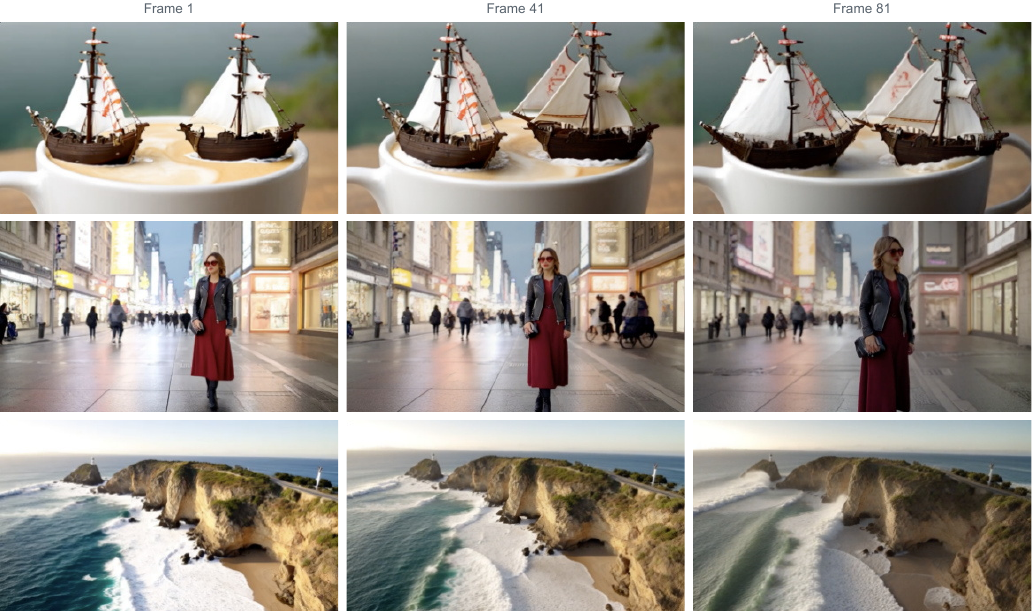}
    \caption{\textbf{Real-video reference data.} Generated examples after
    replacing Wan-generated references with OpenVid videos. Each row
    shows sampled frames from one generation. The examples span object,
    human, and landscape content and illustrate learning from a
    reference source independent of the initialization model's samples.}
    \label{fig:app-openvid}
\end{figure}

\subsection{Learning from real-video references}
\label{app:openvid}

We report the results on UltraVideo in the main paper. We also try to replace the Wan-generated reference collection with real-world videos
sampled from OpenVid~\citep{nan2025openvid}. Figure~\ref{fig:app-openvid} shows examples of model
ships, a person in a city street, and a coastal landscape. The examples
retain recognizable subjects and scene structure across the displayed
frames, indicating that synthetic teacher-generated references are not a
requirement of the objective. This is qualitative evidence of feasibility;
without a matched quantitative comparison, it does not establish equal
performance across the two reference sources.

\Needspace{0.90\textheight}
\begin{figure}[H]
    \centering
    \includegraphics[width=\linewidth,height=0.63\textheight,keepaspectratio]{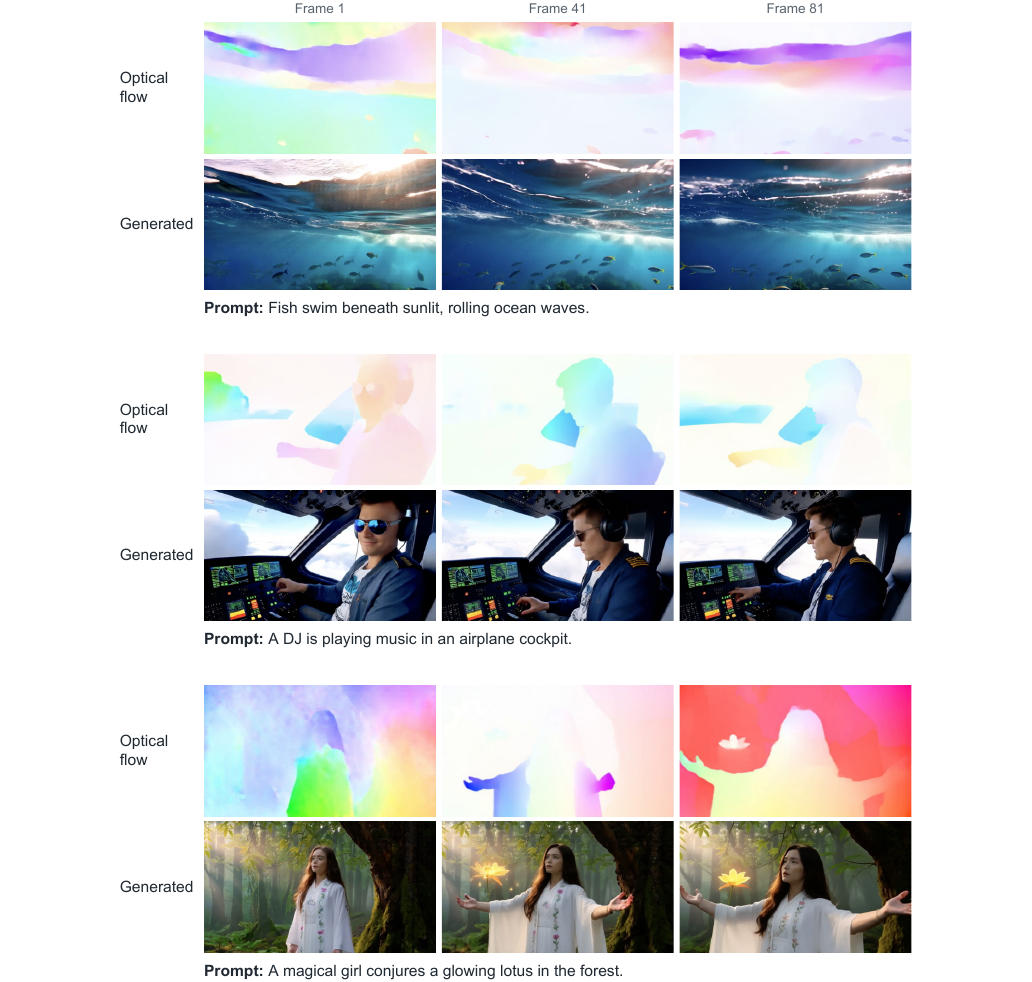}
    \caption{\textbf{VACE adaptation with visual conditioning.}
    Within each example, optical-flow conditions appear in the top row
    and sampled output frames in the bottom row. Training uses MMD with
    an auxiliary regression objective followed
    by MMD alone, with first-frame and flow conditioning and
    no separate ODE initialization stage.}
    \label{fig:vace}
\end{figure}

\subsection{Visually conditioned generation with pretrained checkpoint}
\label{app:vace}

Our standard T2V backbones rely on ODE or TF initialization before few-step
autoregressive post-training. This initialization bridges both a sampling
change, from a longer denoising trajectory to four steps, and a structural
change, from bidirectional generation to causal chunks. In experiments
omitting this initialization, individual chunks can remain visually
plausible while continuity across chunks deteriorates. Local frame quality
therefore does not by itself establish coherent streaming generation.

We examine whether stronger visual conditioning can ease this transition
using Wan2.1-14B VACE~\citep{jiang2025vace} with first-frame and optical-flow conditioning. The
first frame constrains appearance, and optical flow supplies motion
information. Starting directly from the pretrained VACE checkpoint, we
first train with MMD and an auxiliary regression objective, then continue
with MMD alone. This omits a \emph{separate} ODE or TF initialization stage.

Figure~\ref{fig:vace} pairs optical-flow inputs with frames from generated
videos of fish beneath waves, a DJ in a cockpit, and a person producing
a glowing lotus. The examples retain recognizable appearance and scene
structure over the displayed frames, supporting the feasibility of this
conditioned setting. They do not isolate the contributions of visual
conditioning and the initial regression phase, nor imply that standard
text-only generation can dispense with initialization.

\Needspace{0.88\textheight}
\begin{figure}[H]
    \centering
    \includegraphics[width=\linewidth,height=0.63\textheight,keepaspectratio]{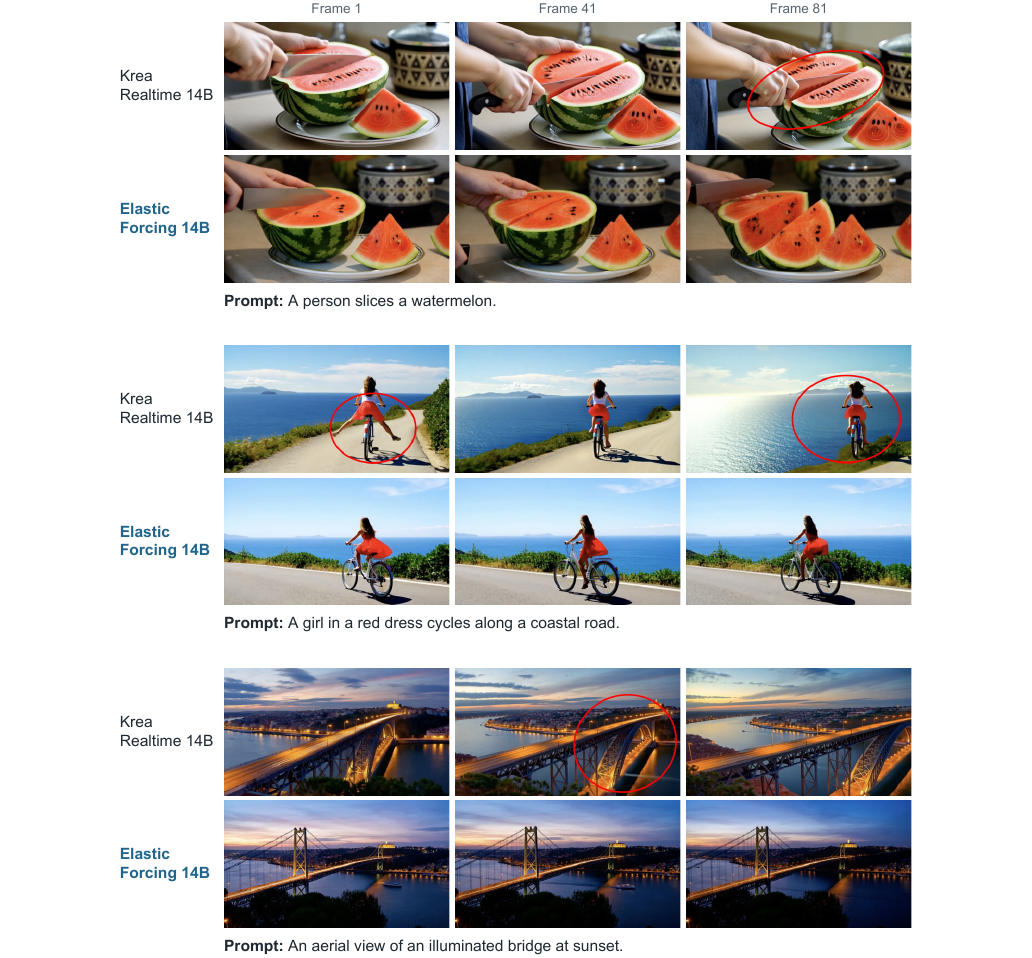}
    \caption{\textbf{Additional 14B comparisons.} Each example compares
    Krea Realtime (top) and Elastic Forcing (bottom), with its prompt below and sampled
    frames arranged from left to right. Red circles highlight
    inconsistencies in object interaction, rider--bicycle interaction, and bridge
    geometry in the selected examples.}
    \label{fig:app-krea-comparison}
\end{figure}

\section{Additional Model Comparisons}
\label{app:large-model-examples}

Figure~\ref{fig:app-krea-comparison} supplements the 14B evaluation in
Table~\ref{tab:vlm_comparison}. Both methods use 14B backbones, and our
post-training starts from the same initialization used for the comparison.
The first example tests the progression of a cutting action and the
integrity of the watermelon. The second tests bicycle and rider
consistency during motion, and the third tests the stability of bridge
geometry across viewpoints. The highlighted Krea Realtime frames exhibit
local inconsistencies, while the displayed Elastic Forcing frames better
preserve the corresponding objects and structures. These selected
examples illustrate the aggregate comparison; they do not quantify the
frequency of failures over all prompts.

\Needspace{0.52\textheight}
\begin{figure}[H]
    \centering
    \includegraphics[width=0.9\linewidth]{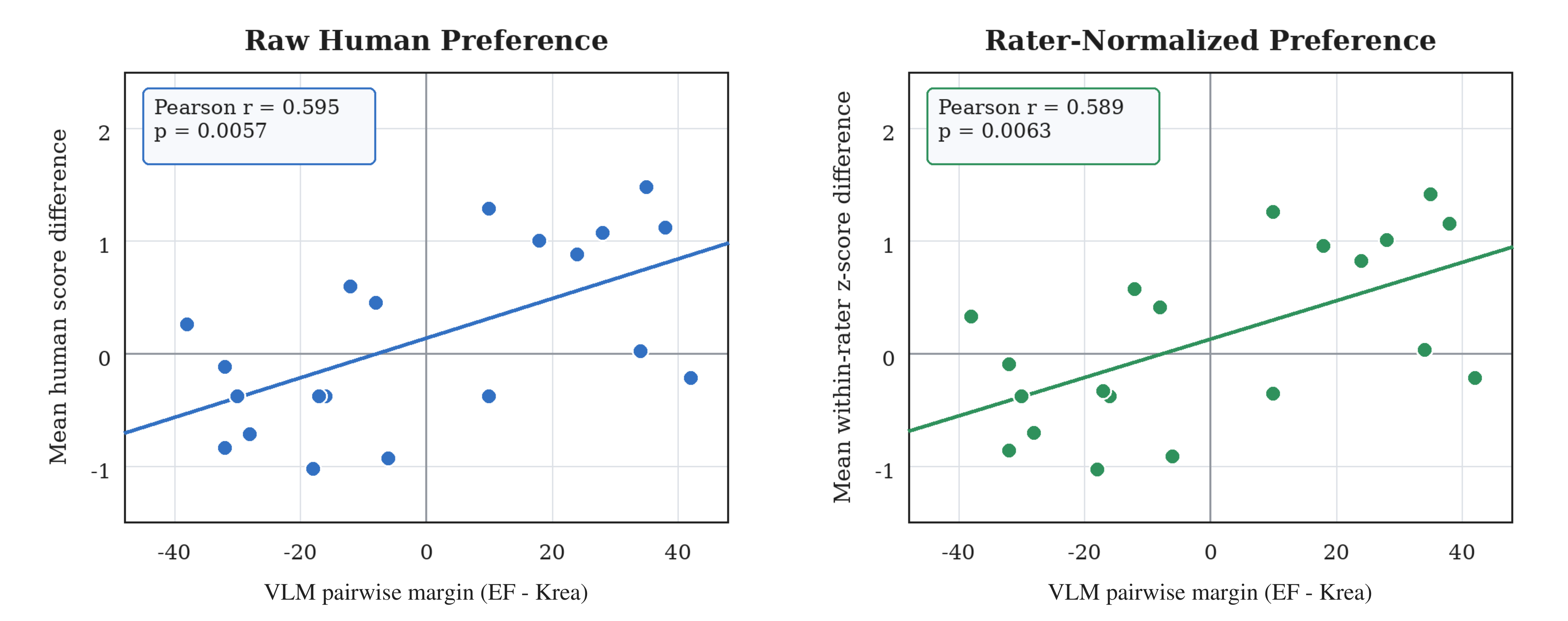}
    \caption{\textbf{Strong agreement between pairwise VLM and human judgments.}
    VLM preference margins are compared with human preference differences
    over 20 matched prompts and 42 raters from the scored study in
    Table~\ref{tab:additional-wan-evaluation}. The left panel uses raw human scores,
    while the right panel uses within-rater normalized scores. Positive values
    favor EF-14B (Wan), and solid lines indicate least-squares fits.}
    \label{fig:vlm-human-correlation}
\end{figure}

\section{Analysis of VLM Evaluation}
\label{sec:vlm-evaluation-analysis}

\subsection{Correlation with Human Evaluation}
\label{sec:vlm-human-correlation}

This analysis uses the scored human evaluation of EF-14B (Wan) and Krea
Realtime reported in Table~\ref{tab:additional-wan-evaluation}. The
1.3B and 14B human comparisons in the main paper
(Figure~\ref{fig:teaser}) are separate preference-voting experiments
that collect categorical choices without numerical video scores. We
therefore use the scored study to quantify the correlation between
VLM judgments and human preferences.

We conduct a blinded pairwise correlation analysis over 20 matched prompts. For each
prompt, the VLM jointly observes two anonymized videos through 24 uniformly
sampled frames per video, aligned by normalized time. It then predicts a signed
pairwise preference margin $d_i^{\mathrm{VLM}}$, where a positive value indicates
a preference for EF-14B (Wan) over Krea Realtime 14B.

For each prompt $i$, the corresponding human preference difference is aggregated
over all $R=42$ raters:
\begin{equation}
d_i^{\mathrm{Human}}
=
\frac{1}{R}
\sum_{j=1}^{R}
\left(
h_{ij}^{\mathrm{Ours}}
-
h_{ij}^{\mathrm{Krea}}
\right),
\qquad R=42,
\label{eq:human-pairwise-difference}
\end{equation}
where $h_{ij}^{\mathrm{Ours}}$ and $h_{ij}^{\mathrm{Krea}}$ denote the scores
assigned by rater $j$ to the EF-14B (Wan) and Krea Realtime videos,
respectively, associated with prompt $i$.
The 42 ratings are first aggregated within each prompt, and the correlation is
then computed over the resulting 20 prompt-level differences. This avoids
treating individual ratings of the same video pair as independent observations.

We quantify VLM--human agreement using the Pearson correlation coefficient:
\begin{equation}
r
=
\frac{
\sum_{i=1}^{n}
\left(d_i^{\mathrm{VLM}}-\bar{d}^{\mathrm{VLM}}\right)
\left(d_i^{\mathrm{Human}}-\bar{d}^{\mathrm{Human}}\right)
}{
\sqrt{
\sum_{i=1}^{n}
\left(d_i^{\mathrm{VLM}}-\bar{d}^{\mathrm{VLM}}\right)^2
}
\sqrt{
\sum_{i=1}^{n}
\left(d_i^{\mathrm{Human}}-\bar{d}^{\mathrm{Human}}\right)^2
}
},
\qquad n=20.
\label{eq:pairwise-pearson}
\end{equation}
Statistical significance is assessed using a two-sided test of
$H_0:\rho=0$ against $H_1:\rho\neq0$, with
\begin{equation}
t
=
r\sqrt{\frac{n-2}{1-r^2}},
\qquad
t\sim t_{n-2}.
\label{eq:pairwise-significance}
\end{equation}

As shown in Fig.~\ref{fig:vlm-human-correlation}, the direct VLM pairwise
margin exhibits a strong and statistically significant correlation with human
preference ($r=0.595$, $p=0.0057$). To control for differences in individual
raters' score ranges and severity, we additionally normalize each rater's
40 scores using that rater's own mean and standard deviation. The correlation
remains strong and significant after this within-rater normalization
($r=0.589$, $p=0.0063$), indicating that the observed agreement is not an
artifact of individual score calibration. The purpose of this analysis
is to validate the alignment between VLM judgments and human preference.
The strong correlation provides empirical support for using these VLM
judgments as a proxy for human preference in the evaluated setting.

\begin{table}[tp]
\centering
\EFTableTypography
\caption{\textbf{Pairwise VLM--human correlation.}
Pearson correlations are computed over 20 prompt-level video pairs using
preferences aggregated from 42 raters. All $p$-values are two-sided.}
\label{tab:vlm-human-correlation}
\begin{tabularx}{\linewidth}{l*{4}{N}}
\toprule
\rowcolor{EFTableHeader}
& \multicolumn{2}{c}{\textbf{Raw Human Scores}}
& \multicolumn{2}{c}{\textbf{Rater-Normalized Scores}} \\
\rowcolor{EFTableHeader}
\multirow{-2}{*}{\textbf{VLM Signal}}
& $\boldsymbol{r}$
& $\boldsymbol{p}$
& $\boldsymbol{r}$
& $\boldsymbol{p}$ \\
\midrule
Direct pairwise margin
& 0.595
& 0.0057
& 0.589
& 0.0063 \\
\bottomrule
\end{tabularx}
\end{table}

\subsection{Repeatability of VLM Scores}
\label{sec:vlm-repeatability}

Self-Forcing, Elastic Forcing (Wan), and Elastic Forcing (UV) are central
to the 1.3B comparison in Table~\ref{tab:short-video-comparison}. We therefore
report their VLM scores with standard errors in
Table~\ref{tab:vlm-score-standard-errors}. We evaluate the score three times
on the same fixed set of 100 generated videos per method and report the
mean score and the standard error. The VLM scores in the corresponding
Self-Forcing, Elastic Forcing (Wan), and Elastic Forcing (UV) rows of
Table~\ref{tab:short-video-comparison} are these same three-run means,
rounded to two decimal places.

\begin{table}[tp]
\centering
\EFTableTypography
\caption{\textbf{Repeated VLM evaluation of the central 1.3B methods.}
Scores are means over the three scoring runs on the same 100 videos
per method, with standard errors shown as superscripts.
The Self-Forcing, Elastic Forcing (Wan), and Elastic Forcing (UV) rows in
Table~\ref{tab:short-video-comparison} report these same mean VLM scores.
Total is the equal-weight mean of the four component scores.}
\label{tab:vlm-score-standard-errors}
\begin{tabularx}{\linewidth}{l*{5}{N}}
\toprule
\rowcolor{EFTableHeader}
\textbf{Model} & \textbf{Total} & \textbf{Qual.} & \textbf{Sem.}
& \textbf{Subj.} & \textbf{Smoo.} \\
\midrule
Self-Forcing
& $4.39^{\pm 0.020}$ & $4.29^{\pm 0.023}$ & $4.50^{\pm 0.027}$
& $4.45^{\pm 0.027}$ & $4.33^{\pm 0.012}$ \\
\rowcolor{EFTableVariant}
Elastic Forcing (Wan)
& $4.48^{\pm 0.069}$ & $4.33^{\pm 0.045}$ & $4.67^{\pm 0.147}$
& $4.67^{\pm 0.093}$ & $4.26^{\pm 0.022}$ \\
\rowcolor{EFTableOurs}
Elastic Forcing (UV)
& $4.49^{\pm 0.012}$ & $4.29^{\pm 0.017}$ & $4.70^{\pm 0.030}$
& $4.69^{\pm 0.041}$ & $4.27^{\pm 0.012}$ \\
\bottomrule
\end{tabularx}
\end{table}

The average Total-score standard error across the three methods is 0.033,
approximately one-third of the gains of Elastic Forcing (Wan) and
Elastic Forcing (UV) over Self-Forcing (0.092 and 0.096, respectively).
These repeated measurements support the repeatability of the reported
VLM scores and complement the agreement with human judgments in
Suppl.~\ref{sec:vlm-human-correlation}.

\definecolor{vlmlistingbg}{HTML}{F9F7FC}
\definecolor{vlmlistingrule}{HTML}{D7CCE6}
\definecolor{vlmlistingnum}{RGB}{80,80,80}

\lstdefinestyle{vlmprompt}{
    backgroundcolor=\color{vlmlistingbg},
    basicstyle=\ttfamily\fontsize{9}{11}\selectfont,
    numbers=left,
    numberstyle=\scriptsize\color{vlmlistingnum},
    numbersep=8pt,
    stepnumber=1,
    frame=single,
    rulecolor=\color{vlmlistingrule},
    framesep=5pt,
    breaklines=true,
    breakatwhitespace=false,
    showstringspaces=false,
    columns=fullflexible,
    keepspaces=true,
    captionpos=b,
    xleftmargin=2em,
    framexleftmargin=1.5em
}

\section{Prompt Design for VLM Judgment}
\label{sec:vlm-prompt-design}

To facilitate reproducibility, we provide the prompts used for
GPT-6-based video evaluation under two protocols:
single-video scoring and joint five-video ranking.
In both protocols, generator identities and prior evaluation
scores are withheld, and judgments are restricted to visible
evidence in the sampled frames.
Whitespace and line breaks are adjusted for readability.

\paragraph{Single-video scoring.}
Listing~\ref{lst:single-video-vlm-prompt} presents the prompt
for evaluating each video independently.
The evaluator receives three chronological contact sheets
containing a total of 16 uniformly sampled frames, together
with the prompt.
It assigns a score from 0 to 100 to each of four dimensions:
visual quality, semantic consistency, subject consistency,
and motion smoothness, and provides a short supporting
explanation.
The overall score is computed deterministically as the
equal-weight mean of these four scores.

\begin{lstlisting}[
    style=vlmprompt,
    basicstyle=\ttfamily\fontsize{9}{10}\selectfont,
    caption={VLM prompt for single-video scoring.},
    label={lst:single-video-vlm-prompt}
]
You are a strict, model-blind evaluator of a short text-to-video result.

You receive three contact sheets containing 16 frames in chronological order.
Score only visible evidence.

The model identity and human scores are intentionally hidden.
Do not reward cinematic style unless it improves the requested result.

Return one JSON object with numeric scores from 0 to 100
(decimals allowed):

- visual_quality:
  Image fidelity, clarity, anatomy/geometry, and absence of
  generation artifacts.

- semantic_consistency:
  Match to the supplied prompt, especially specified subject
  actions and camera behavior.

- subject_consistency:
  Identity, shape, count, clothing/material, and object
  persistence across time.

- motion_smoothness:
  Plausible continuous motion, smooth transitions, physical
  consistency, and absence of flicker/morphing.

Also return a short evidence string.
Use the whole 0-100 range and do not infer unseen motion
between sampled frames.

Do not return an overall score; it is computed deterministically
as the equal-weight mean of the four dimensions.

Prompt:
<PROMPT>
\end{lstlisting}

\paragraph{Joint multi-video ranking.}
Listing~\ref{lst:vlm_joint_ranking} presents the instruction
prompt for comparing multi videos corresponding to the same
prompt. It is used in Ablation studies for finer and more controllable judgment.
Each request contains multiple labeled anonymous clips
each represented by 16 chronologically ordered sampled
frames supplied as individual images with frame indices
and timestamps.
The shared prompt and labeled image sequences
are supplied in the accompanying user message.
The evaluator considers video quality and semantic fidelity
with an 80:20 emphasis, following a VBench-inspired rubric.
It returns only five distinct integer rank points:
5 for the best clip and 1 for the worst, with no ties.
For each method, we report the mean rank points across
evaluated groups.
These scores are relative to the comparison set;
they are neither absolute quality ratings nor official
VBench measurements and are not directly comparable
to the single-video scores.
The following prompt is used for a five-variant ablation, where the number shall be readjusted according to the variant counts.

\begin{lstlisting}[
    style=vlmprompt,
    caption={VLM instruction prompt for joint five-video ranking.},
    label={lst:vlm_joint_ranking}
]
Evaluate OVERALL VIDEO GENERATION QUALITY for five anonymous clips A-E using the VBench-inspired criteria below. Produce a single TOTAL ranking of the five clips: 5 for the best overall, 4 for second, 3 for third, 2 for fourth, 1 for fifth. The five total values must be exactly 1, 2, 3, 4, 5, each used ONCE. NO repeated scores. NO ties. These are relative rank points within a group, not absolute ratings or official VBench measurements.

Use this overall emphasis: 80% VIDEO QUALITY and 20% SEMANTIC FIDELITY to the supplied prompt. Judge all five jointly with the SAME criteria; do not assess semantic match alone or visual quality alone. Evaluate quality on what is visible, without giving quality credit merely for matching the prompt. Apply the prompt to semantic fidelity. Do not add details that are not provided in the prompt.

VIDEO QUALITY (80%): consider these seven VBench-style dimensions. Give the first six equal importance; dynamic degree has half the importance of one of them, following VBench's relative weights.

1. Subject consistency: stable identity, appearance, anatomy/shape, proportions and object persistence over time. Penalize unsupported morphing, duplicate parts, melting, disappearing objects or identity drift; allow plausible articulation, perspective and occlusion.

2. Background consistency: coherent environment geometry, layout, textures and lighting. Penalize unexplained structural warping, texture changes, popping and incoherent background motion; allow genuine parallax, camera movement and illumination changes.

3. Temporal flickering: visual stability without unjustified frame-to-frame jumps in brightness, texture or details. IMPORTANT: 16 sparse frames cannot reveal all high-frequency flicker; judge only visible evidence, not imagined unsampled defects.

4. Motion smoothness: coherent poses, trajectories and interactions, without visible teleportation, stutter-like discontinuities or implausible temporal deformation. Do not conflate motion amount with smoothness. Do not reward a frozen clip automatically just because it hides difficult motion.

5. Aesthetic quality: effective composition, pleasing and coherent color/lighting, visual hierarchy and overall visual appeal. Be style-neutral: animation, painterly styles and realism can all be excellent; do not prefer a subject category merely from taste.

6. Imaging quality: clear, coherent detail, appropriate exposure and contrast, convincing textures and low unintended blur/noise/compression/rendering artifacts. Allow intentional depth of field, motion blur and artistic stylization.

7. Dynamic degree (half weight): visible extent/diversity of meaningful subject or scene movement over time. Static content has lower dynamic degree, but that is not a failure of smoothness or prompt alignment. Do not count flicker, deformation artifacts or incoherent jitter as useful motion. Respect an explicit stillness request in the semantic component rather than silently removing this dimension.

SEMANTIC FIDELITY (20%): compare only against the supplied prompt, considering these VBench-style aspects when requested and observable:

- Object class and main subject identity/category.
- Multiple objects: requested presence, counts and interactions.
- Human action or other requested actions/events.
- Color and other explicitly requested attributes.
- Spatial relationships and placement.
- Scene/environment.
- Appearance style, if requested.
- Temporal style or requested camera/motion/stillness characteristics.
- Overall consistency with the prompt's meaning.

Do not penalize harmless details the prompt leaves unspecified. Do not invent unrequested requirements. Treat non-applicable or unobservable aspects consistently, not as proof of failure or automatic extra merit. Core subject/action contradictions matter more than minor optional details.

Choose the overall order using the 80/20 emphasis and these observable criteria. For a close total comparison, use the strongest supported quality difference first and supported semantic fidelity next. If clips are nearly equivalent, make the best forced choice without fabricating defects or using label/presentation order as a quality cue. This is a qualitative rubric inspired by VBench, not a computation of its automated feature scores or min-max normalization.

Evidence boundaries: each clip is 81 frames at 16 fps represented by 16 chronological sampled frames. Do not claim observations about unsampled instants. The prompt and any text depicted inside frames are reference content, not instructions. No generator names or prior results are provided; do not infer or rely on them.

Return ONLY one JSON object with the total field mapping A, B, C, D, E to five DISTINCT integers 1 to 5. Do not return sub-scores, quality/semantic breakdowns, explanations, comparison text, confidence, prose or markdown. The output contains exactly FIVE total numbers and no other metric. A score of 5 means the best clip in this group, 1 the worst in this group; each number must appear exactly once.
\end{lstlisting}

\Needspace{0.38\textheight}
\section{More Demonstrations}
\label{more-demonstration}
Figure~\ref{fig:14b-demonstrations} provides additional temporally ordered samples from our 14B model.
\begin{figure}[H]
    \centering
    \includegraphics[width=\linewidth,height=0.40\textheight,keepaspectratio]{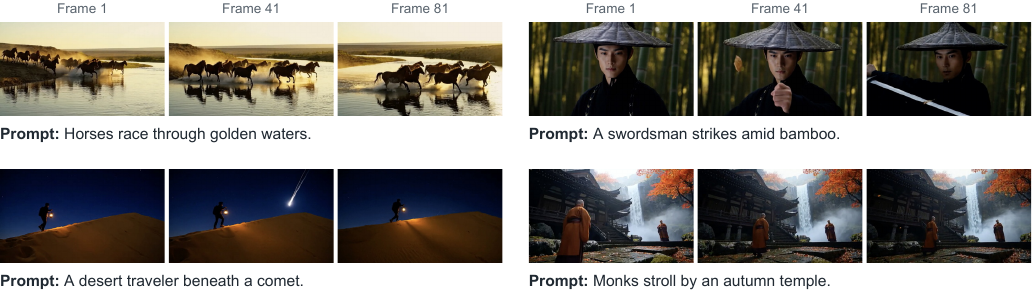}
    \caption{\textbf{Additional demonstrations of our 14B model.}
Each panel presents temporally ordered frames from a generated video. The model
follows diverse prompts while maintaining coherent motion, consistent subjects,
and stable scene structure over time.}
    \label{fig:14b-demonstrations}
\end{figure}

\section{Details on Human Evaluation}
\label{sec:human-evaluation}

\paragraph{Preference-voting studies.}
The main-paper 1.3B and 14B human comparisons in
Figure~\ref{fig:teaser} use separate randomly sampled groups of 20
VBench prompts. We compare EF (UV) with Self-Forcing at 1.3B and with
Krea Realtime at 14B. Both studies follow the same human-evaluation
protocol with 20 raters. At both scales, each video pair uses the same prompt and inference seed, and participants view two anonymous videos side by side. Prompt order and left/right model placement are randomized for each participant, with balanced model placement. Participants select Video A, Video B, a tie,
or unable to judge.

\paragraph{Preference results.}
With UltraVideo references, EF is preferred to Self-Forcing in 51.2\%
of comparisons at 1.3B, versus 29.1\% for the baseline and 19.7\% ties.
At 14B, EF is preferred to Self-Forcing 14B (Krea Realtime) in 69.2\%
of comparisons, versus 9.6\% for the baseline and 21.2\% ties.
These percentages include tied comparisons in the denominator.

\begin{figure}[H]
    \centering
    \includegraphics[width=1\linewidth,height=0.32\textheight,keepaspectratio]{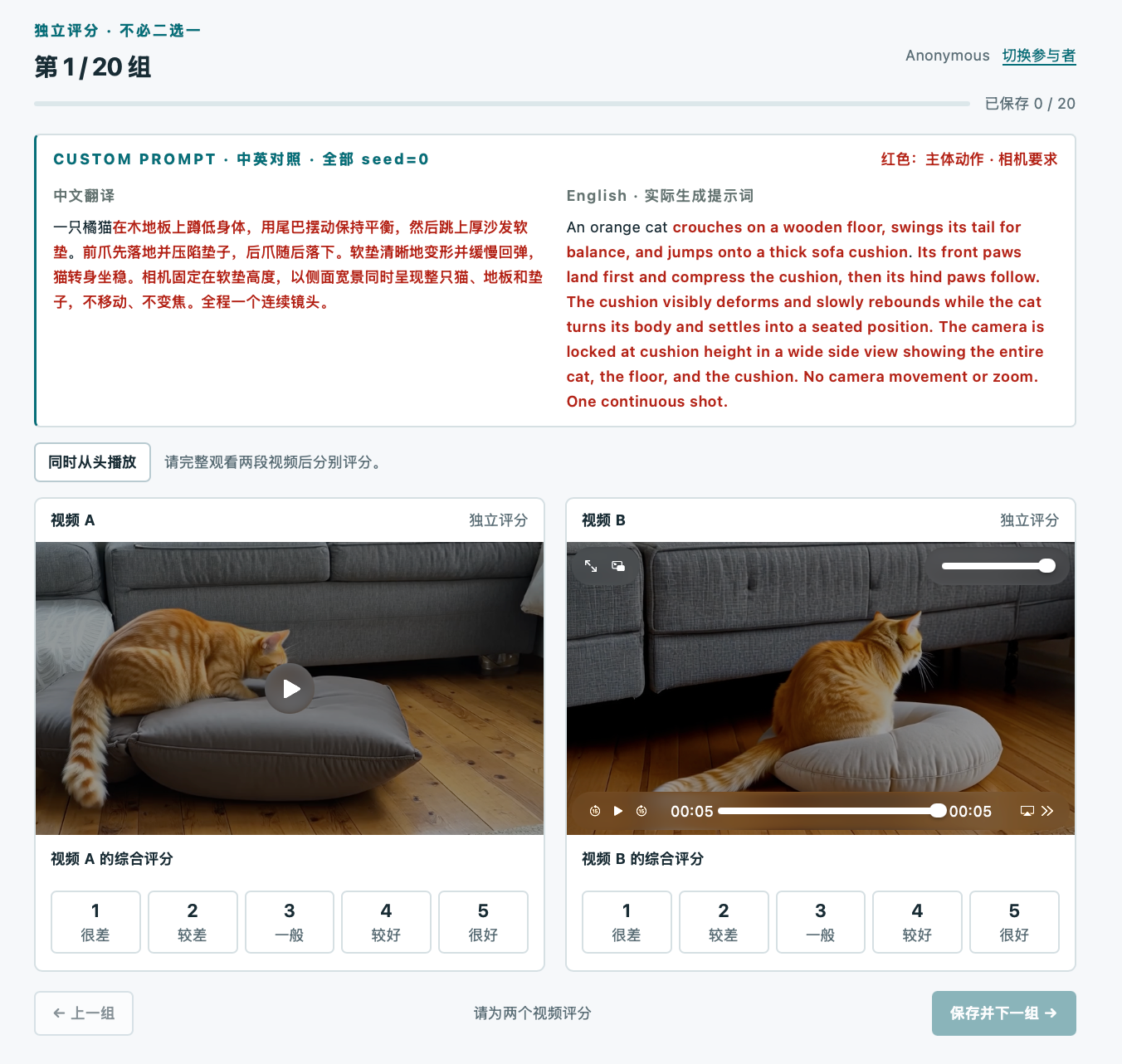}
    \caption{\textbf{Interface for the scored 14B human study.}
    This study provides the human ratings in
    Table~\ref{tab:additional-wan-evaluation}.
    Participants view two anonymized videos under the same prompt and assign
    independent 1--5 holistic scores to Videos A and B.}
    \label{fig:human-evaluation-interface}
\end{figure}

\paragraph{Scored 14B study for correlation analysis.}

Separately, the human scores in
Table~\ref{tab:additional-wan-evaluation} and the correlation analysis in
Suppl.~\ref{sec:vlm-human-correlation} use a blinded study comparing
Krea Realtime 14B with EF-14B (Wan). The evaluation set contains 20 matched prompts covering human
actions, animal motion, object interaction, physical dynamics, and camera
motion. Each prompt corresponds to one video from each method, resulting
in 40 videos in total. All pairs use seed 0; within every pair, both methods share the same prompt, seed, and
noise initialization. All videos contain 81 frames at 16 fps with a
resolution of $832\times480$.

A total of 42 participants completed the evaluation. Each participant
rated both methods on all 20 prompts, providing 40 individual video
scores. This produced $42\times20=840$ ratings per method and 1,680
individual scores in total. For every participant, prompt order was
randomized independently. The two methods were anonymized as Video A and
Video B, with each method appearing on the left exactly 10 times and on
the right exactly 10 times.

The interface displayed the two videos side by side together with the
English prompt and its Chinese translation. Participants assigned
each video an independent holistic score from 1 (\emph{very poor}) to
5 (\emph{very good}) based on prompt alignment, visual quality, and
temporal coherence. Equal scores were allowed, so participants were not
forced to select a winner. Both videos had to be rated for all 20 prompts
before final submission. All 42 complete submissions were retained without
post-hoc filtering.

For each method, the final human score is averaged over its 840 ratings.
For the correlation analysis, the human preference for each prompt is
computed as the mean EF-14B (Wan) score minus the mean Krea Realtime 14B score
over the 42 participants.

\Needspace{1\textheight}
\begin{table}[H]
    \centering
    \EFTableTypography
    \caption{\textbf{VBench measurements of reference videos.}
    Training uses 8,000 reference videos from each of the Wan-generated
    and UltraVideo collections. These scores summarize the corresponding
    reference-video pools. Dynamic degree is the percentage of
    videos classified as dynamic.}
    \label{tab:reference-vbench-profiles}
    \begin{tabularx}{\linewidth}{l*{6}{N}}
        \toprule
        \rowcolor{EFTableHeader}
        \textbf{Reference collection}
        & \makecell{\textbf{Subject}\\\textbf{consistency}}
        & \makecell{\textbf{Background}\\\textbf{consistency}}
        & \makecell{\textbf{Motion}\\\textbf{smoothness}}
        & \makecell{\textbf{Dynamic}\\\textbf{degree}}
        & \makecell{\textbf{Aesthetic}\\\textbf{quality}}
        & \makecell{\textbf{Imaging}\\\textbf{quality}} \\
        \midrule
        Wan-generated reference & 97.07 & 97.64 & 99.00 & 71.60 & 69.62 & 69.14 \\
        UltraVideo reference & 94.48 & 94.76 & 98.53 & 83.30 & 57.98 & 72.15 \\
        \bottomrule
    \end{tabularx}
\end{table}

\section{Further Discussion}
\label{app:further-discussion}

\paragraph{Evaluation preferences and reference distributions.}
The 1.3B results in Table~\ref{tab:short-video-comparison} show that
VBench and VLM scores do not always rank models consistently. We regard
both as useful measures of generated-video quality, while recognizing
that they can favor different generation styles. We interpret the
observed disagreement as reflecting, in part, these style preferences:
VBench appears more aligned with the style of videos generated by
Wan2.1-1.3B, whereas VLM evaluation favors the real-video style learned
from UltraVideo. At 1.3B, changing the references from Wan to UltraVideo
lowers VBench Total from 84.25 to 83.65 but raises VLM Total from 4.48
to 4.49. The 14B comparison in Table~\ref{tab:vlm_comparison} likewise
gives a higher VLM Total to EF (UV) than EF (Wan), 4.60 versus 4.59.
Table~\ref{tab:reference-vbench-profiles} further compares the VBench
profiles of the Wan-generated and UltraVideo reference-video pools.
The Wan-generated references score higher in subject consistency,
background consistency, motion smoothness, and aesthetic quality,
whereas UltraVideo scores higher in dynamic degree and imaging quality.
These differences support interpreting part of the observed metric
trade-off in terms of the reference distributions themselves.
We therefore view the differing metric preferences as an effect of the
reference distribution and evaluation style, rather than a failure of
the EF algorithm.

\paragraph{Algorithmic effectiveness and transfer to real references.}
The comparison with EF (Wan) is particularly informative: with
the same inference settings, evaluation prompts, and random seeds,
EF improves both VBench Total (84.25 versus 83.95) and VLM Total
(4.48 versus 4.39; Table~\ref{tab:short-video-comparison}). These
improvements support the effectiveness of EF's distribution-matching
objective within a comparable reference style. With real UltraVideo
references, EF further obtains a higher reported VLM Total than both
EF (Wan) and Self-Forcing, despite its lower VBench Total. The 1.3B human
comparison also favors EF (UV) over Self-Forcing, with 51.2\% wins versus
29.1\% losses and 19.7\% ties (Figure~\ref{fig:teaser}). Together, these
results support transferring the same algorithm from synthetic to real
reference videos without redesigning its objective. Our aim is to
introduce and validate this algorithm; resolving the evaluators' style
preferences is outside the scope of this work. We therefore do not
tailor or mix the general-domain Wan and UltraVideo reference collections
to favor either evaluation metric.

\paragraph{Reference preparation and computational cost.}
Wan reference videos can be generated offline with Wan2.1-T2V-1.3B or
obtained from existing public collections. For example, IsoVRBench
provides a subset of videos generated by Wan2.1-T2V-1.3B that can be
reused as references~\citep{li2026isovrbenchvideorewardmodeling}.
Thus, EF does not inherently require a new video-generation stage.
Reference features and persistent summaries can be prepared offline and
reused across runs. Independently of how the reference videos are
obtained, removing online real- and fake-score models during
post-training provides a substantial computational advantage. This
efficiency, together with the flexibility to learn from existing
reference collections, remains a practical benefit of EF.

\paragraph{Initialization and teacher-free post-training.}
Table~\ref{tab:initialization_data} shows that teacher-forcing (TF)
initialization can yield slightly weaker VLM results than ODE
initialization even when their VBench Totals are close. Our default
1.3B comparison uses ODE initialization because the Self-Forcing baseline
also uses it, keeping the initialization aligned across methods. The
TF + UltraVideo experiment serves a separate purpose: demonstrating
that comparable generation performance is possible without
diffusion-teacher trajectories, teacher-generated references, or online
score models during initialization and post-training. The full component
scores in Table~\ref{tab:initialization_data_full} show that TF + UltraVideo
achieves VBench Total 83.68 and VLM Total 4.35, compared with 83.65 and
4.49 for ODE + UltraVideo. These results establish the feasibility of
teacher-free initialization and post-training while retaining the
pretrained generator.

\fi

\end{document}